\RequirePackage{fix-cm}
\documentclass[twocolumn]{svjour3}    
\pdfoutput=1

\usepackage{graphicx}
\usepackage{natbib}

\usepackage{url}            
\usepackage{booktabs}       
\usepackage{amsfonts}       
\usepackage{nicefrac}       
\usepackage{microtype}      
\usepackage{graphicx}
\usepackage{enumitem}
\usepackage{bm}
\usepackage{nccmath}
\usepackage{boldline}
\usepackage{booktabs} 
\usepackage{caption}

\usepackage{algorithm}
\usepackage[noend]{algcompatible}

\usepackage{capt-of}
\usepackage{color,soul}
\usepackage[pagebackref=false,breaklinks=true,colorlinks,bookmarks=false,urlcolor=blue, citecolor=blue]{hyperref}

\usepackage{times}
\usepackage{epsfig}
\usepackage{multirow}
\usepackage{xcolor}
\usepackage{amsmath}
\usepackage{amssymb}
\usepackage{wrapfig}
\usepackage{comment}

\usepackage{textcomp}
\usepackage{bm}
\usepackage{soul}

\usepackage{makecell}
\usepackage{subfigure}
\usepackage{arydshln}
\usepackage{ulem}

\newcommand{\ie}{\textit{i}.\textit{e}.}
\newcommand{\eg}{\textit{e}.\textit{g}.}
\newcommand{\etc}{\textit{etc}.}

\useunder{\underline}{\ul}{}
\usepackage{xcolor}

\usepackage{tikz}

\begin{document}

\title{Benchmarking and Analysis of Unsupervised Object Segmentation from Real-world Single Images}

\author{Yafei Yang \and Bo Yang}

\institute{Yafei Yang \and Bo Yang \at
         vLAR Group, Department of Computing \\ 
         The Hong Kong Polytechnic University \\ 
         \email{ya-fei.yang@connect.polyu.hk; bo.yang@polyu.edu.hk}
}
\date{Received: date / Accepted: date}

\maketitle

\sethlcolor{white}\hl{}

\begin{abstract}
In this paper, we study the problem of unsupervised object segmentation from single images. We do not introduce a new algorithm, but systematically investigate the effectiveness of existing unsupervised models on challenging real-world images. We first introduce seven complexity factors to quantitatively measure the distributions of background and foreground object biases in appearance and geometry for datasets with human annotations. With the aid of these factors, we empirically find that, not surprisingly, existing unsupervised models fail to segment generic objects in real-world images, although they can easily achieve excellent performance on numerous simple synthetic datasets, due to the vast gap in objectness biases between synthetic and real images. By conducting extensive experiments on multiple groups of ablated real-world datasets, we ultimately find that the key factors underlying the failure of existing unsupervised models on real-world images are the challenging distributions of background and foreground object biases in appearance and geometry. Because of this, the inductive biases introduced in existing unsupervised models can hardly capture the diverse object distributions. Our research results suggest that future work should exploit more explicit objectness biases in the network design. 

\end{abstract}

\section{Introduction}
The ability of automatically identifying individual objects from complex visual observations is a central aspect of human intelligence \citep{Spelke1992}. It serves as the key building block for higher-level cognition tasks such as planning and reasoning \citep{Greff2020}. In last years, a plethora of models have been proposed to segment objects from single static images in an unsupervised fashion: from the early AIR \citep{Eslami2016}, MONet \citep{Burgess2019} to the recent SPACE \citep{Lin2020}, SlotAtt \citep{Locatello2020}, GENESIS-V2 \citep{Engelcke2021}, \etc{} They jointly learn to represent and segment multiple objects from a single image, without needing any human annotations in training. This process is often called perceptual grouping/binding or object-centric learning. These methods and their variants have achieved impressive segmentation results on numerous synthetic datasets such as dSprites \citep{Matthey2017} and CLEVR \citep{Johnson2017}. Such advances come with great expectations that the unsupervised techniques would likely close the gap with fully-supervised methods for real-world visual understanding. However, few work has systematically investigated the true potential of the emerging unsupervised object segmentation models on complex real-world images such as COCO dataset \citep{Lin2014}. This naturally raises an essential question:

\textit{Is it promising or even possible to segment generic objects from real-world single images using (existing) unsupervised methods?} 

\textbf{What is an object?} 
To answer the above question involves another fundamental question: what is an object? Exactly 100 years ago in Gestalt psychology, Wertheimer \citep{Wertheimer1923} first introduced a set of \textit{perceptual grouping} and \textit{figure-ground organization} principles to heuristically define visual data as individual objects and/or backgrounds. Perceptual grouping principles such as proximity and similarity specify how foreground visual elements are grouped into individual objects, whereas figure-ground organization principles investigate how visual data are just separated as the foreground against background, \eg{} a relatively small area tends to be seen as foreground \citep{Wagemans2012}.

However, these principles are highly subjective, whilst the real-world scenes and objects are far more complex with extremely diverse appearances and shapes. Therefore, it is practically impossible to quantitatively define what is an object, \ie{}, the objectness, from visual inputs (\eg{}, a set of image pixels). Nevertheless, to thoroughly understand whether unsupervised methods can truly learn objectness akin to the psychological process of humans, it is vital to investigate the underlying factors that potentially facilitate or otherwise hinder the ability of unsupervised models. In this regard, by drawing on Gestalt principles, we instead define a series of new factors to quantitatively measure the complexity of individual foreground objects and the background in Section \ref{sec:complexity_factors}. By taking into account both appearance and geometry, our complexity factors explicitly assess the difficulty of segmenting single objects and the background. For example, it is harder to segment a chair with colorful textures from a cluttered background than a single-color ball from a clean background for unsupervised methods. With the aid of these factors, we extensively study whether and how existing unsupervised models can discover objects in Section \ref{sec:experimental_res}. 

\begin{figure}[t]
\setlength{\abovecaptionskip}{ 0 pt}
\setlength{\belowcaptionskip}{ -6 pt}
\centering
   \includegraphics[width=1\linewidth]{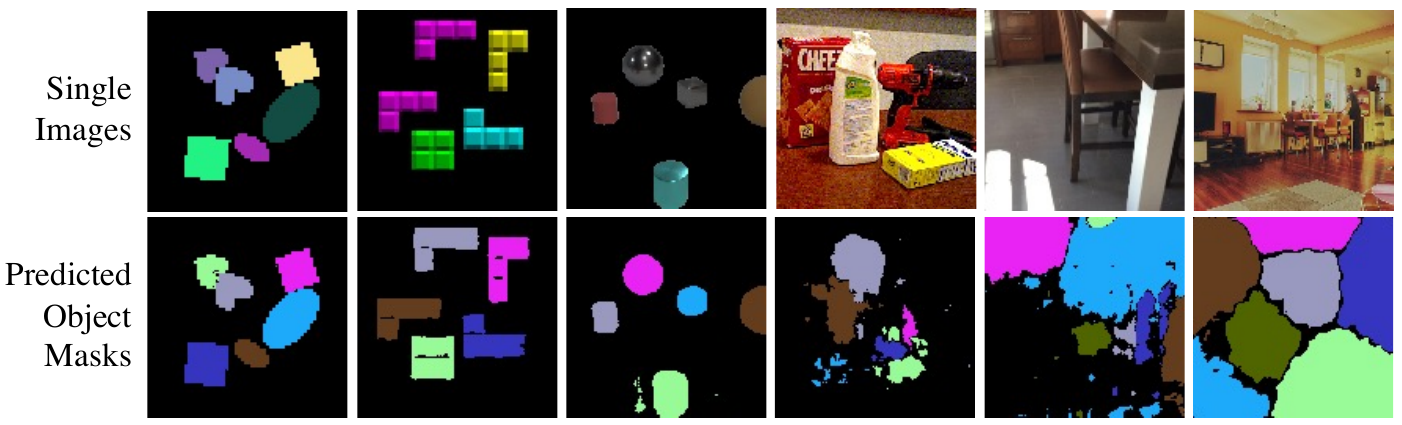}
\caption{SlotAtt can perfectly segment simple objects on three synthetic images (left-hand side) but clearly fails on three real-world images (right-hand side).}
\label{fig:opening}
\vspace{-0.5cm}
\end{figure}

\textbf{What is the problem of unsupervised object segmentation from single images?} A large number of models \citep{Yuan2022} aim to tackle the problem of unsupervised object segmentation from single images. They share several key problem settings: 1) all training images do not have any human annotations; 2) every single image has multiple objects optionally with a textured background; 3) each image is treated as a static data point without any dynamic or temporal information; 4) all models are trained from scratch without requiring any pretrained networks on additional datasets. Ultimately, the goal of these models is to segment all individual objects as accurate as the ground truth human annotations. In this paper, we regard these settings as the basic and necessary part of unsupervised object segmentation from single images, and empirically evaluate how successfully the existing models can exhibit on real-world images.

\textbf{Contributions and findings.} This paper addresses the essential question regarding the potential of unsupervised segmentation of generic objects from real-world single images. Our contributions are:
\begin{itemize}[leftmargin=*]
\setlength{\itemsep}{1pt}
\setlength{\parsep}{1pt}
\setlength{\parskip}{1pt}
    \item We first introduce {4} complexity factors to quantitatively measure the difficulty of individual objects and 3 factors to quantify backgrounds. These factors are key to investigating the true potential of existing unsupervised models.
    \item We extensively evaluate current unsupervised approaches in a large-scale experimental study. We implement 4 representative unsupervised methods and train more than {200} models on 4 groups of curated datasets from scratch. The datasets, code and pretrained models are available at {\small{\url{https://github.com/vLAR-group/UnsupObjSeg}}}   
    \item We analyze our experimental results and find that: 1) existing unsupervised object segmentation models cannot discover generic objects from single real-world images, although they can achieve outstanding performance on synthetic datasets, as qualitatively illustrated in Figure \ref{fig:opening}; 2) the challenging distributions of both foreground object and the background biases in appearance and geometry from real-world images are the key factors incurring the failure of existing models; 3) the inductive biases introduced in existing unsupervised models are fundamentally not matched with the objectness biases exhibited in real-world images, and therefore fail to discover the real objectness.  
\end{itemize}

\textbf{Related Work.} Recently, ClevrTex {\citep{Karazija2021}} and the concurrent work  {\citep{Papa2022}} also study unsupervised object segmentation on single images. Through evaluation on (complex) synthetic datasets only, both works focus on benchmarking the effectiveness of particular network designs of baselines. By comparison, our paper aims to explore what and how the objectness distribution gaps between synthetic and real-world images incur the failure of existing models. The recent work by \citep{Weis2021} which investigates video object discovery is orthogonal to our work as the motion signals do not exist in single images.     

\begin{figure*}[t]
\vspace{0.2cm}
\setlength{\abovecaptionskip}{ 5 pt}
\setlength{\belowcaptionskip}{ -9 pt}
\raisebox{0pt}[\dimexpr\height-1.\baselineskip\relax]{
  \centering
  \includegraphics[width=1\linewidth]{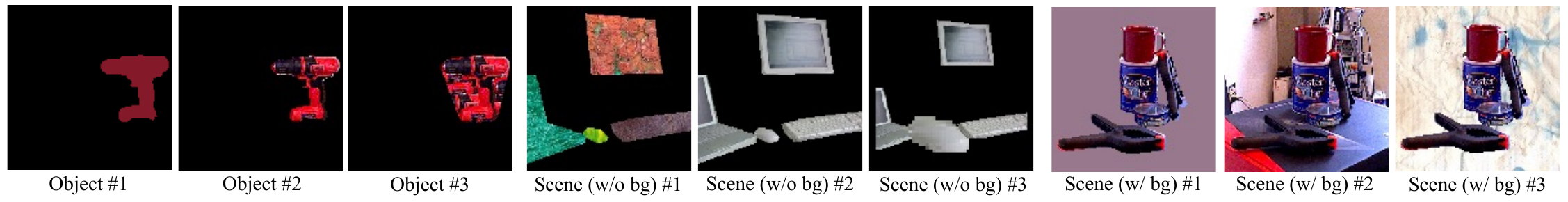}
}
\caption{Complexity in appearance and geometry for objects and scenes.}
\label{fig:main_complexity_factors}
\vspace{-0.2cm}
\end{figure*}

\textbf{Scope of this research.} 
\hl{In addition to our core study on purely unsupervised object segmentation approaches from single images, we also include discussions about more recent methods such as Odin \mbox{\citep{Henaff2022}}, DINOSAUR \mbox{\citep{seitzer2022bridging}}, FreeSOLO \mbox{\citep{wang2022freesolo}}, and CutLER \mbox{\citep{wang2023cut}} that require pretrained models to obtain objectness biases on monolithic object images such as ImageNet \mbox{\citep{Russakovsky2015}} in Section \mbox{\ref{sec:pretrained_model_discussion}}. However, this paper does not investigate object segmentation from saliency maps \mbox{\citep{Wang2021}}, static multi-views \mbox{\citep{Yuan2022}}, or dynamic videos  \mbox{\citep{Singh2022a,Singh2022}}, because the input information of these methods is clearly different from that on single images.}

A preliminary version of this work has been published in  \citep{Yang2022} and our new technical extensions include: 1) 3 additional complexity factors to quantitatively measure the difficulty of background in each image in Section \ref{sec:background_complexity_factors}, 2) new benchmarks with 6 more datasets, a new baseline, and 3 extra metrics in Section \ref{sec:exep_design}, 3) and more comprehensive experiments to analyze object segmentation of existing models.

\section{Complexity Factors}\label{sec:complexity_factors}

As illustrated in the left three images of Figure \ref{fig:main_complexity_factors}, an individual object, represented by a set of color pixels within a mask, can vary significantly given different types of appearance and geometric shape. A specific scene, represented by a set of objects placed on a clean canvas, can also differ vastly given different relative appearances and geometric layouts between objects, as illustrated in the middle images. If these objects are instead placed on diverse backgrounds, they tend to appear more differently as shown in the right images.

Unarguably, such variations and complexity of appearance and geometry in object level, scene level, and background level, directly affect human's ability to precisely separate all objects. Naturally, the performance of unsupervised segmentation models is also expected to be influenced by these variations. In this regard, we carefully define the following three groups of factors to quantitatively describe the complexity of visual scenes in datasets.

\subsection{Object-level Complexity Factors}
\label{sec:object_level_complexity_factors}

As to an object, all its information can be described by appearance and geometry. Therefore we define the below two factors to measure the complexity of appearance and geometry respectively. Notably, both factors are nicely invariant to the object scale. 

\begin{figure}[h]
\vspace{-0.4cm}
\setlength{\abovecaptionskip}{ -0 pt}
\setlength{\belowcaptionskip}{ -4 pt}
\centering
\includegraphics[width=0.8\linewidth]{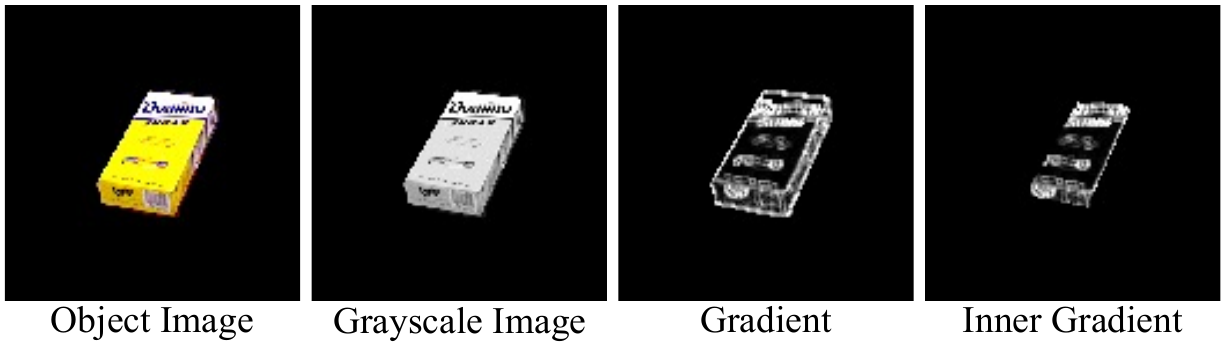}
\caption{Illustrations for Object Color Gradient.}
\label{fig:main_object_color_gradient}
\vspace{-0.2cm}
\end{figure}
    
\textbf{Object Color Gradient:} This factor aims to calculate how frequently the appearance changes within the object mask. As shown in Figure \ref{fig:main_object_color_gradient}, given the RGB image and mask of an object, we firstly convert RGB into grayscale and then apply Sobel filter \citep{Sobel1973} to compute the gradients horizontally and vertically for each pixel within the mask. The final gradient value is obtained by averaging out all object pixels. Note that, the object boundary pixels are removed to avoid the interference of background. Numerically, the higher this factor is, the more complex texture and/or lighting effect the object has, and therefore it is likely harder to segment.
    
\begin{figure}
\setlength{\abovecaptionskip}{ 0 pt}
\setlength{\belowcaptionskip}{ -4 pt}
\centering  
\includegraphics[width=0.8\linewidth]{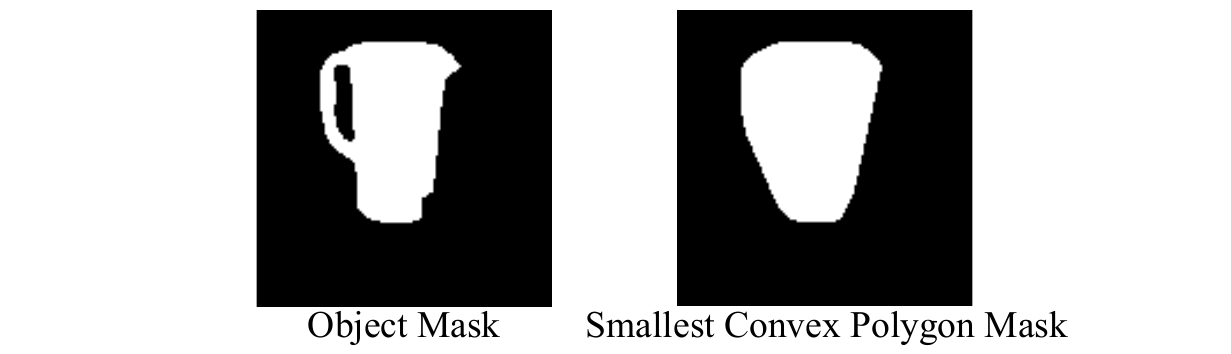}
\caption{\footnotesize{Illustrations for Object Shape Concavity.}}
\label{fig:main_object_shape_concavity}
\vspace{-0.2cm}
\end{figure}

\textbf{Object Shape Concavity:} This factor is designed to evaluate how irregular the object boundary is. As shown in Figure \ref{fig:main_object_shape_concavity}, given an object (binary) mask, denoted as $\boldsymbol{M}_{obj}\in \mathbb{R}^{H\times W}$, we firstly find the smallest convex polygon mask ($\boldsymbol{M}_{cvx} \in \mathbb{R}^{H\times W}$) that surrounds the object mask using an existing algorithm \citep{Eddins2011}, and then the object shape concavity value is computed as: $1-\sum\boldsymbol{M}_{obj} / \sum\boldsymbol{M}_{cvx}$. Clearly, the higher this factor is, the more irregular object shape is, and segmentation is trickier. 

\subsection{Scene-level Complexity Factors}
\label{sec:scene_level_complexity_factors}
Given an image, in addition to the object-level complexity, the spatial and appearance relationships between all objects can also incur extra difficulty for segmentation. We define the following two factors to quantify the complexity of relative appearance and geometry between objects in an image. 

\begin{figure}[h]
\vspace{-0.4cm}
\setlength{\abovecaptionskip}{ 2 pt}
\setlength{\belowcaptionskip}{ -8 pt}
\centering
\includegraphics[width=0.8\linewidth]{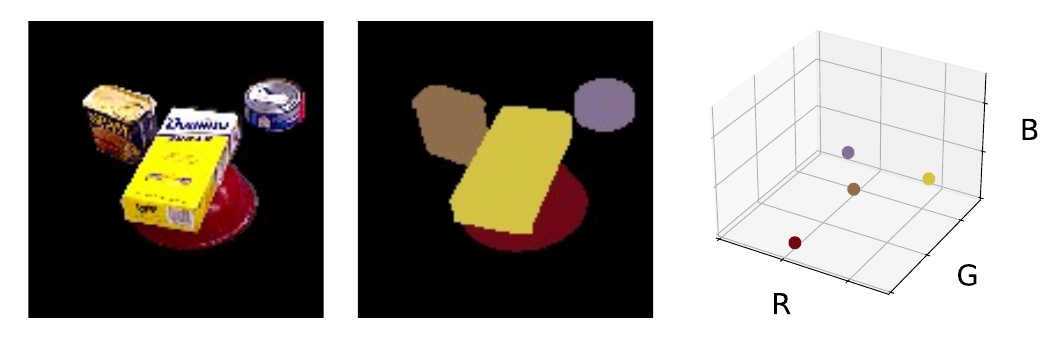}
\caption{Illustrations for Inter-object Color Similarity.}
\label{fig:main_inter_object_color_similarity}
\vspace{-0.3cm}
\end{figure}

\begin{figure*}[t]
\setlength{\abovecaptionskip}{ 2 pt}
\setlength{\belowcaptionskip}{ -8 pt}
\centering
   \includegraphics[width=0.8\linewidth]{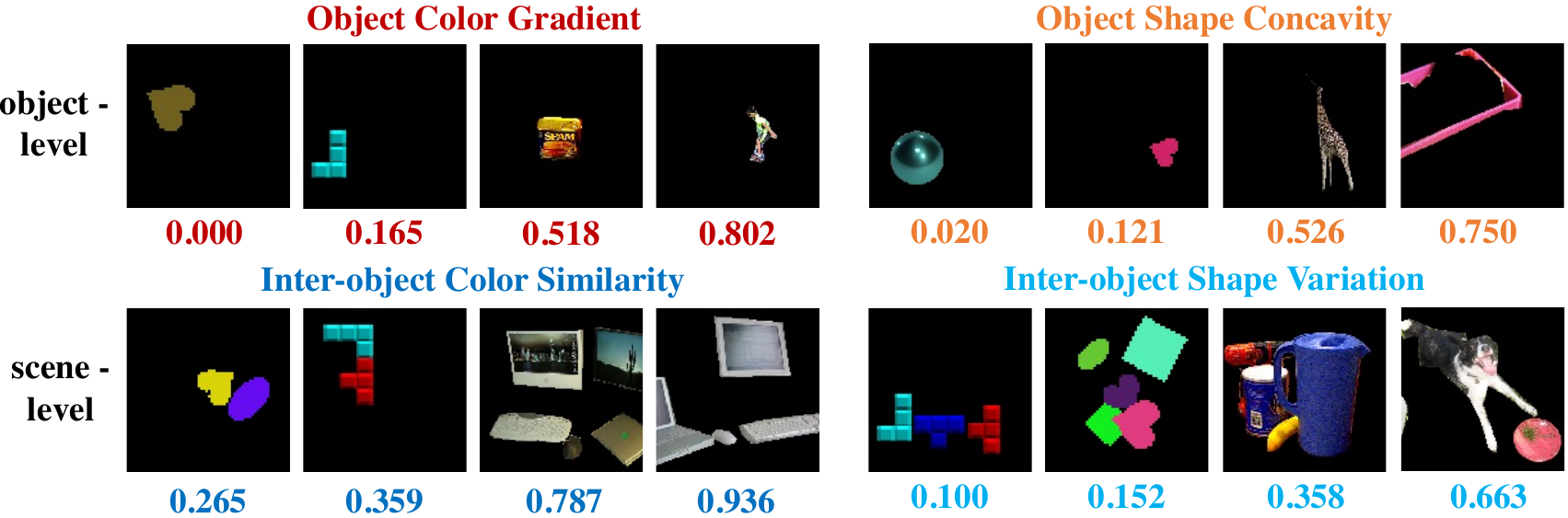}
\caption{\hl{Sample objects and scenes for the four factors at different complexity values.}}
\label{fig:main_factor_example}
\end{figure*}
    
\textbf{Inter-object Color Similarity:} This factor intends to assess the appearance similarity between all objects in the same image. As shown in Figure \ref{fig:main_inter_object_color_similarity}, we first calculate the average color for each object, and then compute the pair-wise Euclidean distances of object colors, obtaining a $K\times K$ matrix where $K$ represents the object number. The \textit{average color distance} is calculated by averaging the matrix excluding diagonal entries, and the final inter-object color similarity is computed as: $1-$\textit{average color distance}$/(255\times\sqrt{3})$. Intuitively, the higher this factor is, the more similar all objects appear to be, the less distinctive each object is, and the harder it is to separate each object.

\begin{figure}[h]
\vspace{-0.3cm}
\setlength{\abovecaptionskip}{ -0 pt}
\setlength{\belowcaptionskip}{ -2 pt}
\centering
\captionsetup{justification=centering}
\includegraphics[width=0.70\linewidth]{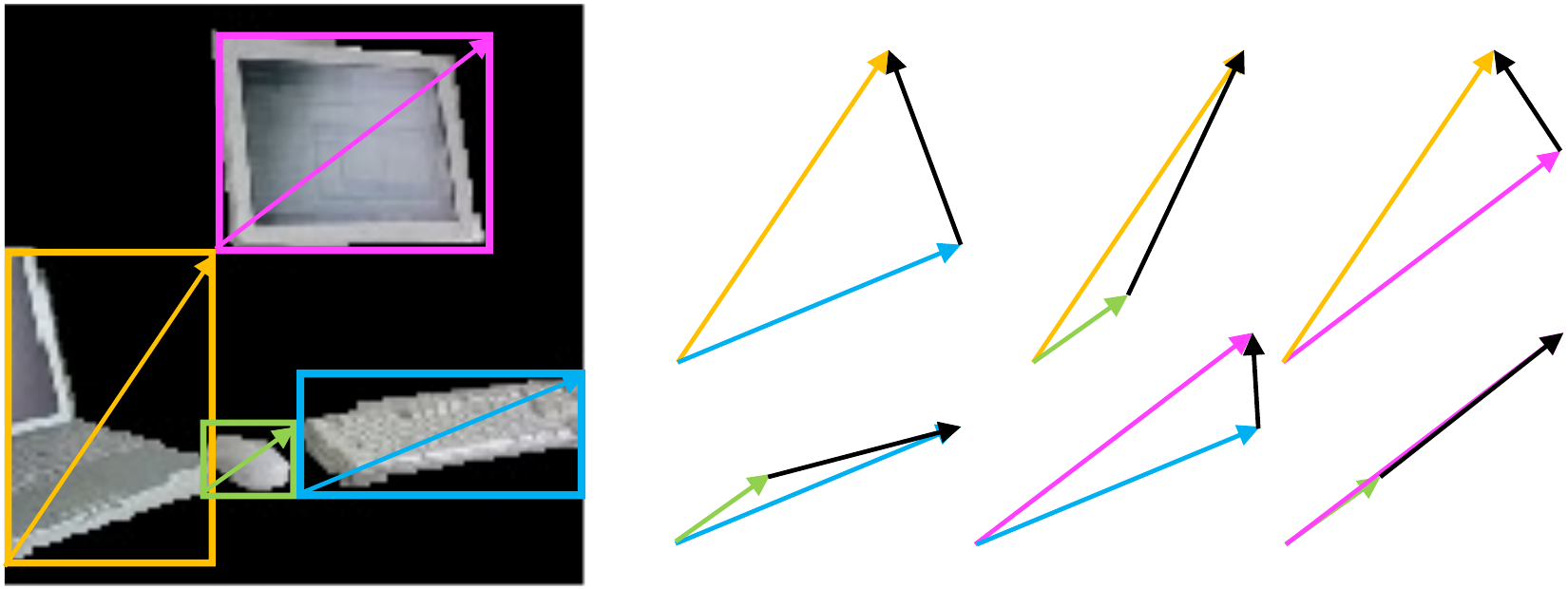}
\caption{\hl{Illustrations for Inter-object Shape Variation.}}
\label{fig:main_inter_object_shape_variation}
\vspace{-0.5cm}
\end{figure}

\textbf{Inter-object Shape Variation:} 
This factor aims to measure the relative geometry diversity between all $K$ objects in an image. \hl{We first find the diagonal vector of each object bounding box. Then we calculate the differences between each pair of diagonal vectors, resulting in $\frac{K \times (K-1)}{2}$ vectors (denoted as black vectors in Figure \mbox{\ref{fig:main_inter_object_shape_variation}}). The final inter-object shape variation is the averaged norm of these $\frac{K \times (K-1)}{2}$ vectors.} The higher this factor, objects within an image are more diverse with imbalanced sizes, and therefore segmenting both gigantic and tiny objects is likely harder.

By capturing the appearance and geometry of objects in both object and scene levels, the four factors in Sections \ref{sec:object_level_complexity_factors} \& \ref{sec:scene_level_complexity_factors} are designed to quantify the image complexity only with clean backgrounds. For illustration, Figure {\ref{fig:main_factor_example}} shows sample images for the four factors at different values. The higher the values, the more complex foreground objects are in both object and scene levels. 

In fact, these factors are carefully selected from more than 10 candidates because they are empirically more suitable to differentiate the gaps between synthetic and real-world images, and they eventually serve as key indicators to diagnose existing unsupervised models in Section \ref{sec:experimental_res}. 
Details of other candidates are in the appendix.

\subsection{Background Complexity Factors}
\label{sec:background_complexity_factors}
Since the diversity of image backgrounds also plays a critical role in potentially distinguishing or messing up foreground objects, we introduce the following three factors to measure the complexity of backgrounds. The background here is defined as a collection of all pixels which do not belong to any considered objects in a single image. 

\begin{figure}[h]
\vspace{-0.55cm}
\setlength{\abovecaptionskip}{ -0 pt}
\setlength{\belowcaptionskip}{ -2 pt}
\centering
\includegraphics[width=1\linewidth]{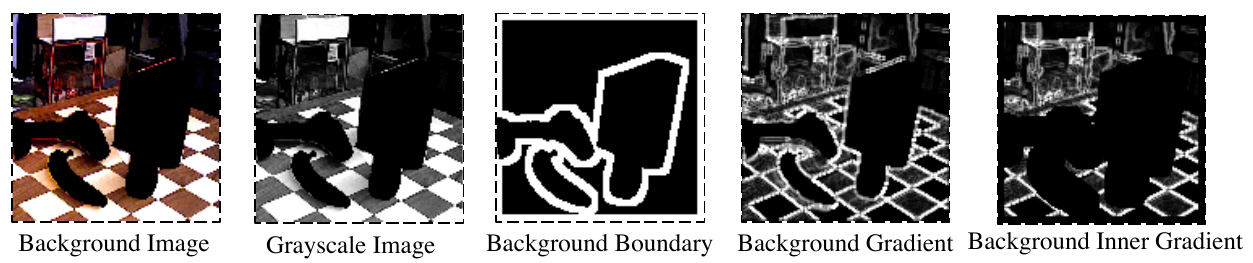}
\caption{Illustrations for Background Color Gradient.}
\label{fig:main_bg_color_gradient}
\vspace{-0.3cm}
\end{figure}

\textbf{Background Color Gradient:} 
This factor is designed to measure how frequently the appearance changes within background pixels. The calculation is the same Object Color Gradient as illustrated in Figure \ref{fig:main_bg_color_gradient}. The background boundary is also removed to avoid the interference of foreground objects. The higher this factor, the more complex texture and/or lighting of the background, and therefore both the background and objects may be more difficult to be segmented.

\textbf{Background-Foreground Color Similarity:} 
This factor aims to measure the appearance similarity between background and foreground pixels in an image. Specifically, we first calculate the Euclidean distance between each background pixel RGB and each foreground pixel RGB, resulting in a $U\times V$ matrix where $U$ and $V$ represent the number of pixels in background and foreground respectively. We then treat it as a form of cost matrix where the Hungarian algorithm is applied to obtain an optimal \textit{cost value} between all background and foreground pixels. The final background-foreground color similarity is computed as: 1 - \textit{cost value}. Intuitively, the higher this factor, the more similar background and foreground appear to be, and it is harder to separate each other. Details are in the appendix.

\begin{figure}[ht]
\vspace{-0.4cm}
\setlength{\abovecaptionskip}{ 2 pt}
\setlength{\belowcaptionskip}{ -4 pt}
\centering
\includegraphics[width=0.95\linewidth]{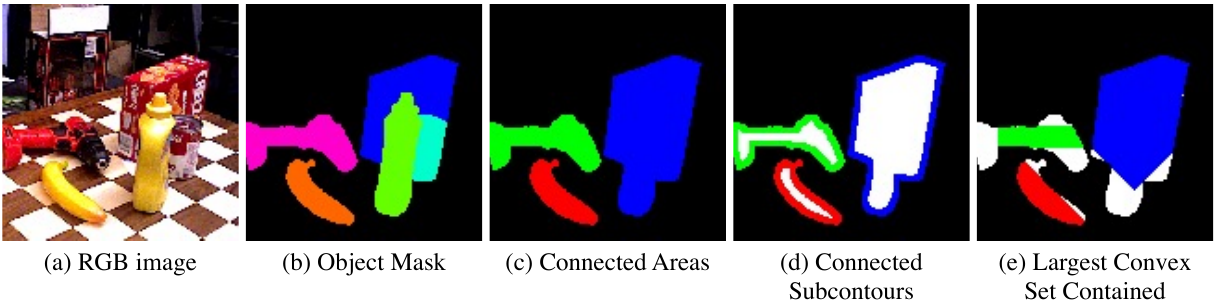}
\caption{\hl{Illustrations for Background Shape Irregularity.}}
\label{fig:main_background_shape_irregularity}
\vspace{-0.3cm}
\end{figure}

\begin{figure*}[t]
\setlength{\abovecaptionskip}{ 2 pt}
\setlength{\belowcaptionskip}{ -8 pt}
\centering
   \includegraphics[width=1.\linewidth]{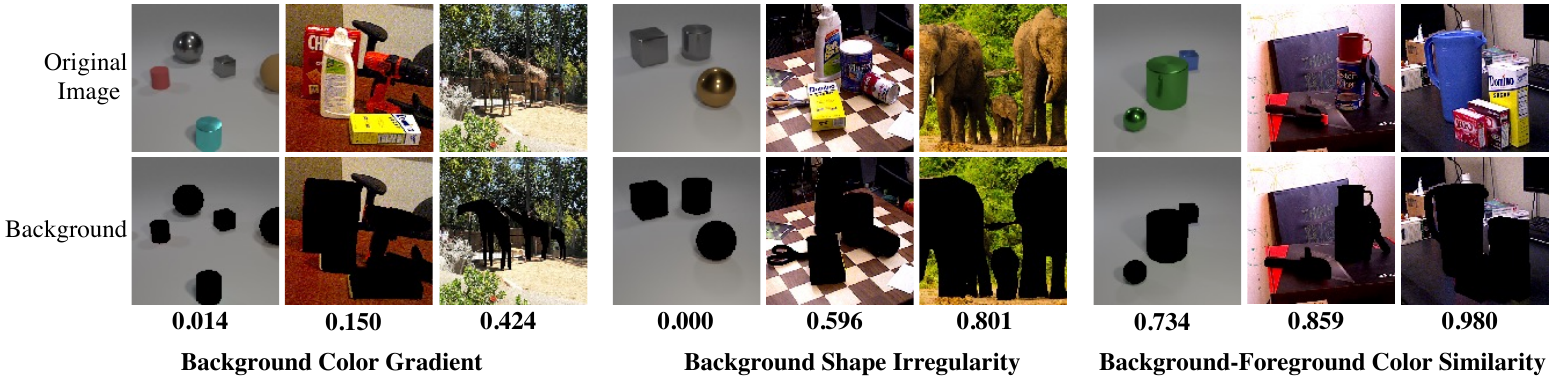}
\caption{\hl{Sample image backgrounds for the three factors at different complexity values.}}
\label{fig:main_bg_factors_illustration}
\vspace{-0.3cm}
\end{figure*}

\textbf{Background Shape Irregularity:} This factor aims to measure the irregularity of the background shape, \ie{}, the full contour where the background intersects with all foreground objects. As illustrated in Figure \ref{fig:main_background_shape_irregularity}, given the (binary) background mask, we first divide its contour into subcontours, where each subcontour is fully self-connected and encloses a region. \hl{In fact, that region may originally contain one or more foreground objects, which means that the previously designed Inter-object Shape Variation factor for foreground objects cannot be simply reused to measure the shape complexity of background.
For each enclosed region, we first compute its maximal inscribed convex set based on an existing algorithm \mbox{\citep{borgefors2005approximation}} whose details are provided in the Appendix. The area of each enclosed region is denoted as $A_{i}$ and the area of its inscribed convex set is denoted as $C_{i}$. The irregularity of each subcontour is then calculated as $1- \frac{C_{i}}{A_{i}}$. The final Background Shape Irregularity score for each image is calculated as the average of irregularity score for all subcontours within the image as: $\frac{1}{N}\sum_{i=1}^{N} (1-\frac{C_{i}}{A_{i}}) $, where $N$ is the total number of enclosed regions within an image. }
Clearly, the higher this factor, the more irregular the background shape (contour), and therefore the harder to separate background pixels from foreground objects. 

Above all, the three factors together are designed to quantify the complexity of image backgrounds in terms of  appearance and shape. Figure \ref{fig:main_bg_factors_illustration} shows sample images for three background complexity factors at different values. The original full images are also included for reference.

\section{Experimental Design}\label{sec:exep_design}
\subsection{Considered Methods}

A range of works have explored unsupervised object segmentation in recent years. They are typically formulated as (variational) autoencoders (AE/VAE) \citep{Kingma2014} or generative adversarial networks (GAN) \citep{Goodfellow2014}. GAN based models \citep{Chen2019,Arandjelovic2019,Bielski2019,VanSteenkiste2020,Azadi2020,Voynov2021,Abdal2021} are usually limited to identifying a single foreground object and can hardly discover multiple objects due to the training instabilities, therefore not considered in this paper. There is another line of works that adopt Energy-based Models (EBMs) \citep{lecun2006tutorial,du2021unsupervised,Liu2022} on scene decomposition. However, they do not explicitly generate object masks but encode objects into energy functions, therefore not discussed in this paper.
As shown in Table \ref{tab:existing_models}, the majority of existing models are based on AE/VAE and can be generally divided into two groups according to the object representation: 

\begin{itemize}[leftmargin=*]
\vspace{-0.1cm}
\setlength{\itemsep}{1pt}
\setlength{\parsep}{1pt}
\setlength{\parskip}{1pt}
\item \textbf{Factor-based models}: Each object is represented by explicit factors such as size, position, appearance, \etc{}, and the whole image is a spatial organization of multiple objects. Basically, such representation explicitly enforces objects to be bounded within particular regions. 
\item \textbf{Layer-based models}: Each object is represented by an image layer, \ie{}, a binary mask, and the whole image is a spatial mixture of multiple object layers. Intuitively, this representation does not have strict spatial constraints, and instead is more flexible to cluster similar pixels as objects. 
\end{itemize}

\begin{table*}[th] 
\vspace{0.2cm}
\centering
\setlength{\abovecaptionskip}{ 4 pt}
\setlength{\belowcaptionskip}{ -10 pt}
\caption{Existing unsupervised models for object segmentation on single images. Each model includes different inductive biases, such as variational autoencoding (VAE), iterative inference (Iter), object relationship regularization (Rel), \etc{}.}
\label{tab:existing_models}
\resizebox{\textwidth}{!}
{
\begin{tabular}{rlccc|rlccc}
\toprule[1.0pt]
\multicolumn{2}{c}{\multirow{2}{*}{Factor-based Models}} & \multicolumn{3}{c|}{Inductive Biases} & 
\multicolumn{2}{c}{\multirow{2}{*}{Layer-based Models}} & \multicolumn{3}{c}{Inductive Biases} \\
\multicolumn{2}{c}{} & VAE & Iter & Rel & \multicolumn{2}{c}{} & VAE & Iter & Rel \\
\toprule[1.0pt]
CST-VAE \citep{Huang2016} & ICLRW'16 &\checkmark& & & Tagger \citep{Greff2016} & NIPS'16  & & \checkmark & \\ 
AIR \citep{Eslami2016}    & NIPS'16  &\checkmark& & & RC \citep{Greff2016} & ICLRW'16 & & \checkmark & \\  
SPAIR \citep{Crawford2019} & AAAI'19 &\checkmark& & & NEM \citep{Greff2017} & NIPS'17 & & \checkmark & \\
SuPAIR \citep{Stelzner2019} & ICML'19 &\checkmark& & & MONet \citep{Burgess2019} & arXiv'19 & \checkmark & & \\  
GMIO  \citep{Yuan2019} & ICML'19 &\checkmark&\checkmark& & IODINE \citep{Greff2019} & ICML'19 &\checkmark&\checkmark& \\ 
ASR  \citep{Xu2019} & NeurIPS'19 &\checkmark& & &ECON \citep{VonKugelgen2020} &ICLRW'20 &\checkmark & & \checkmark \\ 
SPACE \citep{Lin2020} & ICLR'20 &\checkmark& & & GENESIS \citep{Engelcke2020} &ICLR'20 & \checkmark & & \checkmark \\ 
GNM \citep{Jiang2020} & NeurIPS'20 &\checkmark& &\checkmark & SlotAtt \citep{Locatello2020} & NeurIPS'20 & & \checkmark & \\ 
SPLIT \citep{Charakorn2020} & arXiv'20 &\checkmark& & &GENESIS-V2 \citep{Engelcke2021} & NeurIPS'21 & \checkmark & &\checkmark \\
OCIC \citep{Anciukevicius2020} &arXiv'20 &\checkmark&&\checkmark & CAE \citep{Lowe2022} & TMLR'22 & & & \checkmark  \\
GSGN \citep{Deng2021} & ICLR'21 & \checkmark & & \checkmark & BO-QSA \citep{Jia2023}&ICLR'23 & & \checkmark & \\
ISA \citep{biza2023invariant} & ICML'23 & \checkmark & \checkmark &  & ISA \citep{biza2023invariant}& ICML'23 & \checkmark & \checkmark & \\
\bottomrule[1.0pt]
\end{tabular}
}
\end{table*}

To decompose input images into objects against backgrounds, these approaches introduce different types of network architecture, losses, and regularization terms as inductive biases. These biases broadly include: 1) variational encoding which encourages the disentanglement of latent variables; 2) iterative inference which likely ends up with better scene representations over occlusions; 3) object relationship regularization such as depth estimation and autoregressive prior which aims at capturing the dependency of multiple objects; and many other biases. With different combinations of these biases, many methods have shown outstanding performance on synthetic datasets. Among them, we select 4 representative models for investigation: 1) AIR \citep{Eslami2016}, 2) MONet \citep{Burgess2019}, 3) IODINE \citep{Greff2019}, and 4) SlotAtt \citep{Locatello2020}. We also add the fully-supervised Mask R-CNN \citep{He2017a} as an additional baseline for a comprehensive comparison in Sections \ref{sec:exp_main_bg} $\sim$ \ref{sec:exp_why_fail}.

\hl{More recently, a number of new methods such as Odin \mbox{\citep{Henaff2022}}, DINOSAUR \mbox{\citep{seitzer2022bridging}}, FreeSOLO \mbox{\citep{wang2022freesolo}}, and CutLER \mbox{\citep{wang2023cut}} make use of pretrained models on monolithic object images such as ImageNet to segment objects from real-world images. We additionally evaluate and discuss the representative method DINOSAUR \mbox{\citep{seitzer2022bridging}} in Section \mbox{\ref{sec:pretrained_model_discussion}}}.


\vspace{-0.3cm}
\subsection{Considered Datasets}\label{sec:considered_datasets}

We consider four groups of datasets for extensive benchmarking and analysis: 

\begin{itemize}[leftmargin=*]
\vspace{-0.3cm}
\setlength{\itemsep}{1pt}
\setlength{\parsep}{1pt}
\setlength{\parskip}{1pt}
\item \textbf{Group 1:} Three commonly-used synthetic datasets with blank backgrounds: dSprites \citep{Matthey2017}, Tetris \citep{Kabra2019} and CLEVR \citep{Johnson2017};
\item \textbf{Group 2:} One semi-realistic and three real-world datasets with blank backgrounds: \hl{MOViC}\citep{greff2021kubric},  YCB \citep{Calli2017}, ScanNet \citep{Dai2017}, and COCO \citep{Lin2014}, representing \hl{semi-realistic,} small-scale, indoor- and outdoor-level real scenes respectively; 
\item \textbf{Group 3:} One synthetic dataset with synthetic backgrounds: CLEVR$_{bg}$ \citep{Johnson2017}, whose foreground objects are the same as CLEVR in Group 1;
\item \textbf{Group 4:} One semi-realistic and three real-world datasets with backgrounds: \hl{MOViC$_{bg}$} \citep{greff2021kubric}, YCB$_{bg}$ \citep{Calli2017}, ScanNet$_{bg}$ \citep{Dai2017}, and COCO$_{bg}$ \citep{Lin2014}, whose foreground objects are the same as the three datasets in Group 2. 
\end{itemize}

Naturally, objects and scenes in different datasets tend to have very different types of biases. For example, the objects in dSprites tend to have a single-color bias, while COCO does not; the objects in ScanNet$_{bg}$ tend to be influenced by cluttered backgrounds, while ScanNet does not. Generally, the object-level biases can be divided as: 1) appearance biases including different textures and lighting effects for objects, and 2) geometry biases including the object shape and occlusions. Similarly, the scene-level biases include: 1) appearance biases such as the color similarity between all objects, and 2) geometry biases such as the diversity of all object shapes. The background biases can be divided as: 1) appearance biases including textures and lighting effects of backgrounds, and the color similarity between background and foreground, 2) geometry biases such as the irregularity of background shape. In fact, our complexity factors introduced in Section \ref{sec:complexity_factors} are designed to well capture these biases. Table \ref{tab:dataset_biases} \hl{in the appendix} qualitatively summarizes the biases of datasets in Groups 1/2/3/4. We may hypothesize that the large gaps of biases between synthetic and real-world datasets would have a huge impact on the effectiveness of existing models. 

To guarantee the fairness and consistency of all experiments, we carefully prepare the four groups of datasets using the following same protocols. Preparation details of each dataset are provided in the appendix.
\vspace{-0.15cm}
\begin{itemize}[leftmargin=*]
\setlength{\itemsep}{1pt}
\setlength{\parsep}{1pt}
\setlength{\parskip}{1pt}
    \item All images are rerendered or cropped with the same resolution of $128\times 128$. 
    \item Each image has about 2 to 6 solid objects, with a blank background in Groups 1\&2, with a synthetic background in Group 3, and with a (semi)real background in Group 4. 
    \item Each dataset has about 10000 images for training, 2000 images for testing.
\end{itemize}

The datasets in Groups 1\&2 are primarily used to evaluate how the four object- and scene-level complexity factors affect object segmentation performance of existing methods, while the datasets in Groups 3\&4 are used to investigate how the diversity of backgrounds affects the segmentation performance.

\begin{figure*}[th]
    \setlength{\abovecaptionskip}{ 1 pt}
    \setlength{\belowcaptionskip}{ 2 pt}
    \centering
       \includegraphics[width=1\linewidth]{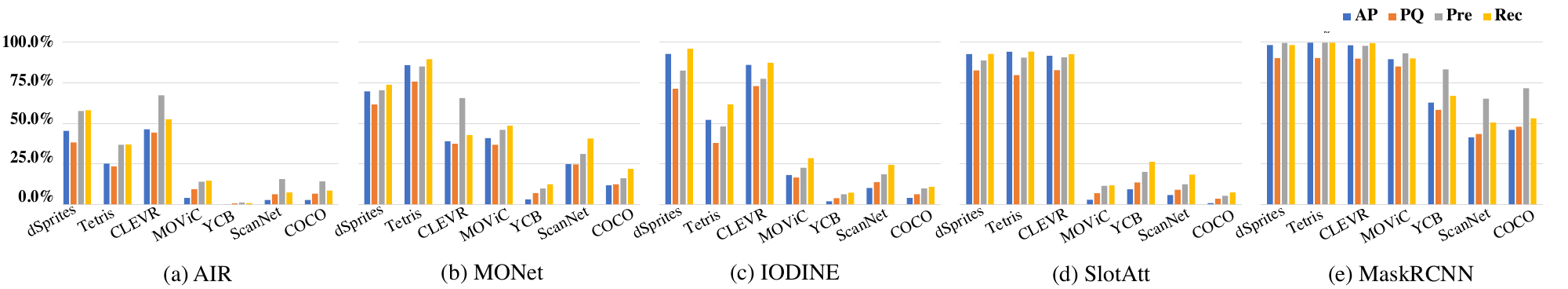}
    \caption{\hl{Quantitative results of object segmentation on 7 datasets from Groups 1\&2 with blank backgrounds.}}
    \label{fig:main_exp}
    \vspace{-0.2cm}
\end{figure*}
\begin{figure*}[t]
    \setlength{\abovecaptionskip}{ 1 pt}
    \setlength{\belowcaptionskip}{ -6 pt}
    \centering
       \includegraphics[width=1\linewidth]{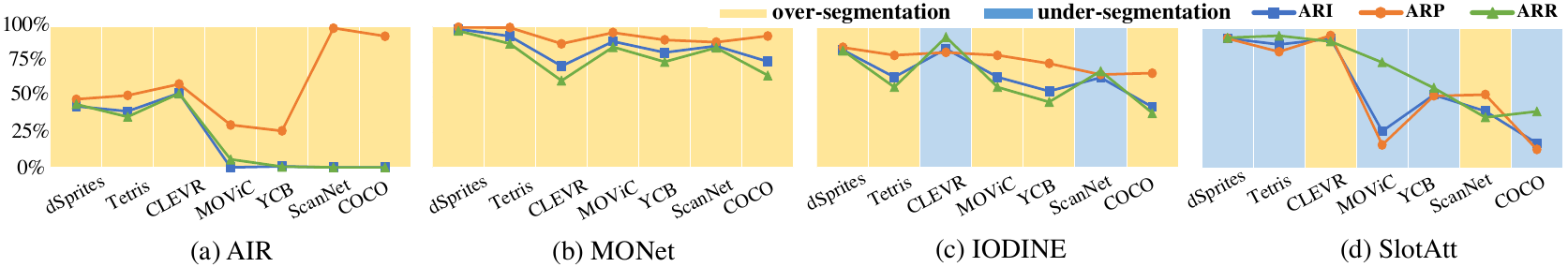}
    \caption{\hl{Quantitative results for comparing over/under-segmentation of 4 unsupervised methods on 7 datasets of Groups 1\&2.}}
    \label{fig:main_arp_arr}
\end{figure*}

\begin{figure*}[t]
    \setlength{\abovecaptionskip}{ 1 pt}
    \setlength{\belowcaptionskip}{ -6 pt}
    \centering
       \includegraphics[width=1\linewidth]{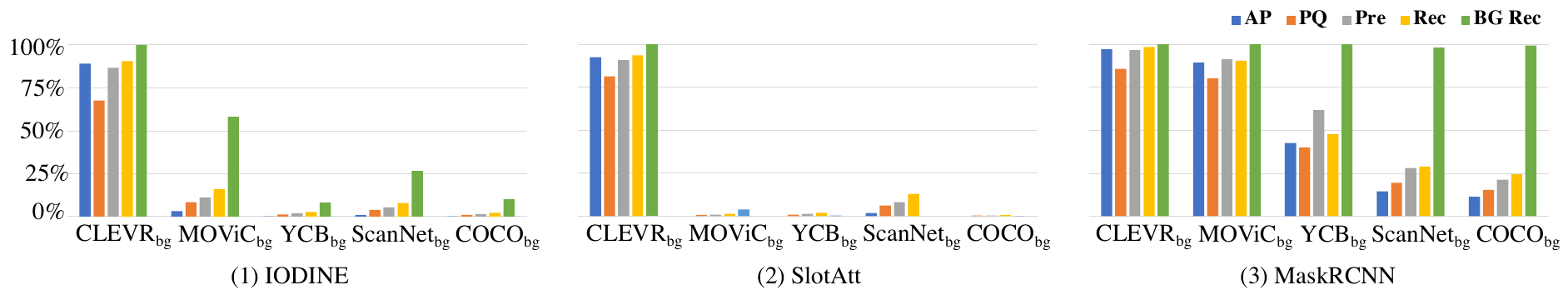}
    \caption{\hl{Quantitative results of object segmentation on 5 datasets of Groups 3\&4 with synthetic or real-world backgrounds.}}
    \label{fig:main_exp_bg}
\end{figure*}

\subsection{Considered Metrics}
Having the \hl{12} representative datasets and existing unsupervised methods at hand, we choose the following metrics to evaluate the object segmentation performance: 1) AP score which is widely used for object detection and segmentation \citep{Everingham2015}, 2) PQ score which is used to measure non-overlap panoptic segmentation \citep{Kirillov2019}, and 3) Precision and Recall scores. A predicted mask is considered correct if its IoU against a ground truth mask is above 0.5. All objects are treated as a single class. The blank background is not taken into account for foreground object segmentation. To compute AP, we simply treat the mean value of the soft object mask as the object confidence score. 
\hl{In order to quantitatively analyze over-/under-segmentation issues, we additionally calculate ARP/ARR scores \mbox{\citep{zimmermann2023sensitivity}} and ARI score \mbox{\citep{Rand1971}}}.
Note that, the alternative segmentation covering (SC) \citep{Arbelaez2011} is not considered as it can be easily saturated.

\section{Key Experimental Results}\label{sec:experimental_res}

\subsection{Can current unsupervised models succeed on real-world datasets?}\label{sec:exp_main_bg}

\begin{figure}[ht]
\vspace{-0.5cm}
    \setlength{\abovecaptionskip}{ 2 pt}
    \setlength{\belowcaptionskip}{ -4 pt}
     \centering
\includegraphics[width=1\linewidth]{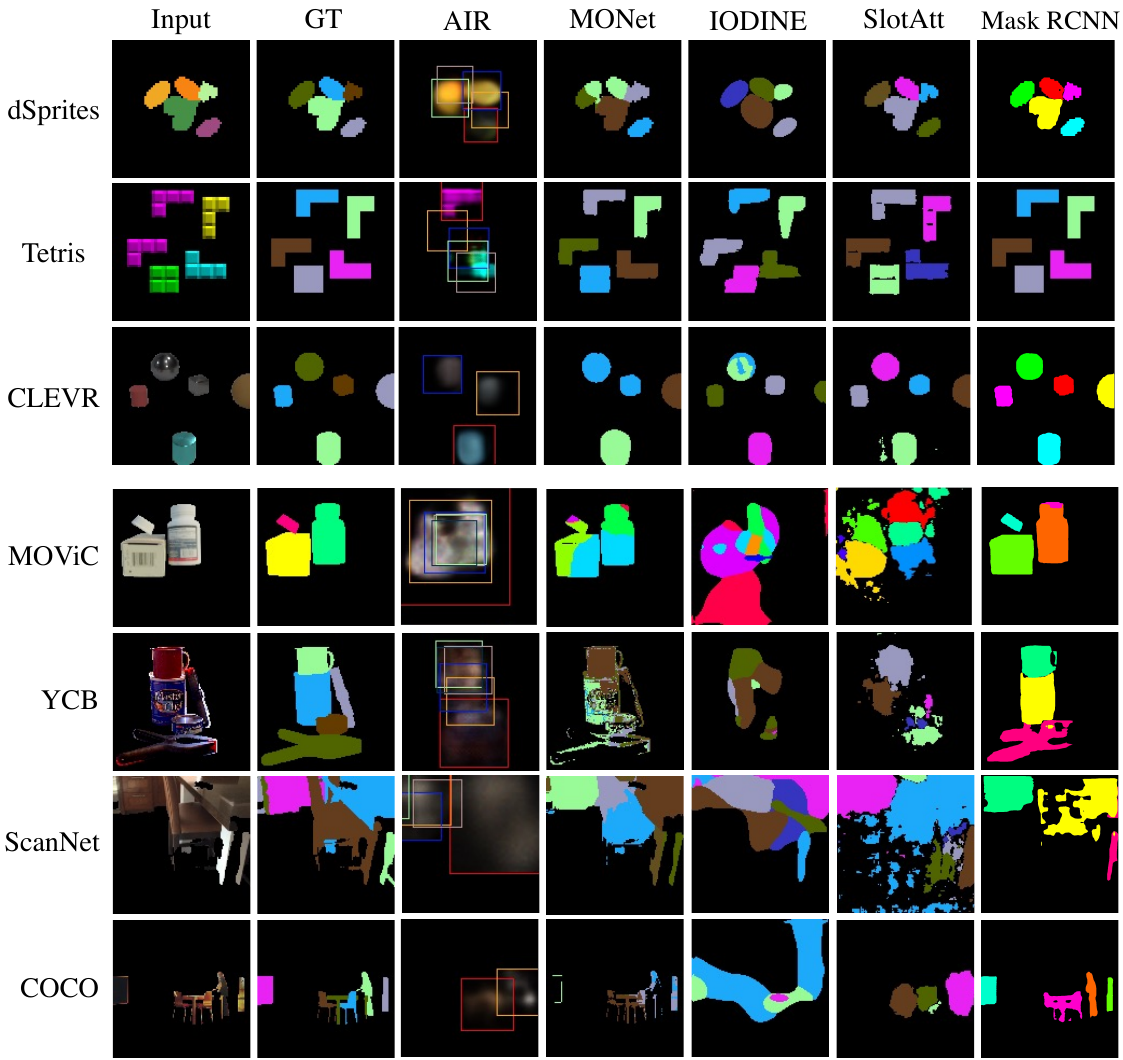}
\caption{\hl{Qualitative results of object segmentation from five methods on datasets of Groups 1\&2 with blank backgrounds.}}
\label{fig:main_exp_vis}
\vspace{-0.6cm}
\end{figure}

\hl{We first evaluate the performance of all five baselines on synthetic datasets (dSprites, Tetris, CLEVR) in Group 1, as well as on the datasets MOViC, YCB, ScanNet, and COCO in Group 2. It is important to note that all images in these datasets only have blank backgrounds. Both quantitative  and qualitative results are presented in Figures \mbox{\ref{fig:main_exp}\&\ref{fig:main_exp_vis}}. Detailed breakdown ARP/ARR scores for comparing over-/under- segmentation are provided in Figure \mbox{\ref{fig:main_arp_arr}}.} 

\hl{Notably, on synthetic datasets, all methods, especially the recent baselines IODINE and SlotAtt, demonstrate satisfactory segmentation outcomes. However, it is not surprising that all unsupervised methods encounter significant challenges when applied to real-world datasets, despite the blank backgrounds observed in all images. In addition, from ARP/ARR scores, the methods AIR, MONet and IODINE are more prone to over-segmentation, while SlotAtt tends to exhibit under-segmentation across all 7 datasets. Such an observation is aligned with results in \mbox{\citep{zimmermann2023sensitivity}}.}

\begin{figure}[ht]
    \setlength{\abovecaptionskip}{ 2 pt}
    \setlength{\belowcaptionskip}{ 0 pt}
     \centering
\includegraphics[width=1\linewidth]{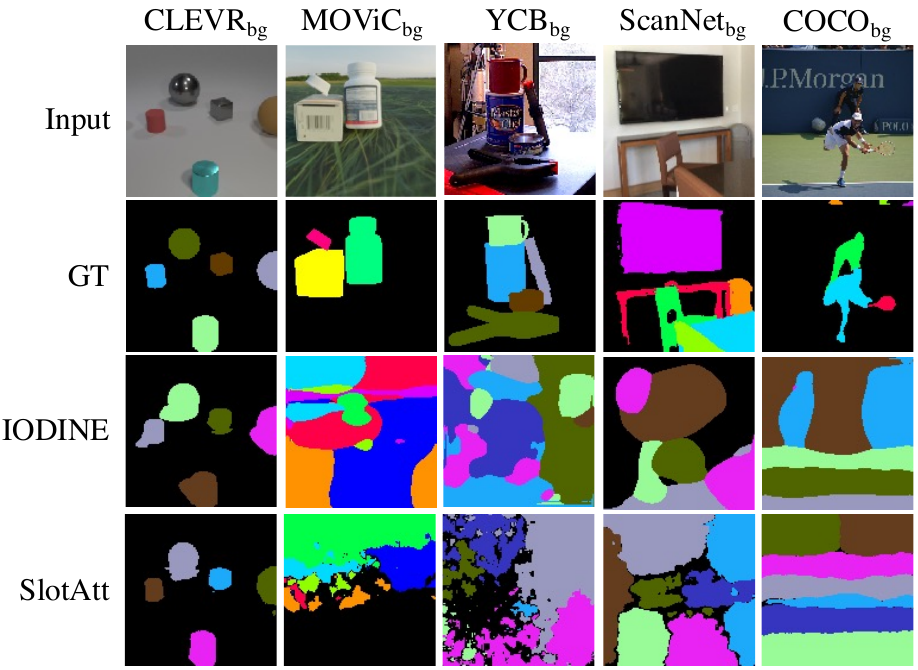}
\caption{\hl{Qualitative results of object segmentation on CLEVR$_{bg}$/ MOViC$_{bg}$/ YCB$_{bg}$/ ScanNet$_{bg}$/ COCO$_{bg}$.}}
\label{fig:main_exp_bg_vis}
\vspace{-0.4cm}
\end{figure}

We then evaluate three baselines (IODINE/SlotAtt/ MaskRCNN, because of their better abilities to model backgrounds) on datasets (CLEVR$_{bg}$/\hl{MOViC$_{bg}$}/YCB$_{bg}$/ScanNet$_{bg}$/COCO$_{bg}$) from Groups 3\&4, where all images have either synthetic or real-world backgrounds. Figures \ref{fig:main_exp_bg}\&\ref{fig:main_exp_bg_vis} show the quantitative/qualitative results respectively. We can see that both unsupervised IODINE\&SlotAtt achieve excellent performance on the synthetic dataset CLEVR$_{bg}$, but again fail on real-world datasets with real backgrounds. 

\begin{figure*}[th]
    \setlength{\abovecaptionskip}{ 2 pt}
    \setlength{\belowcaptionskip}{ -6 pt}
    \centering
    \includegraphics[width=0.95\linewidth]{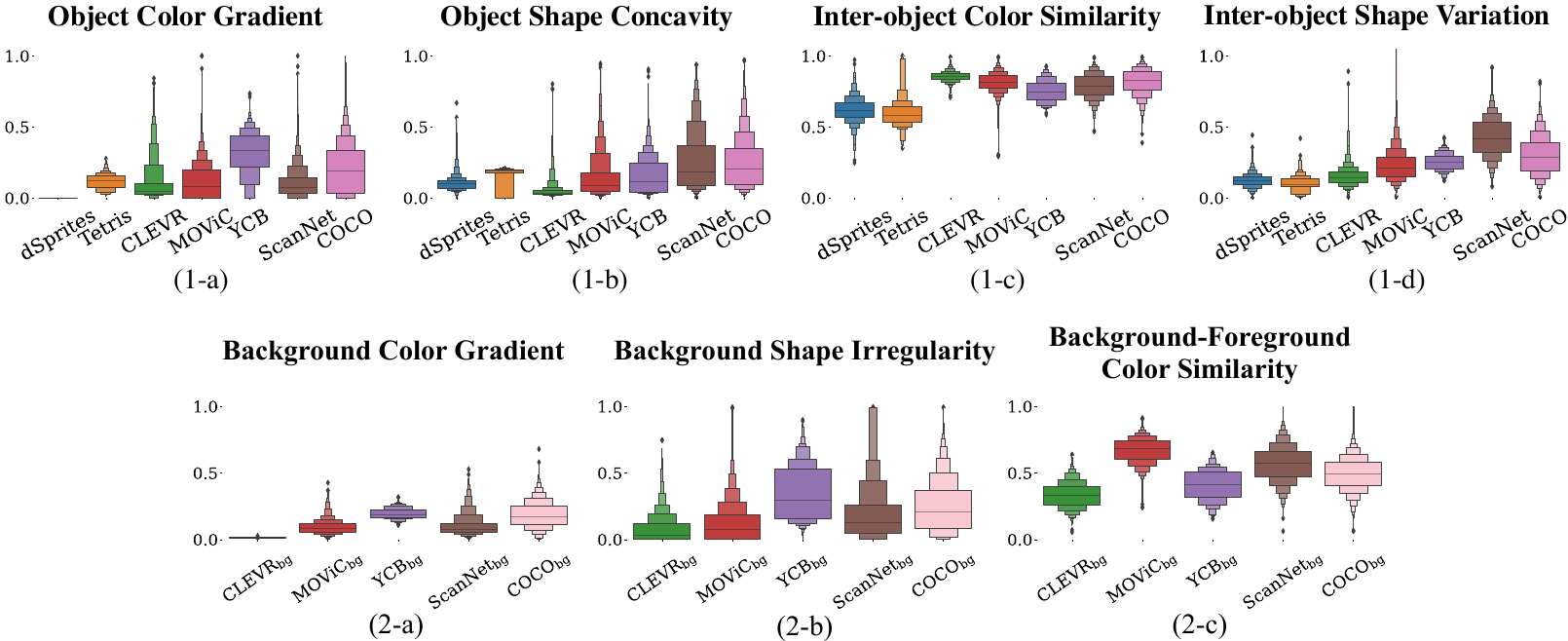}
        \caption{\hl{Distributions of the 7 complexity factors on four groups of datasets.}}
    \label{fig:main_exp_original_factors}
    \vspace{-0.1cm}
\end{figure*}

\textbf{Preliminary Diagnosis:} To diagnose the failure on real-world images, we hypothesize that it is because of the gaps in biases between synthetic and real-world datasets. In this regard, we quantitatively compute the distributions of all 7 complexity factors on the four groups of datasets. In particular, the two object-level factors, \ie{}, Object Color Gradient and Object Shape Concavity, and the two scene-level factors, \ie{}, Inter-object Color Similarity and Inter-object Shape Variation, are computed on the \hl{seven} training splits of Groups 1\&2 datasets (dSprites/Tetris/CLEVR, \hl{MOViC}/YCB/ScanNet/COCO). The three background-level factors, \ie{}, Background Color Gradient, Background-Foreground Color Similarity and Background Shape Irregularity, are computed for each image of the four training splits of Groups 3\&4 datasets (CLEVR$_{bg}$, MOViC$_{bg}$, YCB$_{bg}$/ ScanNet$_{bg}$/COCO$_{bg}$).

As shown in the \textit{top row} of Figure \ref{fig:main_exp_original_factors}, we can see that: 
1) for the two object-level factors \hl{(Subfigs 1-a and 1-b)}, \hl{the complexity scores for synthetic datasets are substantially lower than those of real-world datasets, and the semi-realistic MOViC is in between. This implies that synthetic objects are more likely to have uniform colors and convex shapes; }
 2) for the two scene-level factors \hl{(Subfigs 1-c and 1-d)}, the images in synthetic datasets tend to include less similar objects in terms of color, which means that multiple objects in real-world scenes are less distinctive in appearance. In addition, the multiple objects in synthetic scenes tend to have similar sizes, whereas real-world scenes usually have diverse object scales in single images. As shown in the \textit{bottom row} of Figure \ref{fig:main_exp_original_factors}, we can see that: 1) real-world image backgrounds are more likely to have non-uniform colors and irregular shapes \hl{(Subfigs 2-a and 2-b)}; 2) synthetic images tend to have more distinctive backgrounds against foreground objects than real-world images \hl{(Subfig 2-c)}. 

To validate whether these distribution biases are the true reasons incurring the failure, we conduct extensive ablative experiments in Sections  \ref{sec:exp_object_factors}, \ref{sec:exp_scene_factors}, \ref{sec:exp_joint_factors}, and \ref{sec:exp_background_factor}.

\begin{figure*}[ht]
    \setlength{\abovecaptionskip}{ 4 pt}
    \setlength{\belowcaptionskip}{ -2 pt}
    \centering
       \includegraphics[width=0.97\linewidth]{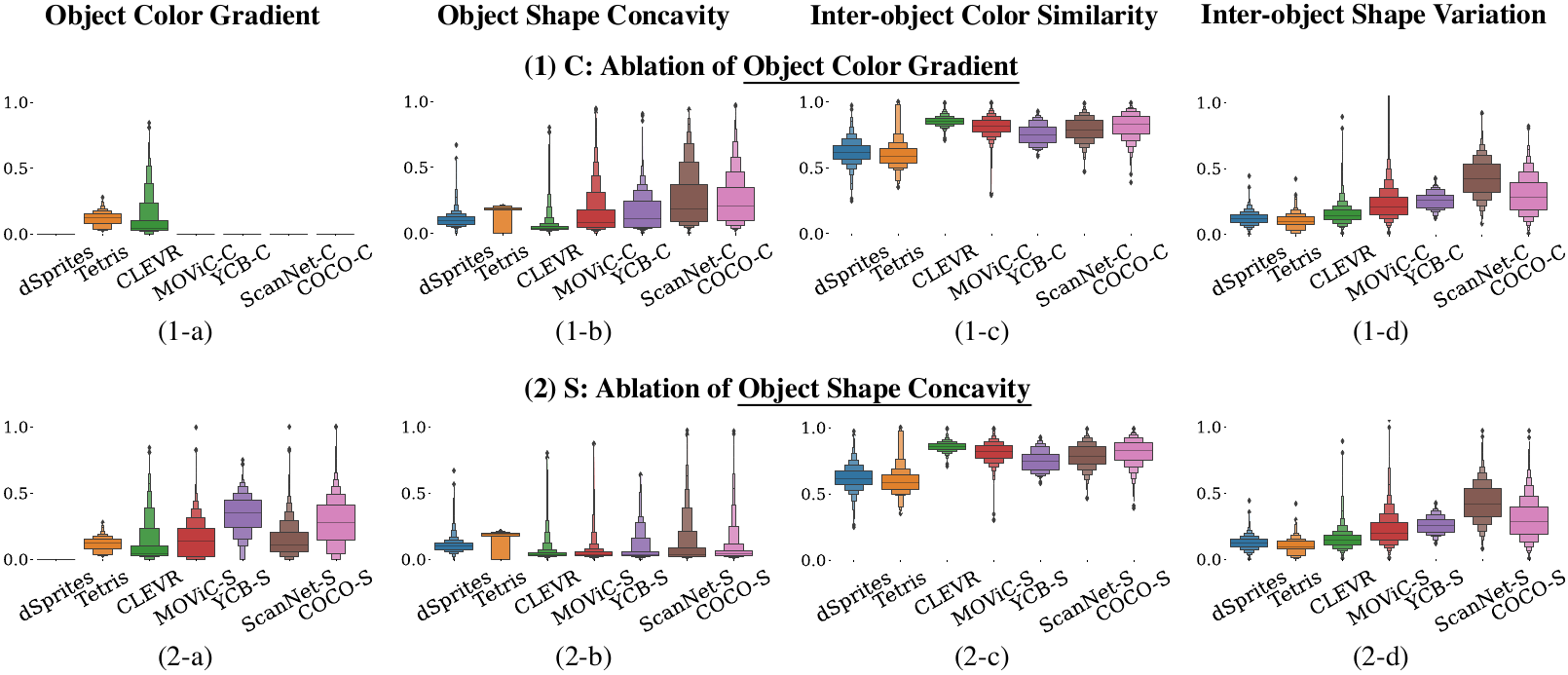}
    \caption{\hl{Distributions of four foreground factors on datasets ablated in object-level.}}
    \label{fig:main_object_level_ablation_factors}
\end{figure*}

\begin{figure*}
    \setlength{\abovecaptionskip}{ 4 pt}
    \setlength{\belowcaptionskip}{ -2 pt}
\centering
\includegraphics[width=0.97\textwidth]{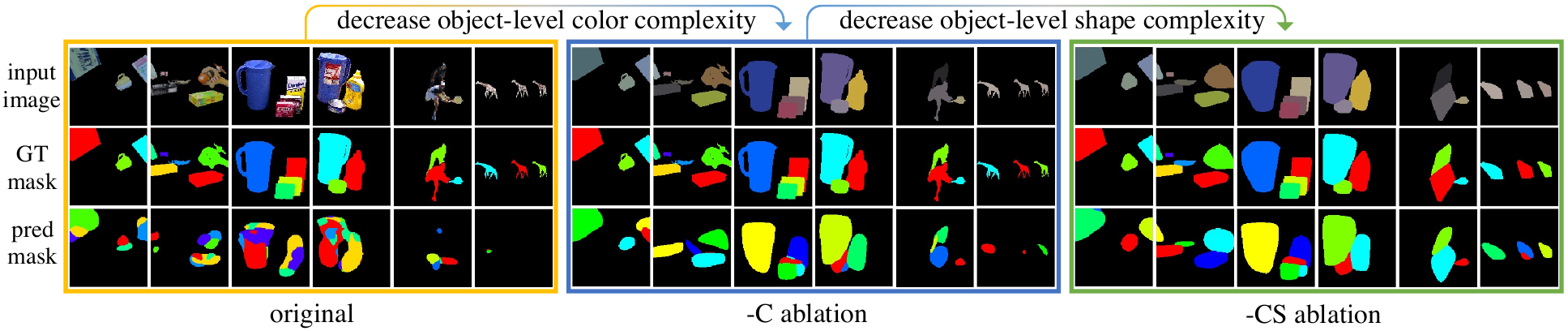}
\caption{\hl{Qualitative results of object-level factors ablations from IODINE.}}
\vspace{-0.3cm}
\label{fig:segmentation_by_object_complexity}
\end{figure*}

\begin{figure*}[t]
    \setlength{\abovecaptionskip}{ 4 pt}
    \setlength{\belowcaptionskip}{ -8 pt}
    \centering
       \includegraphics[width=0.97\linewidth]{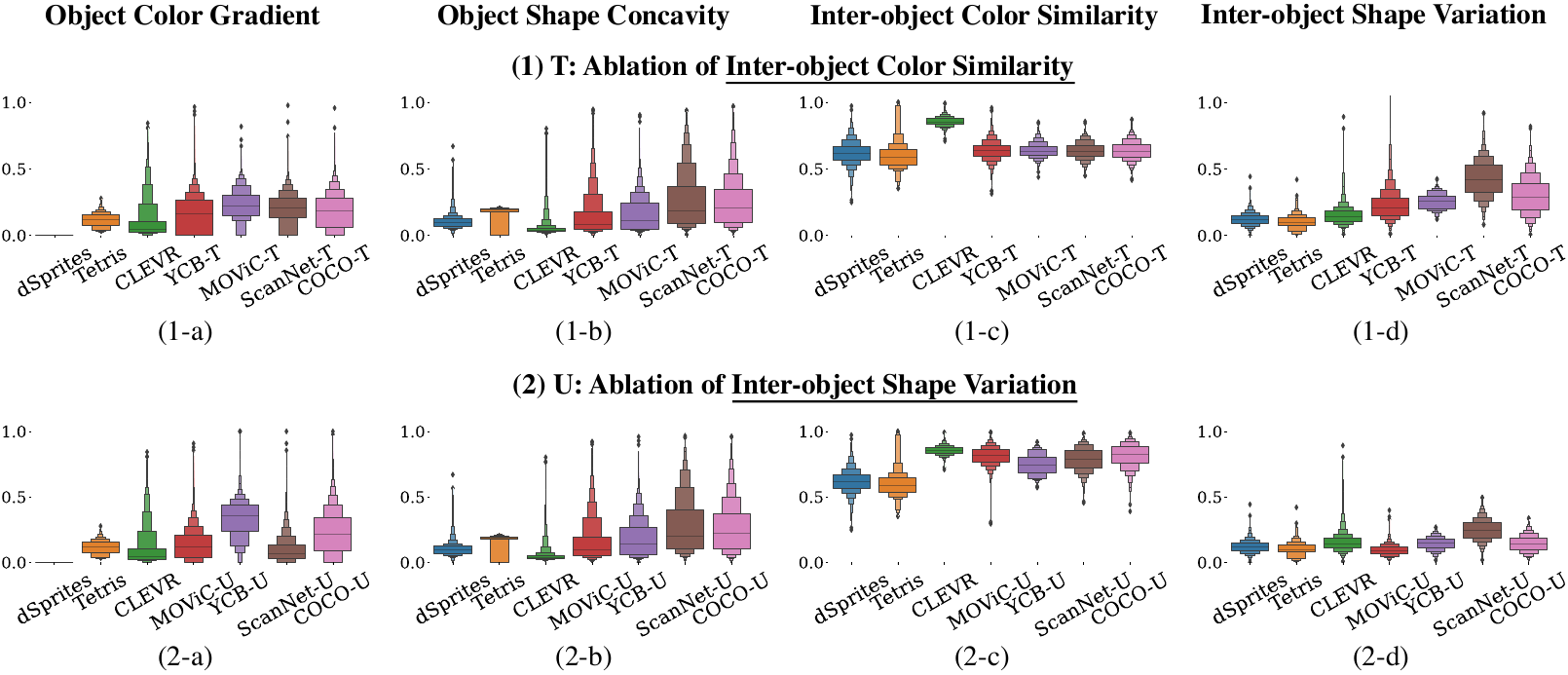}
    \caption{\hl{Distributions of four foreground factors on datasets ablated in scene-level.}}
    \label{fig:scene_level_ablation_factors}
\end{figure*}%

\begin{figure*}
  \centering
  \begin{tabular}{@{}c@{}}
    \includegraphics[width=1\linewidth]{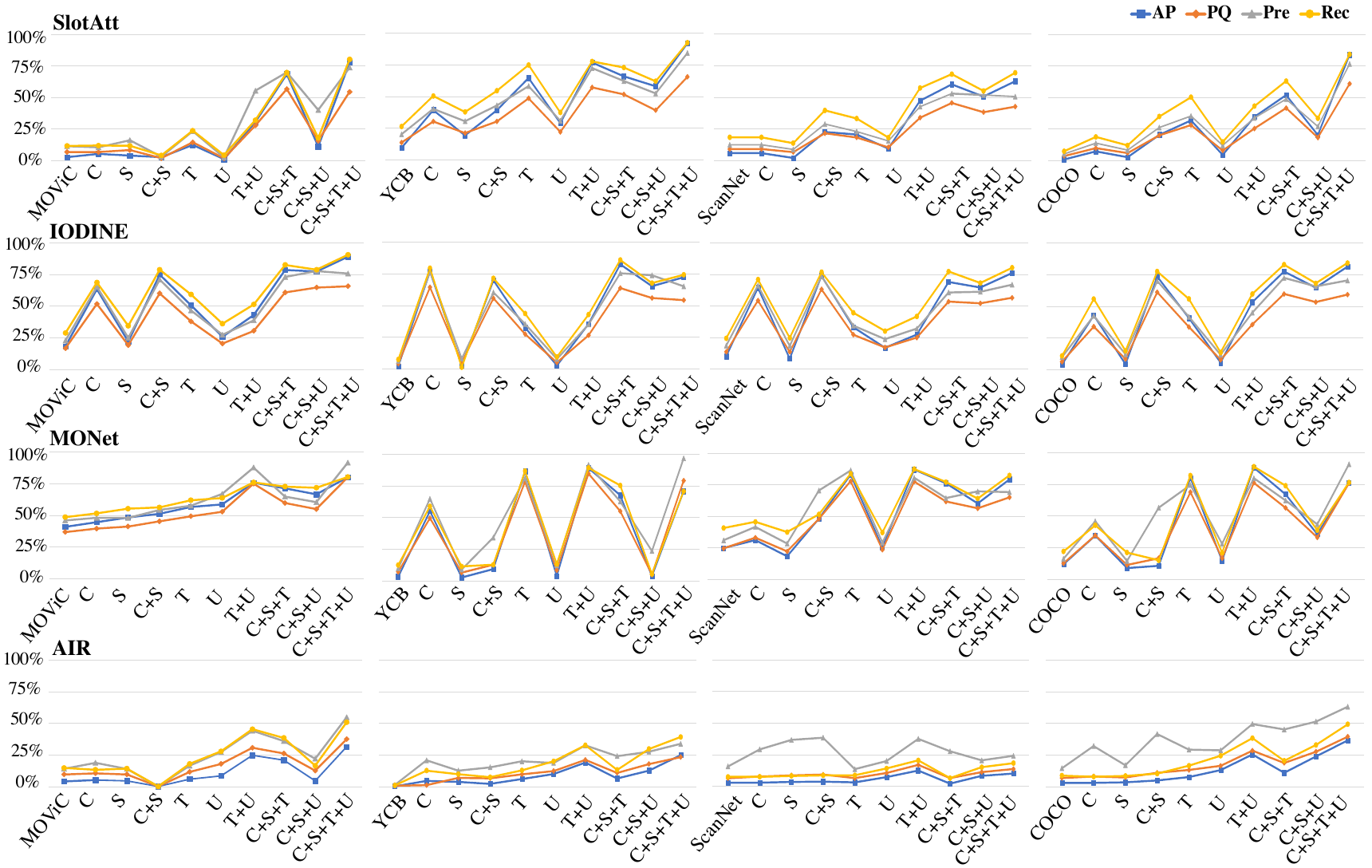} \\
    \small \hl{(a) Quantitative results of baselines on the group 2 datasets (with blank backgrounds) and their variants.}
  \end{tabular}
  \begin{tabular}{@{}c@{}}
    \includegraphics[width=1\linewidth]{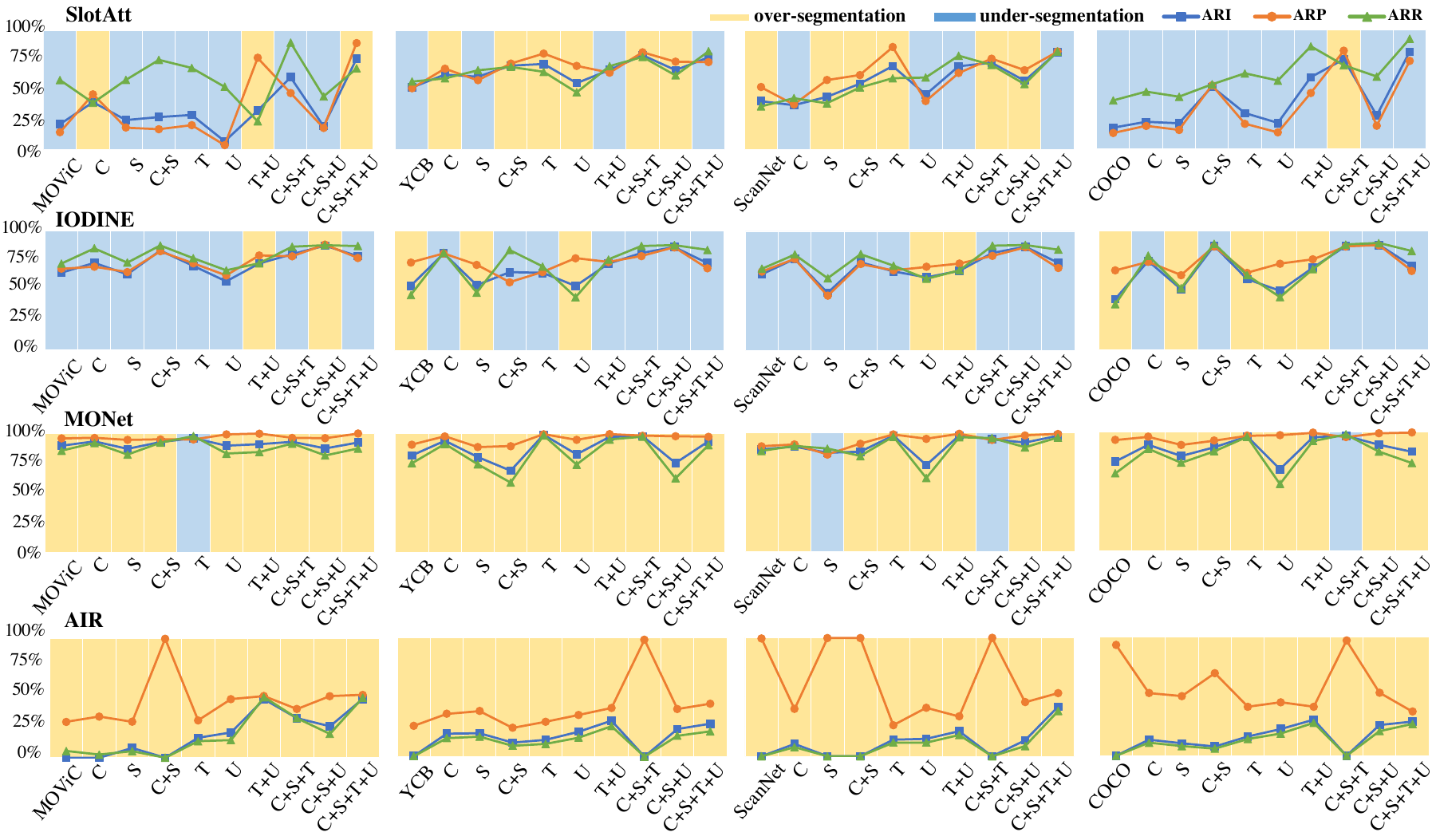} \\
    \small (b) \hl{The ARP/ARR/ARI scores of baselines on the group 2 datasets for analyzing over-/under-segmentation.}
  \end{tabular}
  \caption{\hl{The letters C/S/C+S represent the three ablated datasets in Section {\ref{sec:exp_object_factors}}; T/U/T+U represent the three ablated datasets in Section {\ref{sec:exp_scene_factors}}; C+S+T/C+S+U/C+S+T+U represent the three ablated datasets in Section {\ref{sec:exp_joint_factors}}.}}
  \label{fig:main_obj_ablation_summary}
\end{figure*}

\begin{figure*}
    \setlength{\abovecaptionskip}{ 4 pt}
    \setlength{\belowcaptionskip}{ -8 pt}
\centering
\includegraphics[width=0.96\textwidth]{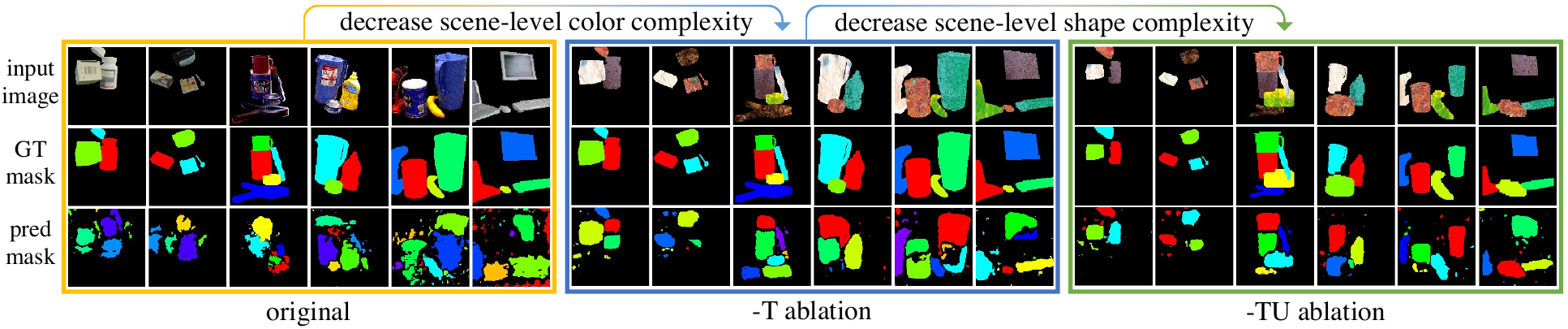}
\caption{\hl{Qualitative results of scene-level factors ablations from SlotAtt.}}
\label{fig:segmentation_by_scene_complexity}
\end{figure*}

\begin{figure*}
    \setlength{\abovecaptionskip}{ 4 pt}
    \setlength{\belowcaptionskip}{ -8 pt}
\centering
\includegraphics[width=0.96\textwidth]{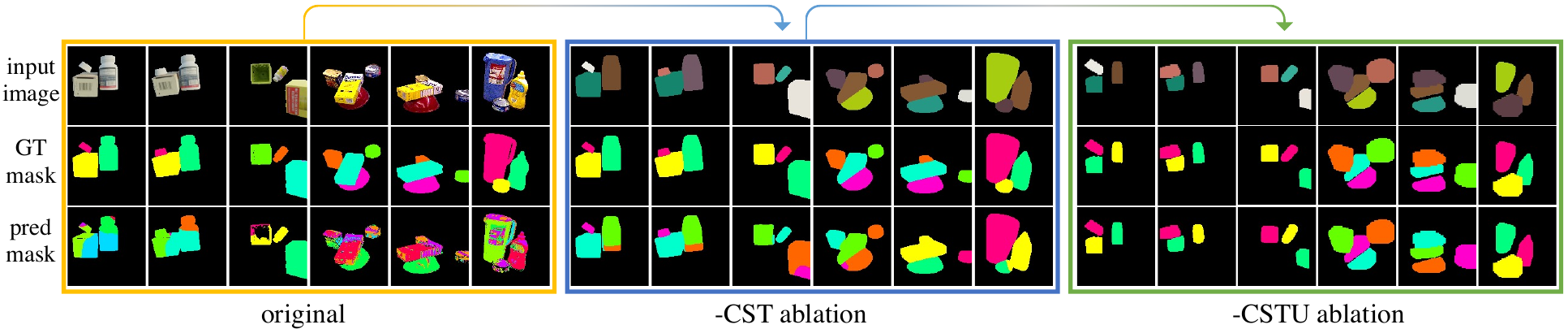}
\caption{\hl{Qualitative results of both object-level and scene-level factors ablations from MONet.}}
\label{fig:segmentation_by_object_and_scene_complexity}
\end{figure*}

\subsection{How do object-level factors affect current models?}\label{sec:exp_object_factors}
\vspace{-0.15cm}
In this section, we aim to verify to what extent the distributions of object-level factors affect the segmentation performance by doing the following three ablative experiments. 
\hl{Examples of object-level ablations can be found in the appendix Figure} \ref{fig:ablation_example_1}.

\vspace{-0.15cm}
\begin{itemize}[leftmargin=*]
\setlength{\itemsep}{1pt}
\setlength{\parsep}{1pt}
\setlength{\parskip}{1pt}
    \item \textit{Ablation of Object Color Gradient}: For each object of Group 2 datasets (\hl{MOViC}/YCB/ScanNet/COCO), we only replace all pixel colors by its average color $rgb$ value within each object mask, without touching the object shapes. In this way, the \textbf{c}olor gradients of each object are totally erased, thus removing the potential impact of Object Color Gradient. The three ablated datasets are: \hl{MOViC-C}/YCB-C/ScanNet-C/COCO-C. 
    \item \textit{Ablation of Object Shape Concavity}: For each object in Group 2 datasets (\hl{MOViC}/YCB/ScanNet/COCO), we find the smallest convex hull \citep{Eddins2011} for its object mask and then fill the empty pixels by shifting original object pixels. Basically, this ablation aims to only reduce the irregularity of object {\textbf{s}}hapes, yet retain the distributions of color gradients. The ablated datasets are: \hl{MOViC-S}/YCB-S/ScanNet-S/COCO-S.  
    \item \textit{Ablation of Object Color Gradient and Shape Concavity}: We combine the above two ablations for each object, getting datasets: {\hl{MOViC-C+S}/YCB-C+S/ScanNet-C+S/COCO-C+S}.
\end{itemize}

\hl{Figures \mbox{\ref{fig:main_object_level_ablation_factors}} (1-a)(1-b)(1-c)(1-d) show new distributions of two object-level factors and two scene-level factors on the -C ablated datasets.} We can see that the Object Color Gradient becomes all zeros in Figure \ref{fig:main_object_level_ablation_factors} (1-a), even simpler than three synthetic datasets. Yet, the distributions of other three factors are almost the same as the original ones, \ie{}, \hl{Figure \mbox{\ref{fig:main_object_level_ablation_factors}} (1-b)(1-c)(1-d) being similar to Figure \mbox{\ref{fig:main_exp_original_factors}} (1-b)(1-c)(1-d).}

Figure \ref{fig:main_object_level_ablation_factors} (2-b) shows that the distributions of Object Shape Concavity of -S ablated datasets now become similar to the synthetic datasets, while the distributions of other three factors are unaffected, \ie{}, Figure \ref{fig:main_object_level_ablation_factors} \hl{(2-a)(2-c)(2-d)} being similar to Figure \ref{fig:main_exp_original_factors} \hl{(1-a)(1-c)(1-d)}. Note that, for the -C+S ablations, the distributions will be the same as shown in Figures \ref{fig:main_object_level_ablation_factors} \hl{(1-a), (2-b), (1-c) or (2-c), (1-d) or (2-d)}. Having the three groups of object-level ablated real-world datasets, we then train and evaluate our four unsupervised baselines from scratch on each of the ablated datasets separately.  

\textbf{Brief Analysis:} \hl{Figure \mbox{ \ref{fig:main_obj_ablation_summary}} shows the quantitative segmentation results and the ARP/ARR scores for over/under-segmentation.
Figure \mbox{\ref{fig:segmentation_by_object_complexity}} qualitatively presents the evolvement of segmentation performance from IODINE according to the adjustment in object-level complexity. More qualitative results for all approaches can be found in Figure \mbox{\ref{fig:object_level_ablation_vis}} in the appendix.} We can see that: 
1) Once the pixels of real-world objects are replaced by their mean colors, \ie{}, no color gradients, the object segmentation performance has been significantly improved for almost all methods. \hl{The ARP and ARR scores tend to be closer with -C ablation, indicating both over- and under-segmentation are alleviated.}
2) Reducing the irregularity of real-world objects can also improve object segmentation, although not significantly. \hl{Simplifying object shapes does not effectively reduce the gap between ARP and ARR scores. In most cases, the over/under-segmentation issues remain the same as on the unablated datasets. }
3) Overall, these results show that existing methods are more likely to learn the objectness represented by uniform colors and/or regular objects. However, compared with Figure \ref{fig:main_exp}, the results of current ablated datasets in Figure \ref{fig:main_obj_ablation_summary}(a) still lag behind the synthetic datasets. This means that there should be some other factors that also potentially affect the object segmentation of existing models.

\vspace{-0.2cm}
\subsection{How do scene-level factors affect current models?} \label{sec:exp_scene_factors}
\vspace{-0.2cm}

\begin{wrapfigure}[6]{r}{0.2\textwidth}\vspace{-0.35cm}
\setlength{\abovecaptionskip}{ 1 pt}
\setlength{\belowcaptionskip}{ -6 pt}
\centering
\raisebox{0pt}[\dimexpr\height-1.\baselineskip\relax]{
  \centering
  \includegraphics[width=0.9\linewidth]{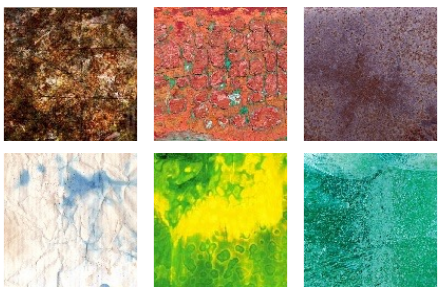}}
\caption{\small{Selected textures.} }
\label{fig:dtd_texture_0}
\end{wrapfigure}

In this section, we investigate to what extent the distributions of scene-level factors affect the segmentation performance with three ablative experiments. \hl{Image examples of scene-level ablations can be found in Figure  \mbox{\ref{fig:ablation_example_1}} in the appendix}.
\vspace{-0.1cm}
\begin{itemize}[leftmargin=*]
\setlength{\itemsep}{1pt}
\setlength{\parsep}{1pt}
\setlength{\parskip}{1pt}
    \item \textit{Ablation of Inter-object Color Similarity}: In each image of datasets of Group 2 (\hl{MOViC}/YCB/ScanNet/COCO), we replace all object textures with a set of new distinctive textures from DTD database \citep{Cimpoi2014}, as shown in Figure \ref{fig:dtd_texture_0}. In this way, the multiple objects look more distinctive in appearance, while the per-object texture gradients are roughly preserved. The ablated datasets are denoted as: \hl{MOViC-T}/YCB-T/ScanNet-T/COCO-T.
    \item \textit{Ablation of Inter-object Shape Variation}: In each image of Group 2 datasets, we normalize the scales of multiple objects by shrinking or expanding the diagonal length of their bounding boxes, such that the new object sizes tend to be uniform. For each object, its shape and texture are linearly scaled up or down. Basically, this aims to remove the diversity of object sizes within single images. The ablated datasets are denoted as: \hl{MOViC-U}/YCB-U/ScanNet-U/COCO-U.   
    \item \textit{Ablation of both Inter-object Color Similarity and Shape Variation}: We simply combine the above two ablation strategies for images. The ablated datasets are denoted as: \hl{MOViC-T+U}/YCB-T+U/ScanNet-T+U/COCO-T+U.   
\end{itemize}

\mbox{Figure \ref{fig:scene_level_ablation_factors}} \hl{(1-a)(1-b)(1-c)(1-d) shows new distributions of object-level and scene-level complexity factors for -T ablated datasets.} We can see that the distributions of Inter-object Color Similarity in Figure \ref{fig:scene_level_ablation_factors} \hl{(1-c)} becomes more similar to synthetic datasets, while the distributions of other three factors are still similar to original ones, \ie{}, Figure \ref{fig:scene_level_ablation_factors} \hl{(1-a)(1-b)(1-d)} being similar to Figure \ref{fig:main_exp_original_factors} \hl{(1-a)(1-b)(1-d)}. 

Figure \ref{fig:scene_level_ablation_factors} \hl{(2-d)} shows that the distributions of Inter-object Shape Variation for ablated datasets now become similar to the synthetic datasets, whereas the distributions of other three factors are still the same as the original ones, \ie{}, Figure \ref{fig:scene_level_ablation_factors} \hl{(2-a)(2-b)(2-c)} being similar to Figure \ref{fig:main_exp_original_factors} \hl{(1-a)(1-b)(1-c)}. Note that, for the ablation of -T+U datasets, the distributions are the same as shown in Figure \ref{fig:scene_level_ablation_factors} \hl{(1-a)(2-b)(1-c)(2-d)}. Having these three groups of scene-level ablated real-world datasets, we then train and evaluate our four unsupervised baselines from scratch on each of the ablated datasets separately.

\textbf{Brief Analysis:} \hl{Figure \mbox{ \ref{fig:main_obj_ablation_summary}} shows the quantitative segmentation results and ARP/ARR/ARI scores for over-/under-segmentation. Figure \mbox{\ref{fig:segmentation_by_scene_complexity}} presents qualitative performance of SlotAtt according to the decrease of scene-level complexity. More qualitative results for all methods can be found in Figure \mbox{\ref{fig:scene_level_ablation_vis}} of the appendix.}
We can see that: 
1) Once the textures of real-world images are replaced by more distinctive textures, \ie{}, with a lower similarity between object appearances, the segmentation performance has been surprisingly boosted remarkably for almost all methods. For MONet and IODINE, the gap between ARR and ARP scores is effectively reduced, indicating a mitigation of over-segmentation (for MONet) and under-segmentation (for IODINE).
2) \hl{Normalizing object sizes over images can also reasonably improve the segmentation performance for AIR through the alleviation of over-segmentation. Its effectiveness, however, is less obvious for other three approaches.}
3) Overall, these results clearly show that existing unsupervised models significantly favor objectness with distinctive appearances in single images. However, compared with Figure \ref{fig:main_exp}, the results on current scene-level ablated datasets are still inferior to synthetic datasets, meaning that the scene-level factors alone are not enough to explain the performance gap.


\begin{figure*}[t]
    \setlength{\abovecaptionskip}{ 1 pt}
    \setlength{\belowcaptionskip}{ -6 pt}
    \centering
       \includegraphics[width=0.8\linewidth]{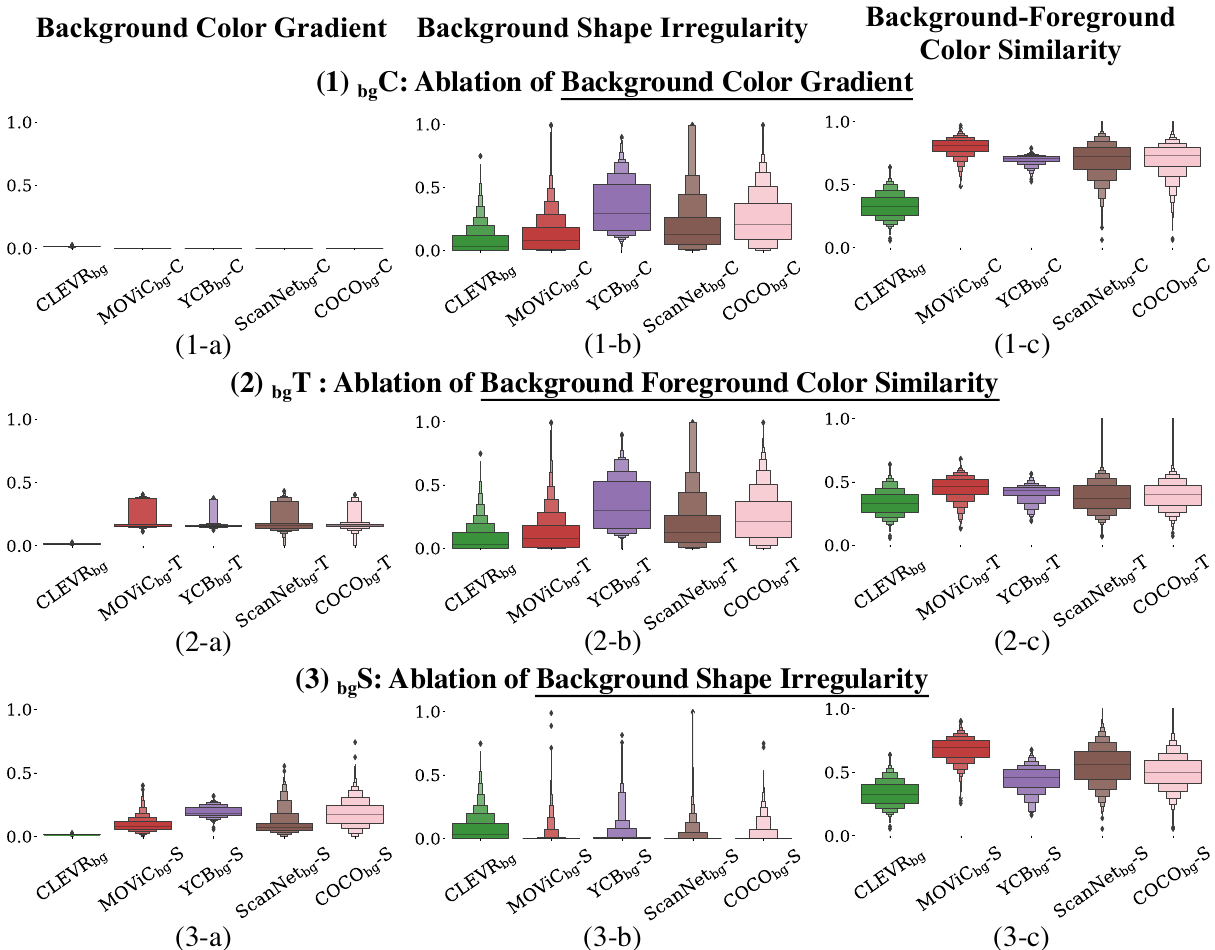}
    \caption{\hl{Distributions of three background-level factors on three types of ablation datasets in Section} \ref{sec:exp_background_factor}.}
    \label{fig:bg_factors_ablation}
\end{figure*}

\vspace{-0.2cm}
\subsection{How do object- and scene-level factors jointly affect current models?} \label{sec:exp_joint_factors}
\vspace{-0.2cm}

In this section, we aim to study how the object- and scene-level factors jointly affect the segmentation performance. We conduct the following three ablative experiments. \hl{Image examples of joint ablations are in Figure \mbox{\ref{fig:ablation_example_2}} in the appendix.}

\vspace{-0.2cm}
\begin{itemize}[leftmargin=*]
\setlength{\itemsep}{1pt}
\setlength{\parsep}{1pt}
\setlength{\parskip}{1pt}
    \item \textit{Ablation of Object Color Gradient, Object Shape Concavity, and Inter-object Color Similarity}: In each image of Group 2 datasets(\hl{MOViC}/YCB/ScanNet/COCO), we replace the object color by averaging all pixels of the distinctive texture, and also replace the object shape with a simple convex hull. The ablated datasets are denoted as: {\hl{MOViC-C+S+T}/YCB-C+S+T/ScanNet-C+S+T/COCO-C+S+T}.   
    \item \textit{Ablation of Object Color Gradient, Object Shape Concavity, and Inter-object Shape Variation}: In each image of the Group 2 datasets (\hl{MOViC}/YCB/ScanNet/COCO), we replace the object color by averaging its own texture, and modify the object shape as convex hull following by size normalization. The ablated datasets are denoted as: {\hl{MOViC-C+S+U}/YCB-C+S+U/ScanNet-C+S+U/ COCO-C+S+U}.   
    \item \textit{Ablation of all four factors}: We aggressively combine all four ablations and get datasets: \hl{MOViC-C+S+T+U}/YCB-C+S+T+U/ ScanNet-C+S+T+U/COCO-C+S+T+U.  
\end{itemize}\vspace{-0.2cm}

Since these ablations are conducted independently, the new distributions of four complexity factors on current jointly ablated datasets are the same as Figures \ref{fig:main_object_level_ablation_factors} \hl{(1-a)(2-b)}, Figure \ref{fig:scene_level_ablation_factors} \hl{(1-c)(2-d)}. We train and evaluate the four baselines from scratch on each of the ablated datasets separately.

\textbf{Brief Analysis:}
\hl{Figure \mbox{\ref{fig:main_obj_ablation_summary}} shows the quantitative segmentation results and the ARP/ARR/ARI scores for comparing over-/under-segmentation. Figure \mbox{ \ref{fig:segmentation_by_object_and_scene_complexity}} presents qualitative performance of MONet according to the decrease of object-level and scene-level complexity.
More qualitative results for all methods can be found in Figure \mbox{\ref{fig:object_scene_level_ablation_vis}} in the appendix.}
 We can see that: 
 1) \hl{Combining the two object-level factors with scene-level color ablation (-C+S+T) can significantly improve segmentation results, especially for SlotAtt as shown in Figure \mbox{\ref{fig:main_obj_ablation_summary}(a)}. Their ARP and ARR scores tend to be closer, suggesting an alleviation of over-/under-segmentation. 
 Combining the two object-level factors with scene-level shape ablation (-C+S+U), however, can have a notable effect only on AIR, where its over-segmentation issue is relieved. }
 2) If the challenging real-world objects and images are ablated in both object- and scene-level, the segmentation performance of all unsupervised models achieves the same level with three synthetic datasets as shown in Figure \ref{fig:main_exp}. 
 3) Overall, these three groups of experiments demonstrate that the failure of unsupervised models on real-world images involves both object- and scene-level dataset biases.

\begin{figure*}[t]
    \setlength{\abovecaptionskip}{ 4 pt}
    \setlength{\belowcaptionskip}{ -12 pt}
    \centering
       \includegraphics[width=0.9\linewidth]{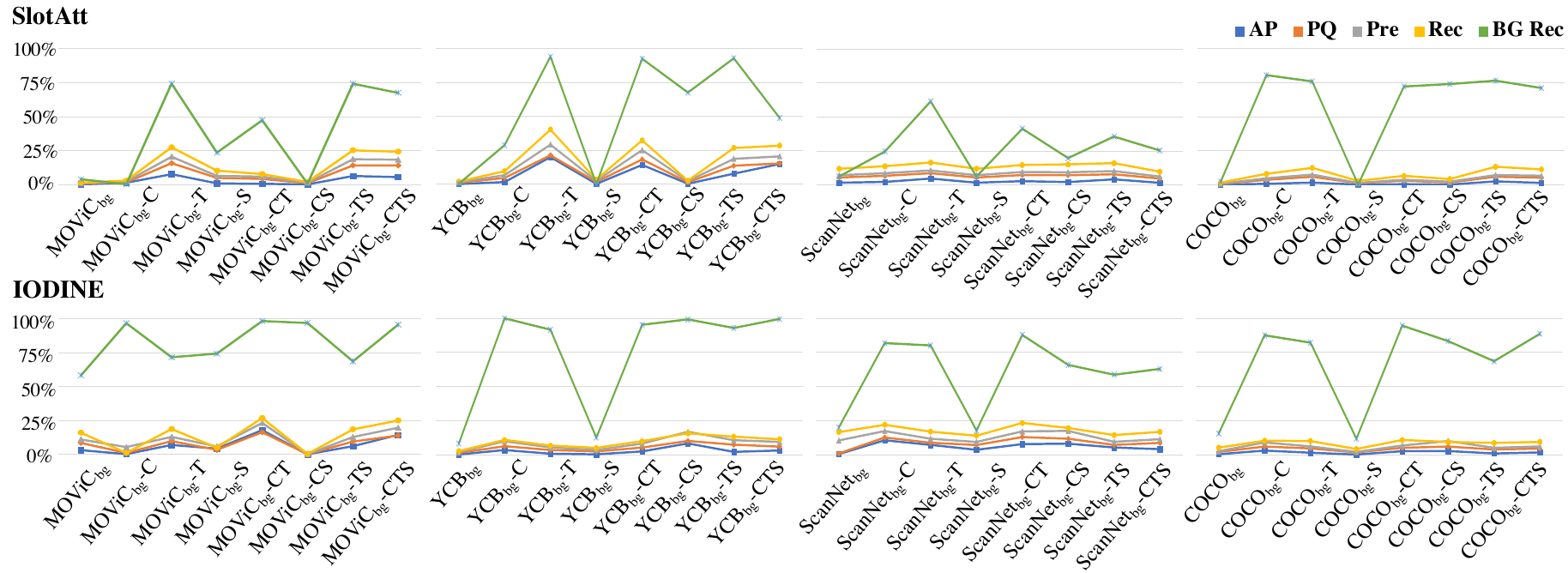}
    \caption{\hl{Quantitative results of IODINE\&SlotAtt on four datasets from Group 4 (with backgrounds) and their ablated datasets.}}
    \label{fig:bg_ablation}
    \vspace{-0.1cm}
\end{figure*}

\begin{figure*}
    \setlength{\abovecaptionskip}{ 4 pt}
    \setlength{\belowcaptionskip}{ -12 pt}
     \centering
     \includegraphics[width=0.9\textwidth]{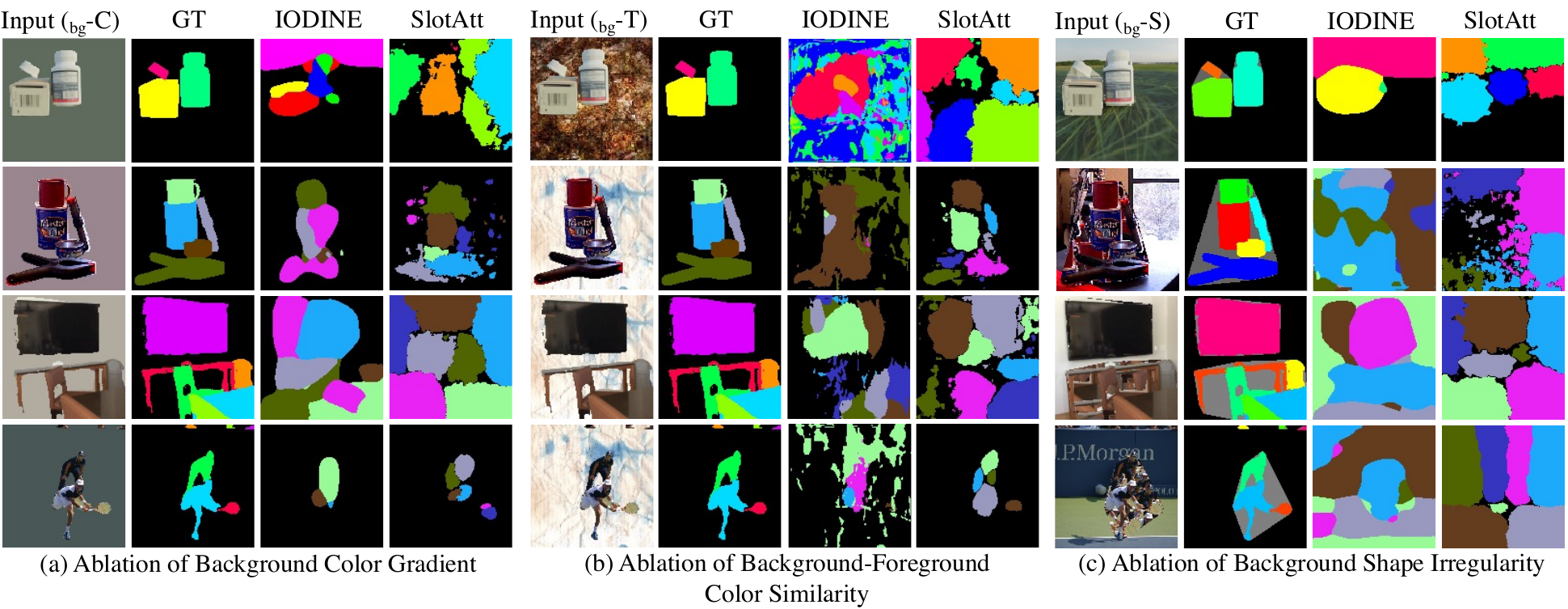}
     \vspace{-0.2cm}
     \label{fig:bg_ablation_vis_a}
\caption{\hl{Qualitative object segmentation results of SlotAtt and IODINE on four datasets of Group 4 (MOVic$_{bg}$/ YCB$_{bg}$/ ScanNet$_{bg}$/ COCO$_{bg}$) with backgrounds.}}
\label{fig:bg_ablation_vis}
\end{figure*}
\vspace{-0.3cm}

\subsection{How do background factors affect current models?}\label{sec:exp_background_factor}
In this section, we investigate to what extent the distributions of background-level factors affect the segmentation performance by conducting the following ablative experiments.

\begin{itemize}[leftmargin=*]
\setlength{\itemsep}{1pt}
\setlength{\parsep}{1pt}
\setlength{\parskip}{1pt}
    \item \textit{Ablation of Background Color Gradient:} For each image of datasets in Group 4, we replace the background pixels with its average color \textit{rgb}. The four ablated datastes are denoted as: \hl{MOViC$_{bg}$-C}/YCB$_{bg}$-C/ ScanNet$_{bg}$-C/COCO$_{bg}$-C.
    \item \textit{Ablation of Background-Foreground Color Similarity:} For each image of Group 4 datasets, we replace the background texture with a new one from DTD database as shown in Figure \ref{fig:dtd_texture_0}. We select the most distinctive color against foreground pixels. In this way, the foreground objects are more distinguishable against the background, while the background gradients are roughly preserved. The three ablated datasets are denoted as: \hl{MOViC$_{bg}$-T}/YCB$_{bg}$-T/ScanNet$_{bg}$-T/COCO$_{bg}$-T.
    \item \textit{Ablation of Background Shape Irregularity:} For each image, we first find all connected subcontours of its background. For each region enclosed by a subcontour, we find the smallest convex hull \citep{Eddins2011} that surrounds it. The original enclosed regions are enlarged to their corresponding convex hulls and filled by shifting around the original region pixels. In this way, the background shape becomes more regular, while the background appearance keeps the same. The three ablated datasets are denoted as: \hl{MOViC$_{bg}$-S}/YCB$_{bg}$-S/ScanNet$_{bg}$-S/COCO$_{bg}$-S.
\end{itemize} \vspace{-0.1cm}

Figure \ref{fig:bg_factors_ablation} shows new distributions of the three background-level factors for the ablated datasets above. Particularly, subfigures (1-a)/(1-b)/(1-c) compares new distributions for the ablated datasets \hl{MOViC$_{bg}$-C}/YCB$_{bg}$-C/ScanNet$_{bg}$-C/COCO$_{bg}$-C. We can see that their Background Color Gradient decreases to zero (Subfig 1-a), and their Background Shape Irregularity (Subfig 1-b) remains the same as original datasets MOViC$_{bg}$/YCB$_{bg}$/ScanNet$_{bg}$/ COCO$_{bg}$ (Subfig (2-b) in Figure \ref{fig:main_exp_original_factors}). 
\hl{Comparing Figure \mbox{\ref{fig:bg_factors_ablation}} (1-c) and Figure \mbox{\ref{fig:main_exp_original_factors}} (2-c), we observe the Background-Foreground Color Similarity is increased inevitably as a byproduct. 
} 

Figure \mbox{\ref{fig:bg_factors_ablation}} (2-a)/(2-b)/(2-c) represents new distributions for the ablated datasets \hl{MOViC$_{bg}$-T}/ YCB$_{bg}$-T/ScanNet$_{bg}$-T/COCO$_{bg}$-T. We can see that their Background-Foreground Color Similarity (Subfig \hl{2-c}) decreases, while Background Color Gradient (Subfig 2-a) and Background Shape Irregularity (Subfig \hl{2-b}) are similar to the original \hl{MOViC$_{bg}$}/ YCB$_{bg}$/ ScanNet$_{bg}$/ COCO$_{bg}$.
Figure \mbox{\ref{fig:bg_factors_ablation}} (3-a)/(3-b)/(3-c) represents new distributions for the ablated datasets \hl{MOViC$_{bg}$-S}/ YCB$_{bg}$-S/ ScanNet$_{bg}$-S/ COCO$_{bg}$-S. We can see that their Background Shape Irregularity (Subfig \hl{3-b}) decreases, while the Background Color Gradient (Subfig 3-a) and Background-Foreground Color Similarity (Subfig \hl{3-c}) remain similar to the original \hl{MOViC$_{bg}$}/YCB$_{bg}$/ScanNet$_{bg}$/COCO$_{bg}$.
We train and evaluate SlotAtt and IODINE from scratch on each of the ablated datasets. Note that, the performance of background segmentation is separately measured by a Recall score, denoted as BG Recall, where a predicted background mask is considered correct if its IoU against a ground truth background mask is above 0.5.

\textbf{Brief Analysis:} Figures \ref{fig:bg_ablation}\&\ref{fig:bg_ablation_vis} show the quantitative and qualitative results respectively. We can see that: 1) Removing the Background Color Gradient or replacing the background texture with a more discriminative one against foreground objects can largely increase BG Recall score. 2) Making the background contours to be more regular alone can hardly benefit the segmentation of objects and backgrounds. 3) Overall, the background can be easily recognized if it is of simple and discriminative color, which however cannot fundamentally alleviate the difficulty of segmenting individual foreground objects. This further confirms that the four object- and scene-level complexity factors introduced in Sections \ref{sec:object_level_complexity_factors}\&\ref{sec:scene_level_complexity_factors} play an essential role in object segmentation of existing models.

\subsection{Why do current unsupervised models fail on real-world datasets?}\label{sec:exp_why_fail}

\vspace{-0.1cm}
As demonstrated in Sections \ref{sec:exp_object_factors}/\ref{sec:exp_scene_factors}/\ref{sec:exp_joint_factors}, once the object- and scene-level complexity factors are removed from the challenging real-world datasets, existing unsupervised models can perform as excellent as on the synthetic datasets, as qualitatively illustrated in Figure \ref{fig:object_scene_level_ablation_vis}. From this,  we can safely conclude that the inductive biases designed in existing unsupervised models are far from able to match with and fully capture the true and complex objectness biases exhibited in real-world images. Nevertheless, from Figure \ref{fig:main_obj_ablation_summary}, each baseline tends to favor different objectness biases. In particular,  

\begin{itemize}[leftmargin=*]
\setlength{\itemsep}{1pt}
\setlength{\parsep}{1pt}
\setlength{\parskip}{1pt}
    \item AIR \citep{Eslami2016}: 
    {As a factor based model, AIR has a strong spatial-locality bias. Despite its poor segmentation performance across all datasets, there is a notable improvement when inter-object shape variation is ablated from real-world datasets (U/T+U/C+S+U/C+S+T+U). More convincingly, even when all other three factors are ablated (C+S+T), it can be hardly improved, showing that object shape variation is a key factor for AIR. \hl{Qualitatively (Figure \mbox{\ref{fig:main_exp_vis}}) and quantitatively (Figure \mbox{\ref{fig:main_arp_arr}}), AIR tends to use multiple bounding boxes for single real-world objects, which leads to over-segmentation. We can see from Figure \mbox{\ref{fig:main_obj_ablation_summary}}(b) that the uniform-scale ablation, especially -T+U datasets, can effectively decrease the gap between ARP and ARR scores, which implies an alleviated over-segmentation.} Since AIR is designed to attend and infer objects as bounding boxes and does not explicitly model backgrounds, the background-level factors are not studied here.}
    \item MONet \citep{Burgess2019}:   
    MONet is more sensitive to color-related factors than shape-related factors. The ablations of object color gradient and inter-object color similarity significantly improve its performance, while ablations of object shape concavity and inter-object shape variation make little difference. For the two color-related factors, the scene-level one is more important than the object-level factor. From this, we can see that MONet has a strong dependency on color. Similar colors tend to be grouped together while different colors are separated apart.
    \hl{From Figure \mbox{\ref{fig:main_obj_ablation_summary}}(b), we observe that the gaps between ARP and ARR scores are relatively small for all experiments on MONet, even when the AP scores are low. This suggests there are both over-segmentation and under-segmentation in the results. An object with various colors tends to be over-segmented, and different objects with similar colors are likely to be grouped together.}
    Due to its heavy reliance on color clues, MONet cannot identify backgrounds without regular and distinctive colors. Thus, background-level factors are not studied for it.
    \item IODINE \citep{Greff2019}:
    IODINE also has a heavy dependency on object-/scene-level color-related factors. However, different from MONet, the ablation on object color gradient brings better performance than inter-object color similarity. We speculate it is because the regularization on shape latent alleviates under-segmentation by biasing towards more regular shapes. 
    \hl{From Figure \mbox{\ref{fig:main_obj_ablation_summary}}(b), we observe both under-/over-segmentation, and the gaps between ARR and ARP scores are not obvious in most cases.}
    Unsurprisingly, as shown in Figure \ref{fig:bg_ablation}, IODINE cannot segment real backgrounds. The ablations of background-level color related factors do improve background-foreground separation, showing that it relies on colors to identify foregrounds from backgrounds. However, the successful background-foreground separation does little help in segmenting individual foreground objects, since the complex biases in real objects still stand.
\begin{figure}[t]
  \centering
  \begin{tabular}{@{}c@{}}
    \includegraphics[width=0.95\linewidth]{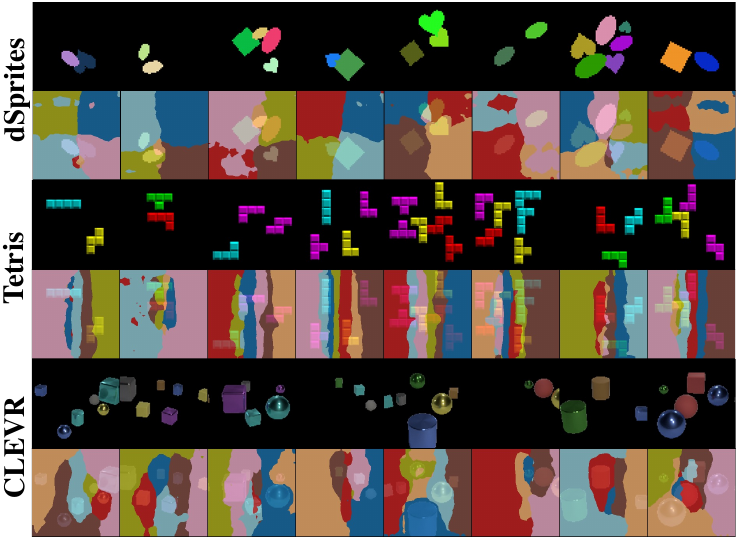} \\
    \small (a)
  \end{tabular}
  \begin{tabular}{@{}c@{}}
    \includegraphics[width=0.98\linewidth]{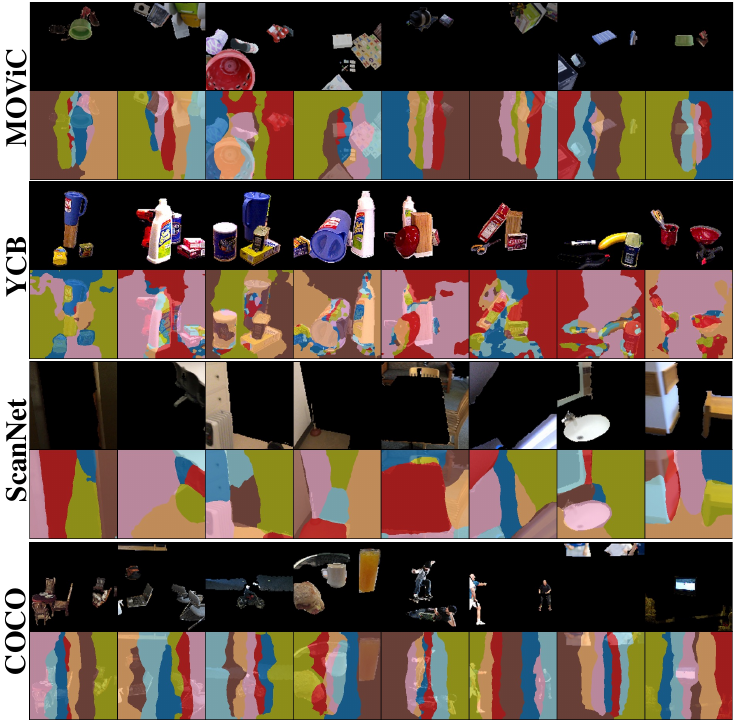} \\
    \small (b)
  \end{tabular}
  \vspace{-0.2cm}
  \caption{\hl{Qualitative object segmentation results of DINOSAUR. (a): On three datasets (dSprites/Tetris/CLEVR) from Group 1 with blank backgrounds; (b): On four datasets (MOViC/YCB/ScanNet/COCO) from Group 2.}}
  \label{fig:dinosaur_result_nobg_1_2}
  \vspace{-0.4cm}
\end{figure}

\begin{figure*}[t]
    \setlength{\abovecaptionskip}{ 4 pt}
    \setlength{\belowcaptionskip}{ -8 pt}
     \centering
     \includegraphics[width=0.85\textwidth]{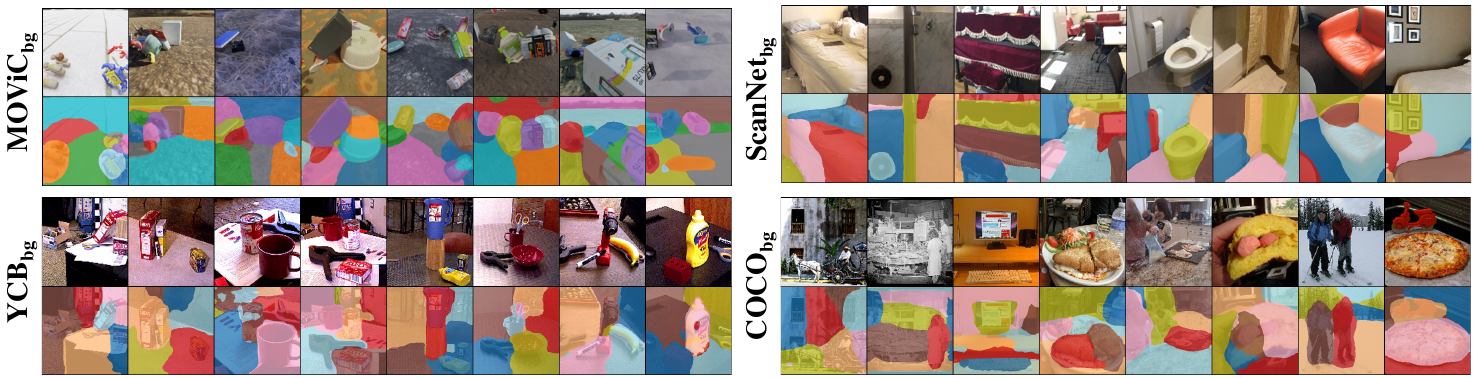}
     \vspace{-0.1cm}
     \label{fig:dinosaur_result_nobg}
\caption{\hl{Qualitative object segmentation results of DINOSAUR on four datasets (MOViC$_{bg}$/YCB$_{bg}$/ScanNet$_{bg}$/COCO$_{bg}$) from Group 4 with backgrounds.}}
\label{fig:dinosaur_result_bg}
\end{figure*}

    \item SlotAtt \citep{Locatello2020}:
    The ablations on all four factors increase the performance of SlotAtt at different levels, among which object- and scene-level color-related factors are more significant. We conjecture that it is because the feature embeddings used by Slot Attention module are learnt from both pixel colors and coordinates, which contributes to its sensitivity to both shape and color factors.
    \hl{By comparing ARP and ARR scores in Figure \mbox{\ref{fig:main_obj_ablation_summary}}, it is observed that MOViC and COCO are more prone to under-segmentation while YCB and ScanNet are more prone to over-segmentation. From Figure \mbox{\ref{fig:main_exp_original_factors}}, it is observed that MOViC and COCO have less distinguishable colors while YCB and ScanNet have more complex shapes. This also illustrates both color and shape are important factors for SlotAtt.}
    As shown in Figure \ref{fig:bg_ablation}, SlotAtt is not able to segment real-world backgrounds. The background-foreground separation in SlotAtt is particularly sensitive to the color contrast between the background and foreground. However, similar to IODINE, a distinguishable background cannot alleviate the burden of foreground object segmentation, as it does not fundamentally remove the complex object- and scene-level factors in real-world images.
\end{itemize}

\subsection{\hl{Can models pretrained on monolithic object images segment real-world images?}}\label{sec:pretrained_model_discussion}

\vspace{-0.5cm}
\begin{table}[ht]
    \setlength{\abovecaptionskip}{ 0 pt}
    \setlength{\belowcaptionskip}{ -6 pt}
    \caption{\hl{Quantitative object segmentation results of DINOSAUR on four datasets (MOViC$_{bg}$/ YCB$_{bg}$/ ScanNet$_{bg}$/ COCO$_{bg}$) of Group 4 with backgrounds. All scores are in percentage (\%).}}
    \label{tab:dinosaur_result}
    \resizebox{0.48\textwidth}{!}{%
    \begin{tabular}{l|ll|llll}
    \textbf{} & \textbf{FG-ARI} & \textbf{mBO} & \textbf{AP} & \textbf{PQ} & \textbf{Pre} & \textbf{Rec} \\ \hline
    \textbf{MOViC$_{bg}$} & 64.9 & 39.2 & 9.2 & 16.9 & 21.6 & 32.9 \\
    \textbf{YCB$_{bg}$} & 52.0 & 24.9 & 1.0 & 3.0 & 4.2 & 6.1 \\
    \textbf{ScanNet$_{bg}$} & 56.2 & 39.1 & 7.6 & 17.2 & 21.0 & 33.3 \\
    \textbf{COCO$_{bg}$} & 51.1 & 27.0 & 1.7 & 7.6 & 9.7 & 16.6
    \end{tabular}%
    }
    \vspace{-0.6cm}
\end{table}

\hl{The models pretrained on monolithic object images have recently achieved promising results on real-world datasets. FreeSOLO \mbox{\citep{wang2022freesolo}} and CutLER \mbox{\citep{wang2023cut}} generate pseudo labels from pretrained features and train detectors interactively with pseudo labels. Odin \mbox{\citep{Henaff2022}} and DINOSAUR \mbox{\citep{seitzer2022bridging}} make use of features pretrained on ImageNet \mbox{\citep{Russakovsky2015}} and apply different clustering strategies. Among this new line of works, we select the representative DINOSAUR to evaluate its segmentation performance on datasets of Groups 1/2/4.}

\hl{\textbf{Brief Analysis:} As shown in Figure \mbox{\ref{fig:dinosaur_result_nobg_1_2}}, DINOSAUR exhibits clear limitations on datasets in Groups 1\&2 where all images just have blank backgrounds. Primarily, this is because DINOSAUR reiles heavily on pretrained ViT features, specifically those extracted from models like DINO \mbox{\citep{Caron2021}} trained on the ImageNet dataset. Such features can hardly generalize to images in Groups 1\&2 datasets where the synthetic objects and blank backgrounds are significantly different from those in ImageNet. 
}

\hl{As shown in Figure \mbox{\ref{fig:dinosaur_result_bg}} and Table \mbox{\ref{tab:dinosaur_result}}, as expected, DINOSAUR demonstrates better and reasonable object segmentation results on the 4 datasets in Group 4. In particular, it shows favorable FG-ARI and mean Best Overlap (mBO) scores, but tends to over-segment objects when the number of available slots exceeds the number of objects present within images. Moreover, it often fails to identify backgrounds as complete entities, instead segmenting them into fragmented pieces. That is why the precision score tend to be lower than the recall score, and the AP score is further diminished.
}

\hl{Overall, by leveraging the large-scale pretrained real-world image features, DINOSAUR shows promising performance to segment objects in real-world images thanks to the feature grouping capability of slot attention based mechanisms \mbox{\citep{Yu2022b,Xu2022a}}. Nevertheless, it is still in its infancy to accurately identify generic objects and separate complex backgrounds. We conjecture that more explicit object biases need to be encoded into the pretrained models or the unsupervised learning process, although the existing pretrained features on monolithic object images may implicitly have the concepts of objectness. More advanced research works are expected in this direction in the future.}

\vspace{-0.6cm}
\section{Conclusions}
We systematically show that existing unsupervised methods are practically impossible to segment generic objects from single real-world images, and investigate the underlying factors that incur the failure. 
With the aid of our carefully designed seven object-, scene-, and background-level complexity factors, we conduct extensive experiments on multiple groups of ablated real-world objects and images, and safely conclude that the distributions of both object-, scene-, and background-level biases in appearance and geometry of real-world datasets are particularly diverse and indiscriminative, such that current unsupervised models cannot segment real objects or backgrounds. Based on this finding, we suggest two main directions for future study: 1) To exploit more discriminative objectness biases such as object motions which expressively describe the ownership of visual pixels as recently explored in \citep{Tangemann2021,Chen2022,Bear2020} for 2D images and in \citep{Song2022} for 3D point clouds. 2) To leverage pretrained features from single-object-dominant datasets which explicitly regard each image as an object as recently studied in \citep{Caron2021,Henaff2022,Seitzer2023}, although such settings are no longer purely unsupervised. \\
\vspace{-0.2cm}


\noindent\textbf{Acknowledgements:}
This work was supported in part by National Natural Science Foundation of China (62271431), in part by Shenzhen Science and Technology Innovation Commission (JCYJ20210324120603011), in part by Research Grants Council of Hong Kong (25207822).

\clearpage
\bibliographystyle{spbasic}      
\bibliography{references}   

\clearpage
\appendix
\section{Appendix}

\subsection{Other Candidates of Object- and Scene-level Complexity Factors }

In addition to the primary four complexity factors in Sections \ref{sec:object_level_complexity_factors}\&\ref{sec:scene_level_complexity_factors}, we also explore other potential complexity factors to quantitatively measure the distributions of object- and scene-level biases in appearance and geometry. Basically, we aim to consider as many aspects as possible to investigate key factors underlying the distribution gaps between synthetic and real-world datasets. 
However, we empirically find that these candidate factors do not show significant discrepancies between synthetic and real-world datasets. Details are shown below.
\vspace{-0.2cm}

\subsubsection{Candidates of Object-level Complexity Factors}

\begin{itemize}[leftmargin=*]
\setlength{\itemsep}{1pt}
\setlength{\parsep}{1pt}
\setlength{\parskip}{1pt}
    \item \textbf{Object Color Count:} 
    This factor is defined as the total number of unique colors within an object mask. Basically, this is to simply measure the diversity of object colors. 
    \item \textbf{Object Color Entropy:}
    Inspired by Shannon entropy \citep{shannon1948mathematical}, we calculate the entropy value at each pixel by applying a $3 \times 3$ filter on the grayscale image converted from RGB. In particular, for each pixel, its color value becomes a discrete value in $[0, 255]$. We compute its entropy score: $H(x) = -\sum_{i=1}^n p(x_i) \log_2 p(x_i)$, where $p(x_i)$ denotes the probability of a specific color value $x_i$ within the $3\times 3$ neighbourhood. Basically, this factor aims to measure the color diversity within $3\times 3$ image patches. The higher this factor, the more frequently the object color changes in small local areas. 
    \item \textbf{Object Shape Non-rectangularity:}
    Given the binary mask of an object ($\boldsymbol{M}_{obj}\in \mathbb{R}^{H\times W}$), we first calculate its axis-aligned bounding box ($\boldsymbol{M}_{bbox} \in \mathbb{R}^{H\times W}$). Object shape non-rectangularity is calculated as $1-\sum\boldsymbol{M}_{obj} / \sum\boldsymbol{M}_{bbox}$. Similar to object shape concavity, this factor is also designed to measure the complexity of object shapes. However, this factor is more likely to be affected by the object orientation since it takes axis-aligned bounding boxes as a reference.
    \item \textbf{Object Shape Incompactness:}
    There are two similar methods to quantify the compactness of object shapes. The first one is Polsby–Popper test \citep{polsby1991third}: $PP(\boldsymbol{M}_{obj}) = 4\pi A(\boldsymbol{M}_{obj}) /P(\boldsymbol{M}_{obj})^2 $. The other is Schwartzberg \citep{schwartzberg1965reapportionment} compactness score: $S(\boldsymbol{M}_{obj}) = (2 \pi \sqrt{A(\boldsymbol{M}_{obj}) / \pi}) / P(\boldsymbol{M}_{obj})$. In both formula, $P(\boldsymbol{M}_{obj})$ is the object perimeter and $A(\boldsymbol{M}_{obj})$ is the object area. For simplicity, we choose $PP(\boldsymbol{M}_{obj})$ to calculate the object shape incompactness score: $1 - PP(\boldsymbol{M}_{obj})$. 
    \item \textbf{Object Shape Discontinuity:}
    Given an object mask ($\boldsymbol{M}_{obj}\in \mathbb{R}^{H\times W}$), we first find the largest connected component ($\boldsymbol{M}_{lcc}\in \mathbb{R}^{H\times W}$) in its binary mask. The discontinuity of shape is calculated as: $1 - \sum\boldsymbol{M}_{lcc} / \sum\boldsymbol{M}_{obj}$. This factor is to evaluate how continuous an object shape is. 
    \item \textbf{Object Shape Decentralization:} Given an object mask, we first calculate its centroid (${\Bar{x}, \Bar{y} }$) by averaging all pixel coordinates in the object. Then, the second moment of this object is calculated as: $\sum_x\sum_y (x - \Bar{x})^2 (y - \Bar{y})^2$, where $(x, y)$ is the coordinates of pixels within the object. The higher this factor, the object shape is less likely to be centralized.  
\end{itemize}
\vspace{-0.2cm}

\begin{figure*}[th]
    \vspace{-0.2cm}
    \setlength{\abovecaptionskip}{ 0 pt}
    \setlength{\belowcaptionskip}{ -2 pt}
    \centering
       \includegraphics[width=0.95\linewidth]{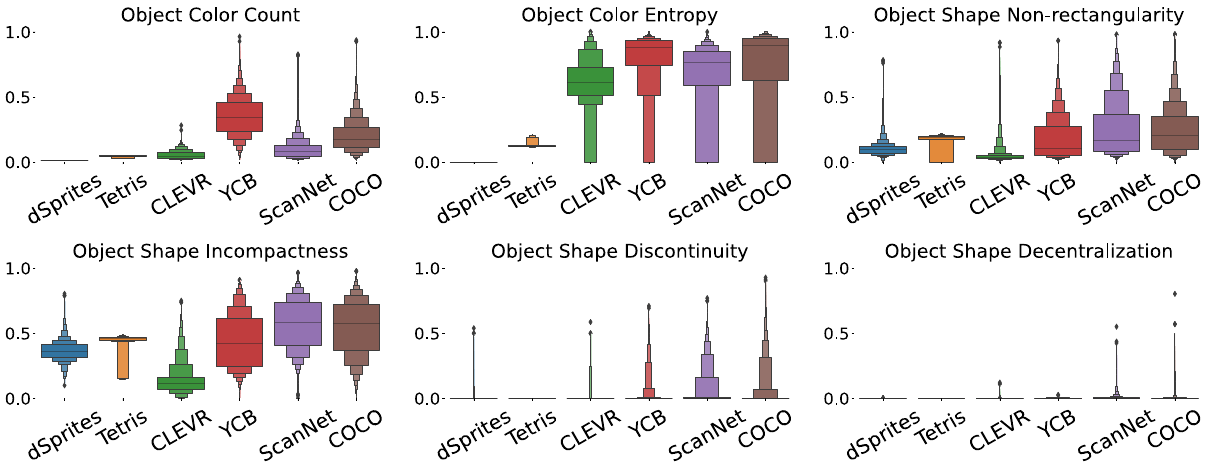}
    \caption{\hl{Distributions of additional candidates of Object-level Complexity Factors.}}
    \label{fig:app_candidate_object_level_factors}
\end{figure*}

As shown in Figure \ref{fig:app_candidate_object_level_factors}, we compare the distributions of the  object-level factor candidates on both synthetic and real-world datasets. It can be seen that the majority of these factors do not show significant gaps between the simple synthetic and the challenging real-world datasets. Therefore, we do not conduct relevant ablation experiments.

\begin{figure*}[h]
    \setlength{\abovecaptionskip}{ 0 pt}
    \setlength{\belowcaptionskip}{ -8 pt}
    \centering
       \includegraphics[width=0.95\linewidth]{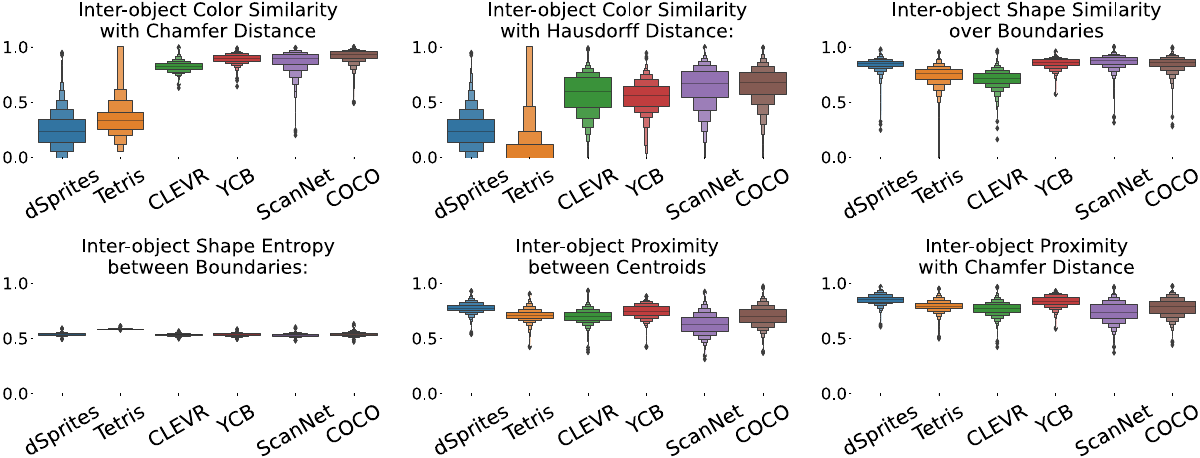}
      \vspace{0.05cm}
    \caption{\hl{Distributions of additional candidates of Scene-level Complexity Factors.}}
    \label{fig:app_candidate_scene_level_factors}
\vspace{-0.2cm}
\end{figure*}

\begin{figure}[h]
    \setlength{\abovecaptionskip}{ -2 pt}
    \setlength{\belowcaptionskip}{ -2 pt}
    \centering
       \includegraphics[width=0.75\linewidth]{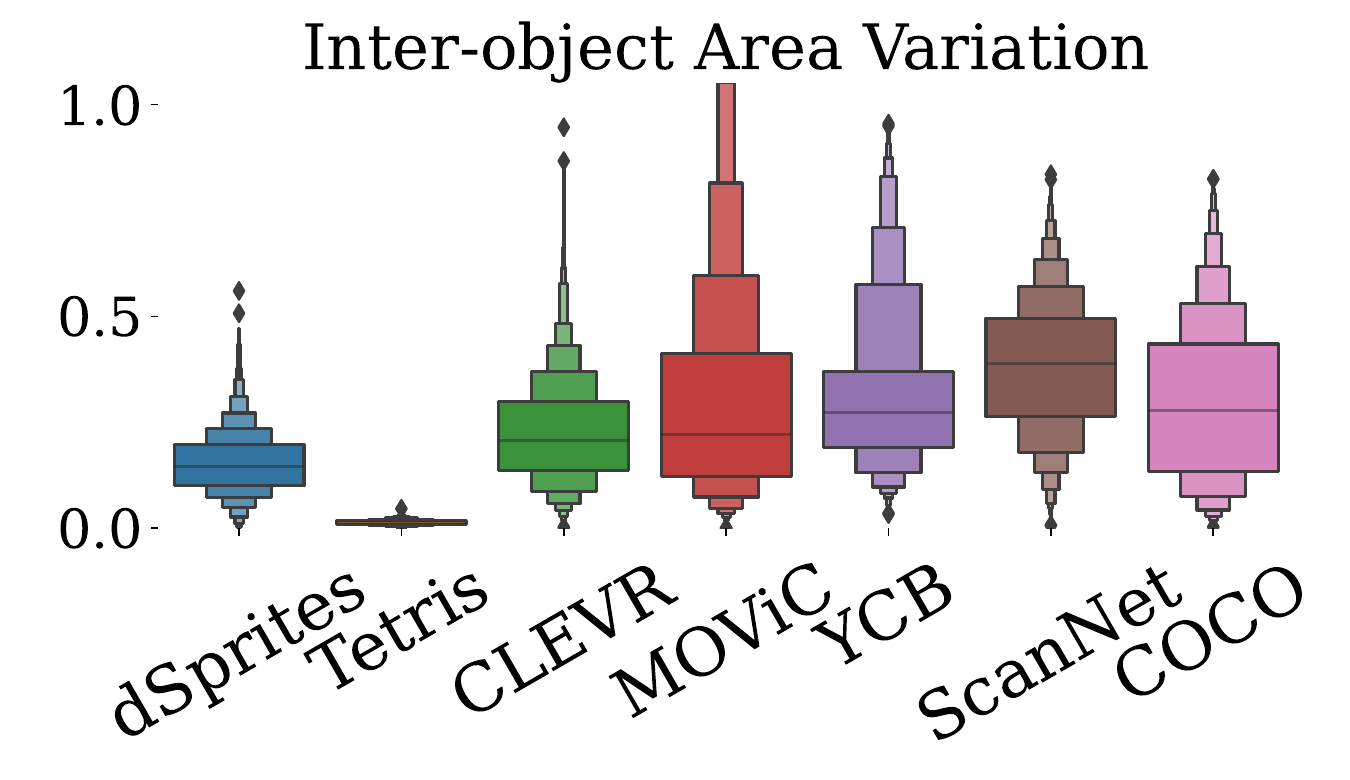}
      \vspace{0.05cm}
    \caption{Distributions of additional candidates of Inter-object Area Variation.}
    \label{fig:area_variation}
\vspace{-0.2cm}
\end{figure}

\subsubsection{Candidates of Scene-level Complexity Factors }

\begin{itemize}[leftmargin=*]
\setlength{\itemsep}{1pt}
\setlength{\parsep}{1pt}
\setlength{\parskip}{1pt}
    \item \textbf{Inter-object Color Similarity with Chamfer Distance:}
    In the calculation of this factor, we first convert each pixel into a point in RGB space. In this way, each object can be represented by a point set in the RGB space. This factor is calculated between each pair of objects by measuring the Chamfer distance of two point sets in the RGB space. Since Chamfer distance is an asymmetric measurement, we calculate and average out the bidirectional Chamfer distances. Compared with Euclidean distance, this measurement favors the most similar colors between two objects.  
    \item \textbf{Inter-object Color Similarity with Hausdorff Distance:} This factor is similar to the previous one. The only difference is that we replace Chamfer distance with Hausdorff distance. Hausdorff distance is also a directed and asymmetric measurement, so the final score is the average of distance values in both directions.
    \item \textbf{Inter-object Shape Similarity over Boundaries:}
    For each object mask, we first find its boundary using the method in \citep{Cheng2021a}, and then crop it with its axis-aligned bounding box. Each bounding box is scaled and fits into a unit box with its original aspect ratio. Lastly, we calculate the IoU between the boundaries of two objects to measure their shape similarity.
    \item \textbf{Inter-object Shape Entropy between Boundaries:}
    We first combine all object masks into a single image by assigning different indices to different objects. Then we compute the entropy of each pixel with a $3 \times 3$ filter. The final factor score is calculated by averaging all non-zero entropy values. Note that, the interior part of objects and background will not be considered because their entropy values will always be zeros. Basically, this factor is designed to evaluate how crowded an image is. The higher this factor, the more objects are spatially adjacent.
    \item \textbf{Inter-object Proximity between Centroids:}
    We first calculate the centroid (${\Bar{x}, \Bar{y} }$) of each object by averaging all pixel coordinates in the object mask. Euclidean distances between object centroids are then computed pair-wisely before they are averaged to be the final factor score. This factor is designed to measure the spatial proximity of multiple objects in a single image.
    \item \textbf{Inter-object Proximity with Chamfer Distance:}
    To measure the spatial proximity between objects, we also calculate the spatial Chamfer distance between objects. Specifically, each object is represented by a set of $(h/w)$ coordinates, and the average of pair-wise bidirectional Chamfer distances is calculated as the proximity score for each image.
    \item \hl{\textbf{Inter-object Area Variation:}}
    \hl{In order to measure the variation of objects in terms of their scale, We first calculate the area of each object, and then compute the pair-wise absolute difference for all object areas, obtaining a $K \times K$ matrix. The final inter-object area variation is the average of the matrix excluding diagonal entries.}
\end{itemize}

As shown in Figure \ref{fig:app_candidate_scene_level_factors}, we compare the distributions of the scene-level candidate factors on both synthetic and real-world datasets. We can see that both the inter-object color similarity with Chamfer and Hausdorff distances share similar distribution gaps with our primary inter-object color similarity factor defined in Section \ref{sec:complexity_factors}. The remaining four candidate factors relating to inter-object shape complexity do not show significant distribution gaps between the synthetic and real-world datasets. In this regard, we choose not to conduct ablation experiments on these six candidate factors. \hl{The distribution for Inter-object Area Variation is presented in Figure }\ref{fig:area_variation} \hl{. Overall, the inter-object area variation for synthetic datasets is higher than real-world datasets. However, comparing Inter-object Area Variation in Figure } \ref{fig:area_variation} \hl{and Inter-object Shape Variation in Figure } \ref{fig:main_complexity_factors} \hl{(1-d), we find that factor values for Inter-object Shape Variation have smaller variation within each dataset, indicating it better captures the scene-level property of datasets. Besides, Inter-object Shape Variation does not only take into account the scale of objects but also the orientations. Thus, we select Inter-object Shape Variation over Inter-object Area Variation.}


\subsubsection{\hl{Complexity Factors Measured Across Dataset}}

\begin{figure*}[ht]
    \setlength{\abovecaptionskip}{ -1 pt}
    \setlength{\belowcaptionskip}{ -2 pt}
    \centering
       \includegraphics[width=0.95\linewidth]{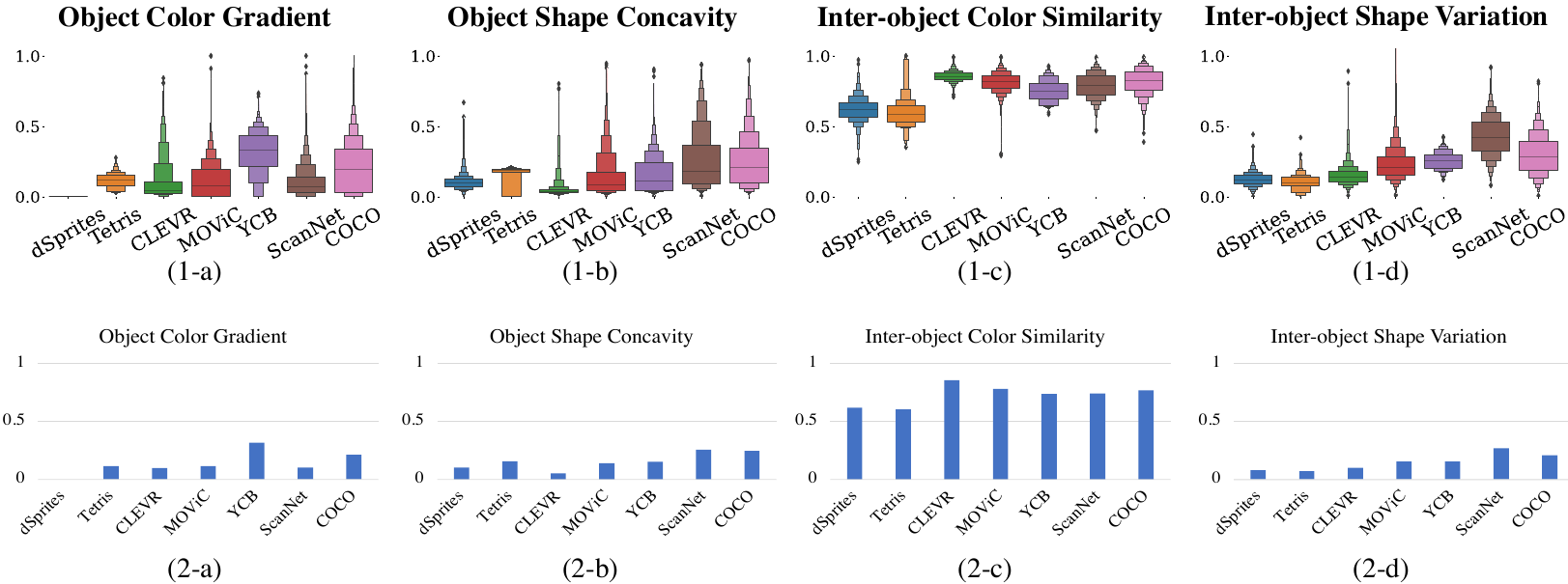}
    \caption{\hl{Complexity calculated across object/image v.s complexity calculated across the dataset.}}
    \label{fig:dataset_level_factor}
\end{figure*}

\hl{Object-level complexity factors are calculated for each object as shown in Figure} \ref{fig:dataset_level_factor}\hl{(1-a)(1-b). Scene-level complexity factors are calculated for each image as shown in Figure} \ref{fig:dataset_level_factor}\hl{(1-c)(1-d). We additionally measure the four complexity factors across the datasets as shown in Figure} \ref{fig:dataset_level_factor}\hl{(2-a)(2-b)(2-c)(2-d). To measure object-level complexity factors across the datasets, we directly calculate the average of all object complexity values. To measure the scene-level complexity factors across the datasets, we make the color/shape comparison across the dataset and calculate the factor value. We find that complexity factors calculated across the dataset follow similar patterns as those calculated across the object/image. }

\begin{figure*}[h]
    \setlength{\abovecaptionskip}{ -1 pt}
    \setlength{\belowcaptionskip}{ -2 pt}
    \centering
       \includegraphics[width=0.5\linewidth]{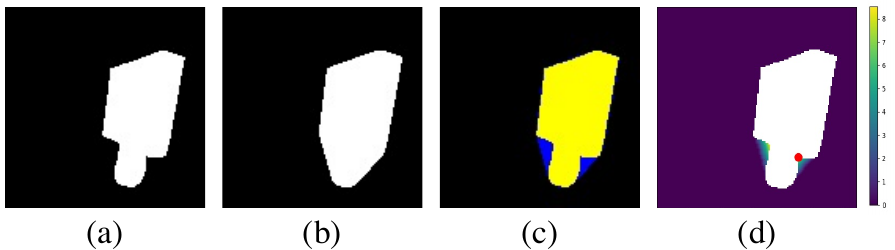}
      \vspace{0.05cm}
    \caption{\hl{Find the deepest concavity for a 2D shape.}}
    \label{fig:maximum_inscribed_convex_1}
\end{figure*}
\begin{figure*}[h]
    \setlength{\abovecaptionskip}{ -1 pt}
    \setlength{\belowcaptionskip}{ -2 pt}
    \centering
       \includegraphics[width=0.95\linewidth]{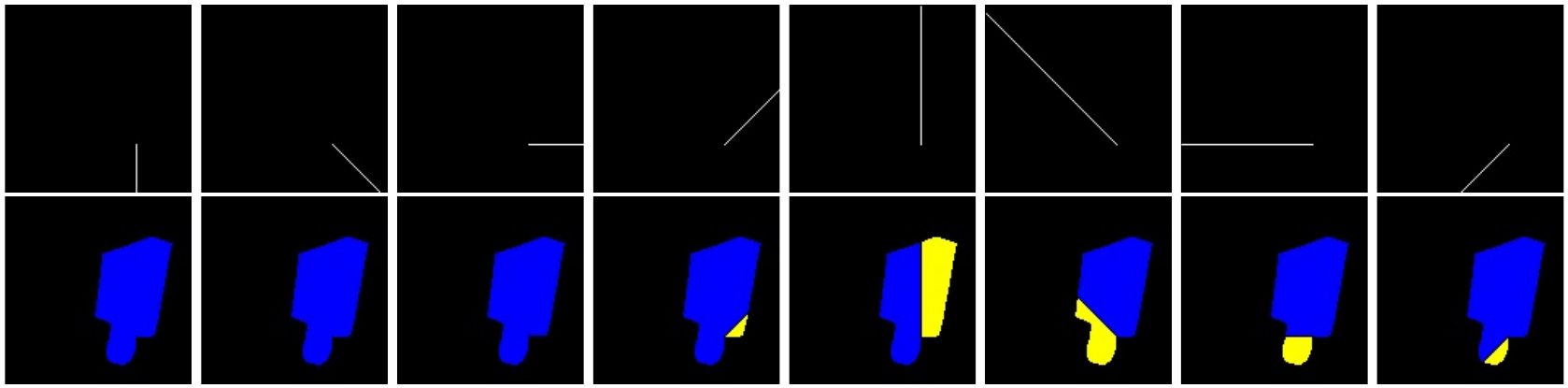}
    \caption{\hl{Find the cut from the deepest concavity that leads to minimum area reduction. The first row shows eight rays with pre-defined directions. The second row shows the sub-regions generated from cuts. }}
    \label{fig:maximum_inscribed_convex_2}
\vspace{-0.2cm}
\end{figure*}

\subsection{More Details of Background-level Complexity Factors}
\begin{itemize}[leftmargin=*]
\setlength{\itemsep}{1pt}
\setlength{\parsep}{1pt}
\setlength{\parskip}{1pt}
    \item \textbf{Background-Foreground Color Similarity:} 
    Given all background pixels and the foreground pixels in a single image, we calculate the Euclidean distance between each background pixel and each foreground pixel in the RGB space. This results in a $U \times V$ matrix ${E}$  where $U$ and $V$ represent the number of pixels in background and foreground respectively. We treat $255\sqrt{3} - {E}$ as the cost matrix for a Hungarian algorithm and solve the optimal assignment between foreground pixels and background pixels. Each background pixel will be assigned to a distant foreground pixel in the RGB space. Given the assignment between foreground pixels and background pixels, the background-foreground color distance is calculated as the Euclidean distance between assigned pairs. The final factor value for background-foreground color similarity is computed as: 1 - background-foreground color distance / $255\sqrt{3}$. The higher this factor, the more similar background and foreground appear to be, and it is harder to separate each other.
    \item \hl{\textbf{Background Shape Irregularity:} }
    \hl{ In order to compute the maximal inscribed convex set of a connected region, we follow the algorithm proposed by }\cite{borgefors2005approximation}\hl{. The main idea is to iteratively find the deepest concavity and remove it with minimum area cost.  Given a randomly shaped connected region $R \in \mathbb{R}^{H\times W}$ (Figure }\ref{fig:maximum_inscribed_convex_1}\hl{ (a)), we first compute the smallest convex polygon mask (convex hull) $C \in \mathbb{R}^{H\times W}$ as shown in Figure }\ref{fig:maximum_inscribed_convex_1}\hl{(b). The convex deficiency $D$ is defined as the difference between the region $R$ and its convex hull $C$: $D = C - R, D \in \mathbb{R}^{H\times W}$ as shown in the blue region in Figure }\ref{fig:maximum_inscribed_convex_1}\hl{(c). Then we calculate the constrained distance transform }\citep{piper1987computing}\hl{ for the convex deficiency $D$ where the region $R$ is the constraint. On the calculated distance transform map, the deepest concavity $dc \in \mathbb{R}^{2}$ is the element with the largest distance label as shown in Figure }\ref{fig:maximum_inscribed_convex_1}\hl{(d). In order to remove the deepest concavity $dc$, we cut the region with a straight line from $dc$ with 8 directions $( \frac{n \times \pi}{4}, n=0,1,...7)$ as shown in Figure }\ref{fig:maximum_inscribed_convex_2}]\hl{. The cut that generates the smallest sub-region is selected and the smallest sub-region is removed. The above process is repeated until the remained sub-region is convex ($distance\: label(dc) <= 3$). The remained convex region is the maximal inscribed convex set.}  

\end{itemize}

\clearpage
\subsection{Implementation Details of Baselines}

\indent \textbf{AIR \citep{Eslami2016}}
\begin{itemize}[leftmargin=*]
    \setlength{\itemsep}{1pt}
    \setlength{\parsep}{1pt}
    \setlength{\parskip}{1pt}
    \item \textbf{Source Code:} We refer to {\scriptsize{\url{https://pyro.ai/examples/air.html}}} and {\scriptsize{\url{https://github.com/addtt/attend-infer-repeat-pytorch}}} for the implementation.
    \item \textbf{Important Adaptations:} We use an additional parameter to weight the KL divergence loss and reconstruction loss. For each experiment, we choose the weight for KL divergence from 1, 10, 25 and 50. The highest AP score is kept.
    \item \textbf{Training Details:} All experiments of AIR \citep{Eslami2016} are conducted with a batch size of 64. The learning rate is set to $1\mathrm{e}{-4}$ for training inference networks and decoders, $1\mathrm{e}{-3}$ for baselines which are the same as the original paper. Since the number of objects in our datasets ranges between 2 and 6, we set the maximum number of steps at inference to be 6 for all experiments. 
    All models are trained on a single GPU for 1000 epochs. We perform an evaluation every 50 epochs and select the highest AP score.
    \item \textbf{\hl{Implementation Details:}} \hl{LSTMs have 256 cell units and object appearances are coded with 50 units. The hidden dimensions for appearance encoder and decoder are both 200. Images are normalized to hold values between 0 and 1 and the likelihood function is a Gaussian with a fixed standard deviation equal to 0.3. The presence prior p(n) is fixed to a geometric distribution that favors sparse reconstructions (p=0.01). The location prior is a normal distribution with mean [3, 0, 0] and standard deviation [0.2, 1, 1], presenting the scale, x-position and y-position respectively. The appearance prior is a standard normal distribution with a dimension of 50. }
\end{itemize}
\vspace{0.1cm}
        
\noindent\textbf{MONet \citep{Burgess2019}}
\begin{itemize}[leftmargin=*]
    \setlength{\itemsep}{1pt}
    \setlength{\parsep}{1pt}
    \setlength{\parskip}{1pt}
    \item \textbf{Source Code:} We refer to \citep{Engelcke2020}'s re-implementation at: \url{https://github.com/applied-ai-lab/genesis}.
    \item \textbf{Important Adaptations:} We train MONet with the GECO objective \citep{rezende2018taming} following the protocol mentioned in \citep{Engelcke2020}.
    \item \textbf{Training Details:} All experiments on MONet are conducted with a batch size of 32 and a learning rate of $1\mathrm{e}{-4}$. Since the maximum number of components is 7, including 1 background and 6 objects, we set the number of steps to be 7 for all experiments. All models are trained on a single GPU for 200 epochs with the training loss converged. We perform an evaluation every 10 epochs and select the highest AP score.
    \item \textbf{\hl{Implementation Details:}} \hl{The VAE encoder is a standard CNN with 3x3 kernels, stride of 2, and ReLU activations. It employs an attention network with [32, 32, 64, 64, 64] filters in the encoder and the reverse in the decoder It receives the concatenation of the input image x and the attention mask in logarithmic units, $log m_k$ as input. The CNN layers output (32, 32, 64, 64) channels respectively. The CNN output is flattened and fed to a 2 layer MLP with output sizes of (256, 32). The MLP output parameterises the $\mu$ and $log\sigma$ of a 16-dim Gaussian latent posterior. The VAE uses a broadcast decoder }\citep{watters2019spatial} \hl{consisting of a four-layer CNN with no padding, 3x3 kernels, stride 1, 32 output channels and ReLU activations. A final 1x1 convolutional layer transforms the output into 4 channels. For all experiments, the output component distribution is an independent pixel-wise Gaussian with fixed scales (std=0.7). The attention network is a standard UNet } \citep{ronneberger2015u} \hl{ with five blocks. Each block consists of the following: a 3x3 bias-free convolution with stride 1, followed by instance normalization with a learned bias term, followed by a ReLU activation, and finally downsampled or upsampled by a factor of 2 using nearest neighbour-resizing (no resizing occurs in the last block of each path).}
\end{itemize}
\vspace{0.1cm}

\newpage
\noindent\textbf{IODINE \citep{Greff2019}}
\begin{itemize}[leftmargin=*]
    \setlength{\itemsep}{1pt}
    \setlength{\parsep}{1pt}
    \setlength{\parskip}{1pt}
    \item \textbf{Source Code:} We use the official implementation at: {\scriptsize{\url{https://github.com/deepmind/deepmind-research/tree/master/iodine}}}
    \item \textbf{Important Adaptations:} The architecture is set the same as what is used for CLEVR dataset \citep{Johnson2017} in the original paper \citep{Greff2019}.
    \item \textbf{Training Details:} Since we use a single GPU for the training of all models, the batch size is adjusted to be 4 and the learning rate $0.0001 \times \sqrt{1/8}$. The number of slots $K$ is set as 7 and the inference iteration $T$ as 5. We train each model for $500K$ iterations until the loss is fully converged.
     \item \textbf{\hl{Implementation Details:}} \hl{The convolutional layers in the encoder and decoder have five layers with size-5 kernels, strides of [1, 2, 1, 2, 1], and filter sizes of [32, 32, 64, 64, 64] and [64, 32, 32, 32, 32], respectively. Fully-connected layers are used at the lowest resolution. The autoregressive prior is implemented as an LSTM with 256 units. The conditional distribution is parameterized by a multilayer perception (MLP) with two hidden layers, 256 units per layer.}
\end{itemize}
\vspace{0.1cm}
    
\noindent\textbf{SlotAtt \citep{Locatello2020}}
\begin{itemize}[leftmargin=*]
    \setlength{\itemsep}{1pt}
    \setlength{\parsep}{1pt}
    \setlength{\parskip}{1pt}
    \item \textbf{Source Code:} We use the official implementation at: {\scriptsize{\url{https://github.com/google-research/google-research/tree/master/slot_attention}}}.
    \item \textbf{Important Adaptations:} The architecture is set the same as what is used for CLEVR dataset \citep{Johnson2017} in the original paper \citep{Locatello2020}.
    \item \textbf{Training Details:} All experiments of SlotAtt are conducted with a batch size of 32 and learning rate selected from [$4\mathrm{e}{-4}$, $4\mathrm{e}{-5}$]. The number of slots $K$ is set as 7 and the number of iterations $T$ is set as 3. All models are trained on a single GPU for $500K$ iterations until the loss is fully converged.
    \item \textbf{\hl{Implementation Details:}} \hl{For all experiments, we use a slot feature dimension of 64. The GRU has a 64-dimensional hidden state and the feedforward block is an MLP with a single hidden layer of size 128 and ReLU activation followed by a linear layer. The CNN Encoder used in our experiments consists of 5 convolutional layers where each filter is 5x5 with channel 64, and ReLU activations are used after each convolutional layer. After this backbone, we add position embeddings and then flatten the spatial dimensions. After applying a layer normalization, we finally add $1 \times 1$ convolutions which we implement as a shared MLP applied at each spatial location with one hidden layer of 64 units. The spatial broadcast decoder} \citep{watters2019spatial} \hl{is applied independently on each slot representation with shared parameters between slots. We first copy the slot representation vector of dimension 64 onto a grid of shape $8\times8\times64$, after which we add a positional embedding. Finally, this representation is passed through several de-convolutional layers symmetrical to the encoder.}
\end{itemize}
\vspace{0.1cm}

\noindent\textbf{{Mask R-CNN} \citep{He2017a}}
\begin{itemize}[leftmargin=*]
    \setlength{\itemsep}{1pt}
    \setlength{\parsep}{1pt}
    \setlength{\parskip}{1pt}
    \item \textbf{{Source Code:}} {We use the implementation at:} \url{https://github.com/matterport/Mask_RCNN}.
    \item \textbf{{Important Adaptations:}} {We use the same settings as training for COCO in the above repository}. 
    \item \textbf{{Training Details:}} {Training for all datasets starts from the pre-trained COCO weights (mask\_rcnn\_coco.h5) from} \url{https://github.com/matterport/Mask_RCNN/releases}. {All models are trained on a single GPU for 30 epochs until the loss is fully converged.}
    \item \textbf{{Training Details:}} \hl{The stride for the backbone is set to [4, 8, 16, 32, 64]. The maximum number of instances to detect in one image is 100. The minimum confidence is 0.5 and NMS ratio is 0.3. The scale of the RPN anchor is (32, 64, 128, 256, 512) and stride is 2. }
\end{itemize}    

\clearpage
\noindent\textbf{\hl{DINOSAUR} \citep{seitzer2022bridging}}
\begin{itemize}[leftmargin=*]
    \setlength{\itemsep}{1pt}
    \setlength{\parsep}{1pt}
    \setlength{\parskip}{1pt}
    \item \textbf{\hl{Source Code:}} \hl{We use the official implementation at: }{\tiny{\url{https://github.com/amazon-science/object-centric-learning-framework}}}.
    \item \textbf{\hl{Important Adaptations:}} \hl{We use a similar model architecture as for MOViC dataset in the original paper. In light of the fact that our datasets only consist of 2-6 objects, we adapt the number of slots to be 7. And since we are using a single GPU for training, we adopt 64 as batch size. }
    \item \textbf{\hl{Training Details:}} \hl{We train DINOSAUR using the Adam optimizer with a learning rate of $4\mathrm{e}{-4}$, a linear learning rate warm-up of 10 000 optimization steps and an exponentially decaying learning rate schedule. All models are trained on a single GPU with a batch size of 64. }
    \item \textbf{\hl{Implementation Details:}} \hl{ We use a ViT with patch size 8 as the feature extractor and the MLP decoder for all datasets considered. The ViT-8 we used has the token dimensionality of 384 and 6 heads. All models use 12 Transformer blocks, linear patch embedding and additive positional encoding. The output of the last block (not applying the final layer norm) is passed on to the Slot Attention module and used in the feature reconstruction loss, after removing the entry corresponding to the CLS token. For the pre-trained weights, we use the timm library for DINO and the specific timm model name is \textit{vit\_small\_patch8\_224\_dino}. We use a four-layer MLP with ReLU activations with hidden layer sizes of 1024.}
\end{itemize}

\subsection{Details of Benchmark Datasets}
\begin{figure*}[ht]
    \setlength{\abovecaptionskip}{ 1 pt}
    \setlength{\belowcaptionskip}{ -2 pt}
    \centering
       \includegraphics[width=0.6\linewidth]{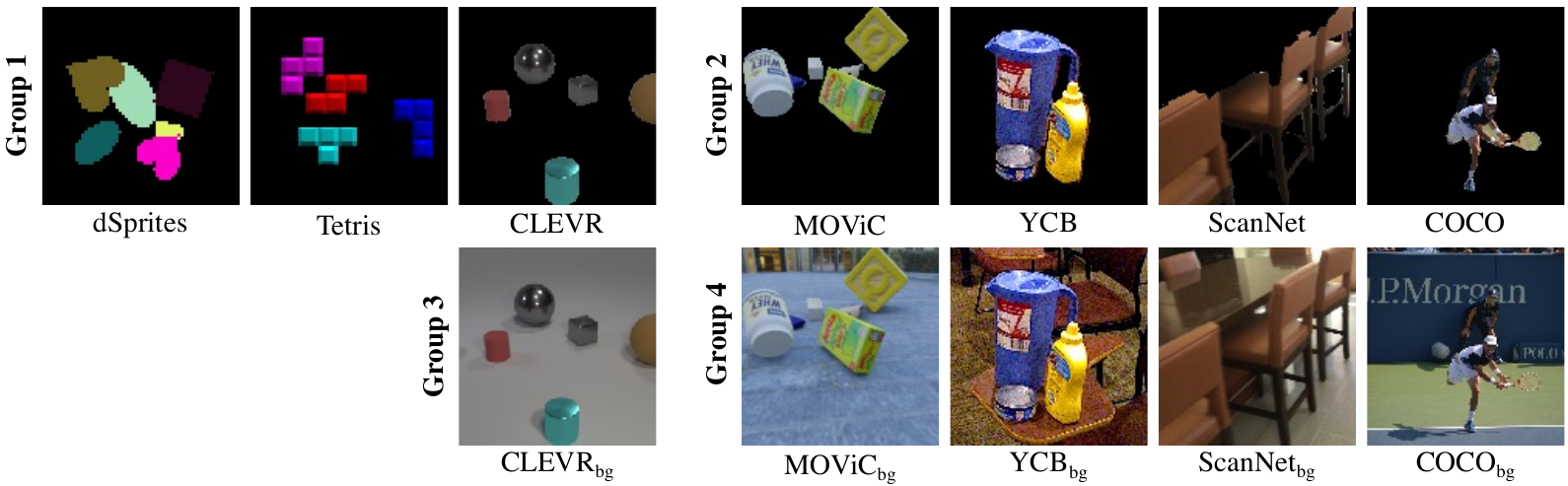}
    \caption{\hl{Samples from all datasets considered.}}
    \label{fig:dataset_sample_images}
\end{figure*}

In this section, we present the details of synthetic and real-world datasets. \hl{Figure} \ref{fig:dataset_sample_images} \hl{presents sample images from all four dataset groups.
Table} \ref{tab:dataset_biases} \hl{qualitatively summarizes the biases of datasets in Groups 1/2/3/4.}

\begin{table*}[th] 
\vspace{0.2cm}
\centering
\setlength{\abovecaptionskip}{ 4 pt}
\setlength{\belowcaptionskip}{ -10 pt}
\caption{\hl{The object-, scene-, and background-level biases in appearance and geometry of the datasets in Groups 1/2/3/4.}}
\label{tab:dataset_biases}
\resizebox{\textwidth}{!}
{
\begin{tabular}{lr|ccc|c|ccc} \hline
 & & \multicolumn{3}{|c|}{Synthetic Datasets} & \multicolumn{1}{c}{\hl{Semi-realistic Datasets}} & \multicolumn{3}{c}{Real-world Datasets} \\ 
 & & dSprites & Tetris   &CLEVR/CLEVR$_{bg}$& \hl{MOViC/MOViC$_{bg}$} & YCB/YCB$_{bg}$ & ScanNet/ScanNet$_{bg}$ & COCO/COCO$_{bg}$ \\ \hline 
\multicolumn{9}{c}{ \multirow{2}{*}{\textit{Object-level Biases}}} \\ 
&&&&&&& \\ \hlineB{3} 
\multirow{2}{*}{Appearance} & Texture: & simple  & simple  & simple & \hl{diverse}  & diverse   & simple  & diverse  \\  
                            & Lighting: & no     & no       & synthetic & \hl{synthetic}  & real  & real  & real  \\ \hdashline
\multirow{2}{*}{Geometry}   & Shape:  & simple  & simple  & simple & \hl{simple}  & simple  & diverse  & diverse  \\ 
                            & Occlusion: & minor & no  & minor & \hl{severe} & severe & severe  & severe \\  \hline
\multicolumn{9}{c}{ \multirow{2}{*}{\textit{Scene-level Biases}}} \\ 
&&&&&&& \\ \hlineB{3} 
Appearance                 & Similarity: & low   & low  & high & \hl{high} & high   & high  & high  \\  \hdashline
Geometry                  & Diversity:  & low   & low    & low & \hl{medium} & high   & high  & high  \\  \hline
\multicolumn{9}{c}{ \multirow{2}{*}{\textit{Background-level Biases}}} \\ 
&&&&&&& \\ \hlineB{3} 
\multirow{3}{*}{Appearance} & Texture: & --&--& -- / simple & \hl{-- / simple} & -- / diverse&-- / simple & -- / diverse \\  
           & Lighting: & --  & -- &  -- / synthetic &  \hl{-- / synthetic} & -- / real& -- / real & -- / real \\ 
            & Similarity: & -- & -- & -- / minor & \hl{-- / high} & -- / high & -- / high & -- / high \\  \hdashline
Geometry    & Diversity:  & -- & -- & -- / simple & \hl{-- / simple} & -- / complex & -- / complex  & -- / complex \\ 
\bottomrule[1.0pt]
\end{tabular}
}\vspace{-0.2cm}
\end{table*}

\textbf{dSprites \citep{Matthey2017}}
To generate a specific image for this dataset, we first sample a random integer $K$ from a uniform distribution with interval $[2, 6]$ as the number of objects in that image. Then, $K$ object shapes are selected from the binary dsprites dataset \citep{Matthey2017} also in a uniformly random manner. Each object is assigned a random RGB color by sampling three random integers from a uniform distribution with interval $[0, 255]$. In total, we generate 10000 images for training, and 2000 for testing.
    
\textbf{Tetris \citep{Kabra2019}}
For each image in this dataset, we first sample a random integer $K$ from a uniform distribution with interval $[2, 6]$ as the number of objects in this image. To render one tetris-like object onto the canvas, we randomly pick up a Tetris object from a randomly selected image from \citep{Kabra2019}. Each object is resized to be $88 \times 88$ and then placed onto the canvas. The position of each object is also sampled from a uniform distribution with 2 criteria: 1) all objects shall be on the canvas with complete shapes; 2) all objects shall not overlap with each other.
    
\textbf{CLEVR \citep{Johnson2017}}
We first generate CLEVR images following {\tiny{\url{https://github.com/facebookresearch/clevr-dataset-gen}}}, where the number of objects per image is restricted between 3 to 6. 
Given generated images with a resolution $640 \times 480$, we perform center-cropping and then resize them to be $128 \times 128$. Then, we remove tiny objects which have less than 35 pixels from each image. Subsequently, the images with less than 2 objects are removed. Being consistent with the previous 2 synthetic datasets, all images have a black background.

\hl{\textbf{MOViC \mbox{\citep{greff2021kubric}}} We use the training and validation sets of the released MOVi dataset from \mbox{\tiny{\url{gs://kubric-public/tfds}}}. Specifically, the downsampled variant at a resolution of 128x128 of MOVi-C is used. In the original MOVi-C dataset, each image consists of $3\sim 10$ objects. Only images with $2\sim 6$ objects are kept. All background pixels are replaced by the black color.}

\textbf{YCB \citep{Calli2017}}
We sample single frames from the YCB video dataset \citep{Calli2017} every 20 images. Given the sampled frames with a resolution of $640 \times 480$, we first center crop and then resize them to be $128 \times 128$. Then, the images consisting of less than 2 or more than 6 objects are removed. Similarly, all background pixels are replaced by a black color.
    
\textbf{ScanNet \citep{Dai2017}}
We sample single frames from the ScanNet dataset \citep{Dai2017} every 20 images. Given the selected frames with a resolution of $1296 \times 968$, we first center crop the images with a size of $800 \times 800$ and then resize them to be $128 \times 128$. For each resized image, we remove objects that contain more than $128 \times 128 \times 0.2$ pixels or less than $128 \times 128 \times 0.007$ pixels. The images with less than 2 or more than 6 are also dropped. All background pixels are replaced by the black color.
    
\textbf{COCO \citep{Lin2014}}
Given images in COCO-2017 \citep{Lin2014} with various resolutions, we first center crop and then resize images to be $128 \times 128$. For each resized image, we use the same criteria applied to ScanNet to remove too-large and too-small objects. The images with $2\sim 6$ objects are kept.
All background pixels are replaced by the black color. Since the number of images that could meet all requirements is less than 2,000 (only 1,597) in the official validation split, we additionally select 403 different images from its official training split for testing. There is no overlap between our training and testing splits.

\textbf{CLEVR$_{bg}$, \hl{MOViC$_{bg}$}/ YCB$_{bg}$/ScanNet$_{bg}$/COCO$_{bg}$} These five datasets are created in the same way as the above (CLEVR, \hl{MOViC/} YCB/ ScanNet/ COCO) datasets, with synthetic or real-world backgrounds being added back to each image.

\subsection{\hl{Examples of the ablated datasets}}
\begin{figure*}[h]
    \setlength{\abovecaptionskip}{ 1 pt}
    \setlength{\belowcaptionskip}{ -6 pt}
    \centering
       \includegraphics[width=0.7\linewidth]{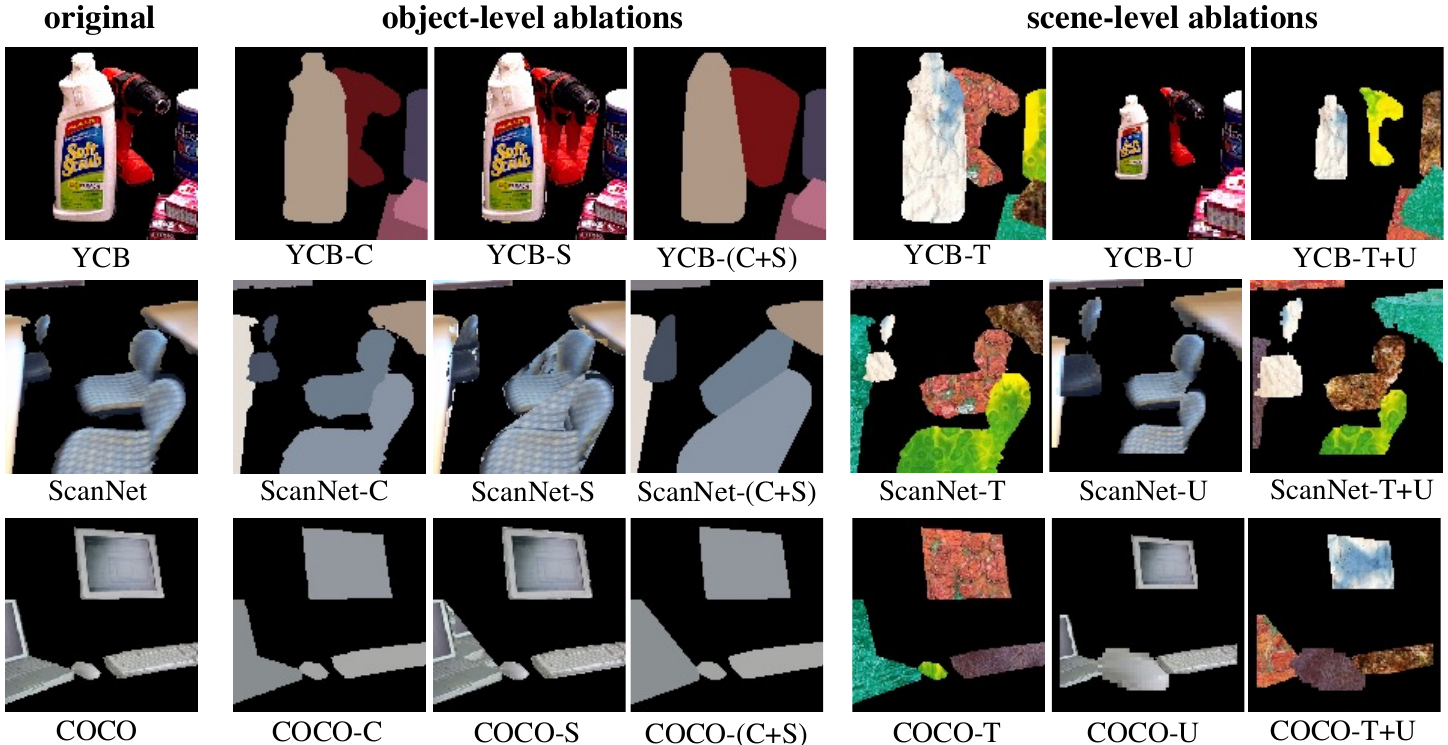}
    \caption{\hl{Example images of real-world datasets ablated with object-level and scene-level factors.}}
    \label{fig:ablation_example_1}
\end{figure*}

\begin{figure*}[h]
    \setlength{\abovecaptionskip}{ 1 pt}
    \setlength{\belowcaptionskip}{ -6 pt}
    \centering
       \includegraphics[width=0.85\linewidth]{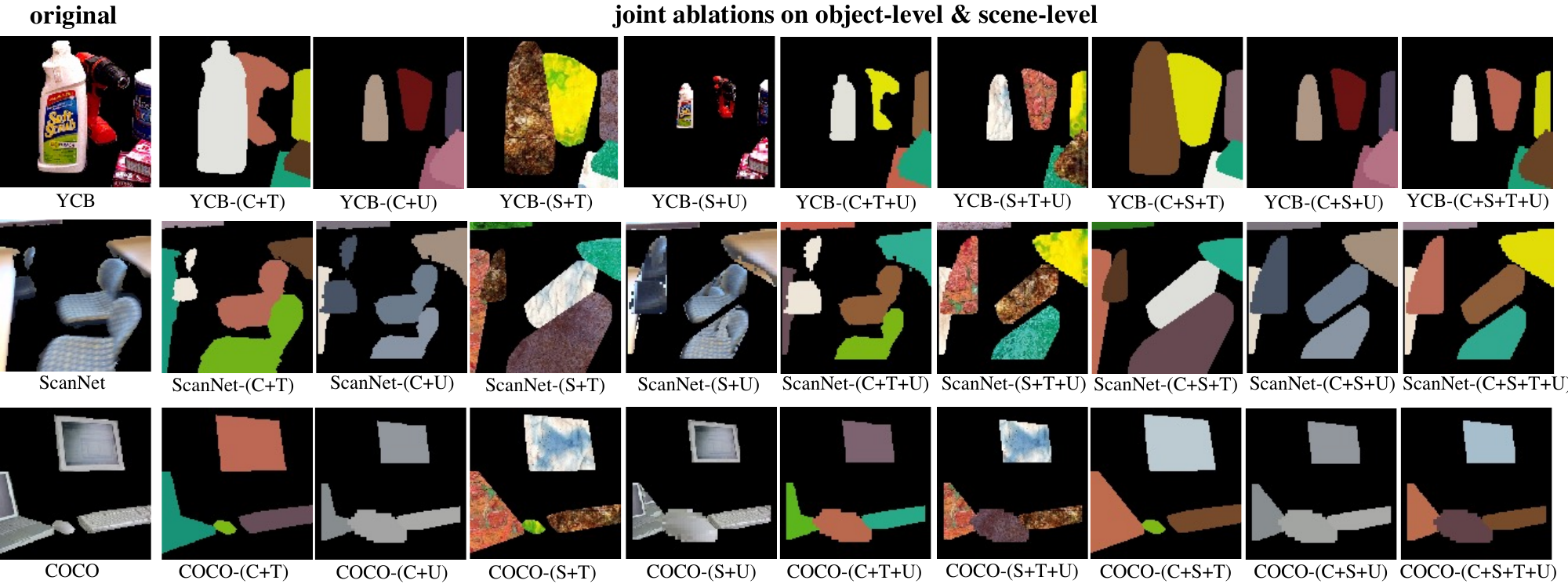}
    \caption{\hl{Example images of real-world datasets ablated with object-level and scene-level factors (continued).}}
    \label{fig:ablation_example_2}
\end{figure*}

\hl{Figure} \ref{fig:ablation_example_1} \hl{presents before/after view for images ablated with object-level factors and scene-level factors respectively. Figure }\ref{fig:ablation_example_2} \hl{shows examples of the joint ablations on both object-level and scene-level.}

\subsection{Details of Experimental Results in Section \ref{sec:exp_main_bg}}
Tables \ref{tab:main_res_nobg}\&\ref{tab:main_res_bg} present the detailed results of experiments conducted in Section \ref{sec:exp_main_bg}. Standard deviations for all scores are calculated over 3 independent runs.

\begin{table*}[th]
\centering
\caption{Quantitative results of object segmentation from the five methods on datasets of Groups 1\&2 with blank backgrounds. This table corresponds to Figures \ref{fig:main_exp}\&\ref{fig:main_exp_vis}. Standard deviations are calculated over 3 runs (marked with \textcolor{blue}{{blue}}{).}}
\label{tab:main_res_nobg}
\resizebox{\textwidth}{!}{
  \begin{tabular}{r c|c|c} \toprule
  & dSprites  & Tetris  & CLEVR   \\ \midrule
              & AP / PQ / Pre / Rec  & AP / PQ / Pre / Rec & AP / PQ / Pre / Rec  \\ 
AIR & 45.4 \textcolor{blue}{1.8} / 38.2 \textcolor{blue}{3.0} / 57.6 \textcolor{blue}{7.4} / 58.1 \textcolor{blue}{7.5} & 25.2 \textcolor{blue}{13.9} / 23.4 \textcolor{blue}{12.4} / 36.8 \textcolor{blue}{20.9} / 39.9 \textcolor{blue}{12.9} & 46.4 \textcolor{blue}{14.0} / 44.3 \textcolor{blue}{12.4} / 67.4 \textcolor{blue}{9.9} / 52.5 \textcolor{blue}{15.9} \\ 
MONet & 69.7 \textcolor{blue}{4.1} / 61.6 \textcolor{blue}{6.0} / 70.4 \textcolor{blue}{8.1} / 73.9 \textcolor{blue}{1.9} & 85.9 \textcolor{blue}{13.0} / 75.8 \textcolor{blue}{13.6} / 85.1 \textcolor{blue}{16.4}/ 89.7  \textcolor{blue}{8.2} & 39.0 \textcolor{blue}{8.5} / 37.3 \textcolor{blue}{6.3} / 65.6 \textcolor{blue}{11.8} / 42.8 \textcolor{blue}{10.8}  \\ 
IODINE & 92.9  \textcolor{blue}{4.3}/ 71.3  \textcolor{blue}{6.1}/ 82.6  \textcolor{blue}{2.3}/ 96.0  \textcolor{blue}{5.2} & 52.2 \textcolor{blue}{2.3} / 37.9 \textcolor{blue}{4.6} / 48.0 \textcolor{blue}{2.3} / 61.8 \textcolor{blue}{1.7} & 82.8 \textcolor{blue}{2.8} / 73.0 \textcolor{blue}{5.7} / 77.5 \textcolor{blue}{3.1} / 87.4 \textcolor{blue}{2.0} \\ 
SlotAtt & 92.8  \textcolor{blue}{1.4} / 82.8  \textcolor{blue}{1.6} / 88.8 \textcolor{blue}{3.4} / 92.9  \textcolor{blue}{1.6} & 94.3 \textcolor{blue}{1.2} / 79.9 \textcolor{blue}{6.4} / 90.5 \textcolor{blue}{3.3} / 94.4 \textcolor{blue}{1.3} & 91.7 \textcolor{blue}{6.4} / 82.9 \textcolor{blue}{10.9} / 90.8 \textcolor{blue}{9.7} / 92.7 \textcolor{blue}{5.3} \\ 
MRCNN & 98.4 \textcolor{blue}{1.5} / 90.2 \textcolor{blue}{2.4} / 99.6 \textcolor{blue}{0.6} / 98.4 \textcolor{blue}{1.4} & 99.8 \textcolor{blue}{0.0} / 90.3 \textcolor{blue}{0.6} / 99.8 \textcolor{blue}{0.2} / 99.8 \textcolor{blue}{0.0} & 98.2 \textcolor{blue}{1.1} / 90.0 \textcolor{blue}{1.4} / 97.8 \textcolor{blue}{0.4} / 99.5 \textcolor{blue}{1.3} \\
\midrule
& YCB  & ScanNet  & COCO   \\ \midrule
              & AP / PQ / Pre / Rec  & AP / PQ / Pre / Rec & AP / PQ / Pre / Rec  \\ 
AIR & 0.0 \textcolor{blue}{0.1} / 0.6 \textcolor{blue}{0.3} / 1.1   \textcolor{blue}{0.4} / 0.8 \textcolor{blue}{0.2} & 2.7 \textcolor{blue}{1.4} / 6.3 \textcolor{blue}{1.7} / 15.6 \textcolor{blue}{2.8} / 7.3 \textcolor{blue}{1.6} & 2.7 \textcolor{blue}{0.1} / 6.7 \textcolor{blue}{0.5} / 14.3 \textcolor{blue}{2.6} / 8.6 \textcolor{blue}{0.8} \\
MONet & 3.1 \textcolor{blue}{1.6} / 7.0 \textcolor{blue}{2.6} / 9.8 \textcolor{blue}{3.6} / 1.2 \textcolor{blue}{0.8} & 24.8 \textcolor{blue}{1.6} / 24.6 \textcolor{blue}{1.6} / 31.0 \textcolor{blue}{1.6} / 40.7 \textcolor{blue}{1.8} & 11.8 \textcolor{blue}{2.0} / 12.5 \textcolor{blue}{1.1}/ 16.1 \textcolor{blue}{0.9}/ 21.9  \textcolor{blue}{1.7} \\
IODINE & 1.8 \textcolor{blue}{0.2} / 3.9 \textcolor{blue}{1.3} / 6.2 \textcolor{blue}{2.0} / 7.3 \textcolor{blue}{1.9} & 10.1 \textcolor{blue}{2.9} / 13.7 \textcolor{blue}{2.7} / 18.6 \textcolor{blue}{4.2} / 24.4 \textcolor{blue}{3.8} & 4.0 \textcolor{blue}{1.2} / 6.3 \textcolor{blue}{1.2} / 9.9 \textcolor{blue}{1.8} / 10.8 \textcolor{blue}{2.0} \\ 
SlotAtt & 9.2 \textcolor{blue}{0.4} / 13.5 \textcolor{blue}{0.9} / 20.0 \textcolor{blue}{1.3} / 26.2 \textcolor{blue}{6.8} & 5.7 \textcolor{blue}{0.3} / 9.0 \textcolor{blue}{1.5} / 12.4 \textcolor{blue}{2.5} / 18.3 \textcolor{blue}{2.7} & 0.8 \textcolor{blue}{0.3} / 3.5 \textcolor{blue}{1.2} / 5.3 \textcolor{blue}{1.7} / 7.3 \textcolor{blue}{2.2} \\ 
MRCNN & 62.9 \textcolor{blue}{4.0} / 58.4 \textcolor{blue}{5.9} / 83.3 \textcolor{blue}{7.2} / 66.9 \textcolor{blue}{3.3} & 41.4 \textcolor{blue}{4.3} / 43.3 \textcolor{blue}{6.2}/ 65.2 \textcolor{blue}{3.5} / 50.5 \textcolor{blue}{1.8} & 46.0 \textcolor{blue}{0.8} / 47.9 \textcolor{blue}{2.1} / 71.7 \textcolor{blue}{4.0} / 53.2 \textcolor{blue}{0.4}\\ 
\bottomrule
  \end{tabular}
}
\end{table*}

\begin{table*}[th]
\centering
\caption{Quantitative results of object segmentation from three methods on datasets of Groups 3\&4 with blank backgrounds. This table corresponds to Figures \ref{fig:main_exp_bg}\&\ref{fig:main_exp_bg_vis}. Standard deviations are calculated over 3 runs (marked with \textcolor{blue}{{blue}}{).}}
\label{tab:main_res_bg}
\resizebox{\textwidth}{!}{
  \begin{tabular}{r c|c|c} \toprule
&  & CLEVR$_{bg}$ & \\ \midrule
&  & AP / PQ / Pre / Rec / BG-Rec &    \\ 
IODINE & & 89.0 \textcolor{blue}{6.1} / 67.5 \textcolor{blue}{4.9} / 86.6 \textcolor{blue}{13.5} / 90.5 \textcolor{blue}{5.5} / 99.9 \textcolor{blue}{0.0}  \\
SlotAtt & & 92.6 \textcolor{blue}{6.2} / 81.5 \textcolor{blue}{8.3} / 90.9 \textcolor{blue}{4.0} / 93.6 \textcolor{blue}{6.1} / 100 \textcolor{blue}{0.0}\\
MRCNN & & 97.2 \textcolor{blue}{1.0} / 85.8 \textcolor{blue}{4.2}/ 96.6 \textcolor{blue}{1.1} / 98.4 \textcolor{blue}{1.0} / 100.0 \textcolor{blue}{0.0}\\
\midrule
& YCB$_{bg}$  & ScanNet$_{bg}$  & COCO$_{bg}$   \\ \midrule
              & AP / PQ / Pre / Rec / BG-Rec & AP / PQ / Pre / Rec / BG-Rec & AP / PQ / Pre / Rec / BG-Rec  \\ 

IODINE & 0.1 \textcolor{blue}{0.4}/1.2 \textcolor{blue}{1.9}/1.8 \textcolor{blue}{0.6}/2.6 \textcolor{blue}{1.7}/8.2 \textcolor{blue}{2.9} & 0.8 \textcolor{blue}{0.3}/3.9 \textcolor{blue}{0.5}/5.4 \textcolor{blue}{0.5}/7.8 \textcolor{blue}{1.6}/26.5 \textcolor{blue}{4.6} & 0.1 \textcolor{blue}{0.1}/1.0 \textcolor{blue}{0.8}/1.3 \textcolor{blue}{1.0}/2.2 \textcolor{blue}{1.9}/10.1 \textcolor{blue}{3.5} \\
SlotAtt & 0.0 \textcolor{blue}{0.0}/ 0.9 \textcolor{blue}{0.1}/1.4 \textcolor{blue}{0.2}/1.9 \textcolor{blue}{0.3}/0.3 \textcolor{blue}{0.2} & 1.8 \textcolor{blue}{0.5}/6.2 \textcolor{blue}{0.4}/8.1 \textcolor{blue}{0.5}/12.9 \textcolor{blue}{0.7}/0.0 \textcolor{blue}{1.3} & 0.0 \textcolor{blue}{0.0}/0.3 \textcolor{blue}{0.0}/0.4 \textcolor{blue}{0.0}/0.7 \textcolor{blue}{0.1}/0.1 \textcolor{blue}{0.1}\\
MRCNN & 42.6 \textcolor{blue}{2.1}/40.1 \textcolor{blue}{0.2}/61.8 \textcolor{blue}{4.5}/ 47.9 \textcolor{blue}{2.5}/100 \textcolor{blue}{0.2}& 14.4 \textcolor{blue}{0.9}/ 19.4 \textcolor{blue}{0.3}/28.1 \textcolor{blue}{0.2}/28.8 \textcolor{blue}{0.7}/ 98.1 \textcolor{blue}{0.0}& 11.3 \textcolor{blue}{0.2}/15.3 \textcolor{blue}{0.0}/21.2 \textcolor{blue}{1.0}/24.5 \textcolor{blue}{0.6}/99.3 \textcolor{blue}{0.3}\\
\bottomrule
\end{tabular}
}
\end{table*}

\subsection{Details of Experimental Results in Sections \ref{sec:exp_object_factors}\&\ref{sec:exp_scene_factors}\&\ref{sec:exp_joint_factors}\&\ref{sec:exp_background_factor}}

Tables \ref{tab:app_S_C_experiments}\&\ref{tab:app_T_U_experiments}\&\ref{tab:app_hybrid_ablation_1}\&\ref{tab:app_bg_experiments} present the detailed results of experiments conducted in Sections \ref{sec:exp_object_factors}\&\ref{sec:exp_scene_factors}\& \ref{sec:exp_joint_factors}\& \ref{sec:exp_background_factor}. Standard deviations for all scores are calculated over 3 independent runs.

\begin{table*}[th]
\centering
\caption{\small{Quantitative results on datasets ablated with object-level factors. Standard deviations are calculated over 3 runs (marked with \textcolor{blue}{blue}). This table corresponds to Section \ref{sec:exp_object_factors}.}}
\label{tab:app_S_C_experiments}
\resizebox{\textwidth}{!}{
  \begin{tabular}{r c|c|c} \toprule
      & YCB-C & ScanNet-C & COCO-C  \\ \midrule
              & AP / PQ / Pre / Rec  & AP / PQ / Pre / Rec & AP / PQ / Pre / Rec \\ 
AIR & 4.4 \textcolor{blue}{0.2} / 1.0 \textcolor{blue}{7.3} / 20.7 \textcolor{blue}{7.1} / 12.5 \textcolor{blue}{1.2} & 2.7 \textcolor{blue}{1.0} / 7.6 \textcolor{blue}{2.1} / 29.1 \textcolor{blue}{20.2} / 7.4 \textcolor{blue}{1.5} & 2.9 \textcolor{blue}{1.8} / 7.9 \textcolor{blue}{2.5} / 32.0 \textcolor{blue}{29.3} / 7.7 \textcolor{blue}{4.5} \\ 
MONet & 55.1 \textcolor{blue}{6.0} / 49.5 \textcolor{blue}{5.2} / 64.8 \textcolor{blue}{1.7} / 58.8 \textcolor{blue}{7.8} & 31.1 \textcolor{blue}{15.9} / 33.1 \textcolor{blue}{9.6} / 41.6 \textcolor{blue}{10.3} / 45.6 \textcolor{blue}{11.1} & 34.5 \textcolor{blue}{7.2} / 34.6 \textcolor{blue}{3.3} / 45.8 \textcolor{blue}{2.0} / 42.6 \textcolor{blue}{11.0}\\ 
IODINE & 76.8 \textcolor{blue}{0.3} / 64.4 \textcolor{blue}{0.1} / 76.5 \textcolor{blue}{0.1} / 79.5 \textcolor{blue}{0.0} & 64.4 \textcolor{blue}{6.4} / 54.3 \textcolor{blue}{5.9} / 66.3 \textcolor{blue}{4.6} / 71.0 \textcolor{blue}{5.9} & 42.8 \textcolor{blue}{14.7} / 34.0 \textcolor{blue}{11.8} / 42.5 \textcolor{blue}{14.4} / 55.8 \textcolor{blue}{11.9}\\ 
SlotAtt & 39.2 \textcolor{blue}{1.9} / 30.2 \textcolor{blue}{0.0} / 40.2 \textcolor{blue}{0.8} / 50.4 \textcolor{blue}{0.7} & 5.7 \textcolor{blue}{8.2} / 9.0 \textcolor{blue}{6.1} / 12.4 \textcolor{blue}{7.6} / 18.3 \textcolor{blue}{8.9} & 7.2 \textcolor{blue}{1.9} / 9.8  \textcolor{blue}{1.7}/ 13.9 \textcolor{blue}{2.2} / 18.7 \textcolor{blue}{3.4}\\ 
\midrule
& YCB-S  & ScanNet-S  & COCO-S   \\ \midrule
              & AP / PQ / Pre / Rec  & AP / PQ / Pre / Rec & AP / PQ / Pre / Rec  \\
AIR & 3.6 \textcolor{blue}{3.0} / 6.8 \textcolor{blue}{3.2} / 12.5 \textcolor{blue}{6.7} / 9.7 \textcolor{blue}{2.7} & 3.3 \textcolor{blue}{2.5} / 8.5 \textcolor{blue}{1.1} / 36.6 \textcolor{blue}{23.0} / 8.1 \textcolor{blue}{8.5} & 3.3 \textcolor{blue}{0.0} / 7.1 \textcolor{blue}{1.1} / 16.8 \textcolor{blue}{5.6} / 8.3 \textcolor{blue}{1.0}\\ 
MONet & 2.5 \textcolor{blue}{1.1} / 6.4 \textcolor{blue}{1.8} / 8.9 \textcolor{blue}{2.6} / 11.3 \textcolor{blue}{3.2} & 18.3 \textcolor{blue}{10.4} / 22.2 \textcolor{blue}{3.9} / 28.4 \textcolor{blue}{4.4} / 37.5 \textcolor{blue}{5.3} & 8.7 \textcolor{blue}{3.2} / 11.2 \textcolor{blue}{1.7} / 14.4 \textcolor{blue}{3.0} / 21.0 \textcolor{blue}{0.7}\\ 
IODINE & 2.5 \textcolor{blue}{2.5} / 5.0 \textcolor{blue}{5.0} / 7.9 \textcolor{blue}{7.8} / 1.1 \textcolor{blue}{1.0} & 8.3 \textcolor{blue}{1.1} / 13.7 \textcolor{blue}{0.4} / 18.4 \textcolor{blue}{0.4} / 24.8 \textcolor{blue}{0.7} & 4.2 \textcolor{blue}{0.4} / 8.3 \textcolor{blue}{0.4} / 12.1 \textcolor{blue}{0.3} / 14.8 \textcolor{blue}{0.6} \\ 
SlotAtt & 19.1 \textcolor{blue}{0.7} / 21.0 \textcolor{blue}{1.7} / 30.4 \textcolor{blue}{2.8} / 37.7 \textcolor{blue}{0.8} & 1.9 \textcolor{blue}{8.5} / 6.6 \textcolor{blue}{7.5} / 8.7 \textcolor{blue}{9.8} / 13.7 \textcolor{blue}{11.9} & 2.6 \textcolor{blue}{1.0} / 5.9 \textcolor{blue}{0.9} / 8.3 \textcolor{blue}{1.3} / 11.9 \textcolor{blue}{10.9}\\ 
\midrule
& YCB-C+S  & ScanNet-C+S  & COCO-C+S   \\ \midrule
              & AP / PQ / Pre / Rec  & AP / PQ / Pre / Rec & AP / PQ / Pre / Rec  \\ 
AIR & 2.0 \textcolor{blue}{0.3} / 6.4 \textcolor{blue}{0.2} / 15.1 \textcolor{blue}{5.2} / 7.3 \textcolor{blue}{4.1} & 3.5 \textcolor{blue}{1.0} / 9.1 \textcolor{blue}{0.5} / 38.3 \textcolor{blue}{25.8} / 8.5 \textcolor{blue}{5.9} & 4.8 \textcolor{blue}{0.1} / 10.8 \textcolor{blue}{0.1} / 41.5 \textcolor{blue}{10.2} / 10.1 \textcolor{blue}{1.7}\\ 
MONet & 9.1 \textcolor{blue}{13.6} / 12.7 \textcolor{blue}{10.5} / 34.1 \textcolor{blue}{4.6} / 12.6 \textcolor{blue}{24.0} & 48.1 \textcolor{blue}{1.0} / 47.9 \textcolor{blue}{6.8} / 70.4 \textcolor{blue}{23.3} / 51.5 \textcolor{blue}{6.5} & 10.5 \textcolor{blue}{24.3} / 16.5 \textcolor{blue}{18.1} / 56.5 \textcolor{blue}{6.9} / 14.9 \textcolor{blue}{28.2}\\ 
IODINE & 69.9 \textcolor{blue}{30.9} / 55.7 \textcolor{blue}{29.6} / 60.3 \textcolor{blue}{28.3} / 71.5 \textcolor{blue}{26.4} & 73.8 \textcolor{blue}{11.4} / 63.2 \textcolor{blue}{10.1} / 73.5 \textcolor{blue}{6.1} / 76.8 \textcolor{blue}{10.8} & 73.5 \textcolor{blue}{2.5} / 61.2 \textcolor{blue}{2.4} / 70.0 \textcolor{blue}{2.9} / 77.6 \textcolor{blue}{1.2} \\ 
SlotAtt & 39.1 \textcolor{blue}{22.3} / 30.2 \textcolor{blue}{13.3} / 42.9 \textcolor{blue}{12.2} / 54.5 \textcolor{blue}{12.7} & 22.5 \textcolor{blue}{7.6} / 21.5 \textcolor{blue}{6.2} / 28.8 \textcolor{blue}{6.0} / 39.6 \textcolor{blue}{5.4} & 20.3 \textcolor{blue}{3.0} / 19.9 \textcolor{blue}{0.4} / 26.1 \textcolor{blue}{0.4} / 34.8 \textcolor{blue}{0.7} \\ \bottomrule
  \end{tabular}}
  \vspace{-0.2cm}
\end{table*}

\begin{table*}[th]
\centering
\caption{Quantitative results on datasets ablated with scene-level factors. Standard deviations are calculated over 3 runs (marked with \textcolor{blue}{blue}). This table corresponds to Section \ref{sec:exp_scene_factors}. }
\label{tab:app_T_U_experiments}
\resizebox{\textwidth}{!}{
  \begin{tabular}{r c|c|c} \toprule
      & YCB-T & ScanNet-T & COCO-T  \\ \midrule
              & AP / PQ / Pre / Rec  & AP / PQ / Pre / Rec & AP / PQ / Pre / Rec \\ 
AIR & 5.9 \textcolor{blue}{3.7} / 9.6 \textcolor{blue}{3.3} / 19.9 \textcolor{blue}{10.3} / 12.7 \textcolor{blue}{1.3} & 2.9 \textcolor{blue}{1.2} / 6.3 \textcolor{blue}{0.1} / 13.4 \textcolor{blue}{3.7} / 8.6 \textcolor{blue}{3.2} & 7.4 \textcolor{blue}{3.1} / 13.0 \textcolor{blue}{4.0} / 29.1 \textcolor{blue}{9.8} / 16.4 \textcolor{blue}{4.3} \\ 
MONet & 86.5 \textcolor{blue}{12.4} / 78.3 \textcolor{blue}{13.3} / 81.1 \textcolor{blue}{11.6} / 86.8 \textcolor{blue}{12.2} & 82.9 \textcolor{blue}{12.3} / 77.8 \textcolor{blue}{7.6} / 86.3 \textcolor{blue}{1.6} / 83.7 \textcolor{blue}{11.9} & 80.0 \textcolor{blue}{9.4} / 68.8 \textcolor{blue}{13.5} / 74.5 \textcolor{blue}{15.0} / 82.0 \textcolor{blue}{5.6} \\ 
IODINE & 32.4 \textcolor{blue}{9.0} / 27.3 \textcolor{blue}{6.8} / 35.3 \textcolor{blue}{8.3} / 43.6 \textcolor{blue}{10.6} & 33.2 \textcolor{blue}{11.9} / 27.3 \textcolor{blue}{6.3} / 34.5 \textcolor{blue}{7.0} / 44.7 \textcolor{blue}{9.3} & 40.8 \textcolor{blue}{12.7} / 33.7 \textcolor{blue}{10.2} / 41.8 \textbf{\textcolor{blue}{10.4}} / 55.9 \textcolor{blue}{15.8} \\ 
SlotAtt & 64.6 \textcolor{blue}{5.3} / 48.5 \textcolor{blue}{3.2} / 58.3 \textcolor{blue}{3.5} / 74.9\textcolor{blue}{4.6} & 20.5 \textcolor{blue}{22.3} / 18.2 \textcolor{blue}{20.1} / 22.9 \textcolor{blue}{23.1} / 33.2 \textcolor{blue}{31.4} & 31.8 \textcolor{blue}{17.8} / 28.0 \textcolor{blue}{11.2} / 35.2 \textcolor{blue}{12.4} / 50.1 \textcolor{blue}{19.8}\\ 
\midrule
& YCB-U  & ScanNet-U  & COCO-U   \\ \midrule
              & AP / PQ / Pre / Rec  & AP / PQ / Pre / Rec & AP / PQ / Pre / Rec  \\
AIR & 9.8 \textcolor{blue}{3.5} / 12.2 \textcolor{blue}{2.6} / 18.6 \textcolor{blue}{4.1} / 19.9 \textcolor{blue}{2.8} & 7.1 \textcolor{blue}{1.4} / 10.4 \textcolor{blue}{0.2} / 19.9 \textcolor{blue}{2.1} / 14.0 \textcolor{blue}{1.0} & 12.9 \textcolor{blue}{4.8} / 16.3 \textcolor{blue}{4.6} / 28.6 \textcolor{blue}{1.8} / 24.3 \textcolor{blue}{9.3} \\ 
MONet & 3.7 \textcolor{blue}{0.5} / 8.3 \textcolor{blue}{1.2} / 12.2 \textcolor{blue}{1.4} / 13.2 \textcolor{blue}{1.4} & 25.4 \textcolor{blue}{8.3} / 23.6 \textcolor{blue}{7.4} / 29.2 \textcolor{blue}{8.9} / 37.2 \textcolor{blue}{12.5} & 14.5 \textcolor{blue}{0.6} / 16.5 \textcolor{blue}{0.8} / 27.9 \textcolor{blue}{1.4} / 20.9 \textcolor{blue}{0.8} \\ 
IODINE & 2.5 \textcolor{blue}{1.4} / 4.9 \textcolor{blue}{2.0} / 7.6 \textcolor{blue}{2.9} / 9.3 \textcolor{blue}{3.6} & 16.9 \textcolor{blue}{0.8} / 17.2 \textcolor{blue}{0.0} / 23.8 \textcolor{blue}{0.8} / 30.2 \textcolor{blue}{0.1} & 5.5 \textcolor{blue}{1.0} / 7.7 \textcolor{blue}{0.0} / 11.7 \textcolor{blue}{0.6} / 13.5 \textcolor{blue}{0.3} \\ 
SlotAtt & 28.9 \textcolor{blue}{2.0} / 21.9 \textcolor{blue}{1.2} / 30.7 \textcolor{blue}{3.1} / 37.3 \textcolor{blue}{2.4} & 9.4 \textcolor{blue}{4.8} / 10.4 \textcolor{blue}{3.8} / 15.4 \textcolor{blue}{4.6} / 18.1 \textcolor{blue}{7.2} & 4.4 \textcolor{blue}{1.1} / 7.9 \textcolor{blue}{0.3} / 11.9 \textcolor{blue}{0.9} / 14.7 \textcolor{blue}{0.7} \\ 
\midrule
& YCB-T+U  & ScanNet-T+U  & COCO-T+U   \\ \midrule
              & AP / PQ / Pre / Rec  & AP / PQ / Pre / Rec & AP / PQ / Pre / Rec  \\ 
AIR & 18.9 \textcolor{blue}{13.7} / 21.0 \textcolor{blue}{9.9} / 32.5 \textcolor{blue}{15.7} / 32.6 \textcolor{blue}{13.2} & 12.3 \textcolor{blue}{5.4} / 16.8 \textcolor{blue}{5.6} / 37.4 \textcolor{blue}{20.8} / 20.4 \textcolor{blue}{1.1} & 25.3 \textcolor{blue}{13.0} / 28.3 \textcolor{blue}{9.0} / 49.4 \textcolor{blue}{13.2} / 38.2 \textcolor{blue}{12.4} \\ 
MONet & 89.1 \textcolor{blue}{0.8} / 84.7 \textcolor{blue}{1.5} / 91.7 \textcolor{blue}{4.1} / 89.3 \textcolor{blue}{0.8} & 87.0 \textcolor{blue}{8.9} / 77.5 \textcolor{blue}{10.3} / 80.5 \textcolor{blue}{11.1} / 87.4 \textcolor{blue}{8.7} & 88.2 \textcolor{blue}{2.9} / 76.2 \textcolor{blue}{0.1} / 79.9 \textcolor{blue}{3.7} / 89.0 \textcolor{blue}{2.9} \\ 
IODINE & 35.3 \textcolor{blue}{1.2} / 26.4 \textcolor{blue}{0.3} / 34.9 \textcolor{blue}{0.6} / 42.8 \textcolor{blue}{0.2} & 27.4 \textcolor{blue}{0.8} / 25.0 \textcolor{blue}{0.8} / 32.1 \textcolor{blue}{0.5} / 41.8 \textcolor{blue}{1.8} & 53.3 \textcolor{blue}{16.7} / 35.6 \textcolor{blue}{8.3} / 45.0 \textcolor{blue}{9.7} / 59.8 \textcolor{blue}{13.0} \\ 
SlotAtt & 76.8 \textcolor{blue}{1.2} / 57.1 \textcolor{blue}{3.8} / 72.4 \textcolor{blue}{3.4} / 77.7 \textcolor{blue}{1.8} & 47.3 \textcolor{blue}{11.2} / 33.8 \textcolor{blue}{9.2} / 42.6 \textcolor{blue}{9.1} / 57.3 \textcolor{blue}{8.4} & 34.5 \textcolor{blue}{17.8} / 25.3 \textcolor{blue}{7.6} / 33.7 \textcolor{blue}{8.3} / 43.0 \textcolor{blue}{14.1}\\ \bottomrule
  \end{tabular}}
\end{table*}

\begin{table*}[th]
\centering
\caption{\small{Quantitative results on datasets ablated with both object- and scene-level factors. Standard deviations are calculated over 3 runs (marked with \textcolor{blue}{blue}). This table corresponds to Section \ref{sec:exp_joint_factors}.}}
\label{tab:app_hybrid_ablation_1}
\resizebox{\textwidth}{!}{
  \begin{tabular}{r c|c|c} \toprule
      & YCB-C+S+T & ScanNet-C+S+T & COCO-C+S+T  \\ \midrule
              & AP / PQ / Pre / Rec  & AP / PQ / Pre / Rec & AP / PQ / Pre / Rec \\ 
AIR & 6.3 \textcolor{blue}{3.7} / 10.7 \textcolor{blue}{3.3} / 24.0 \textcolor{blue}{13.2} / 13.1 \textcolor{blue}{0.5} & 1.9 \textcolor{blue}{3.9} / 6.4 \textcolor{blue}{5.2} / 27.6 \textcolor{blue}{10.7} / 6.3 \textcolor{blue}{13.9} & 10.8 \textcolor{blue}{5.6} / 18.6 \textcolor{blue}{7.0} / 45.1 \textcolor{blue}{14.4} / 20.3 \textcolor{blue}{6.9} \\ 
MONet & 67.7 \textcolor{blue}{4.2} / 55.0 \textcolor{blue}{8.5} / 62.7 \textcolor{blue}{19.8} / 75.4 \textcolor{blue}{10.2} & 76.1 \textcolor{blue}{7.4} / 61.7 \textcolor{blue}{2.6} / 64.4 \textcolor{blue}{2.1} / 77.1 \textcolor{blue}{2.8} & 67.1 \textcolor{blue}{15.3} / 56.4 \textcolor{blue}{10.1} / 62.2 \textcolor{blue}{6.2} / 73.9 \textcolor{blue}{17.3} \\ 
IODINE & 82.6 \textcolor{blue}{2.6} / 63.8 \textcolor{blue}{0.9} / 75.5 \textcolor{blue}{5.1} / 86.0 \textcolor{blue}{1.6} & 69.0 \textcolor{blue}{14.9} / 53.5 \textcolor{blue}{14.1} / 60.7 \textcolor{blue}{14.5} / 77.3 \textcolor{blue}{9.4} & 77.6 \textcolor{blue}{7.8} / 59.8 \textcolor{blue}{6.8} / 72.5 \textcolor{blue}{1.8} / 83.0 \textcolor{blue}{7.5} \\ 
SlotAtt & 65.9 \textcolor{blue}{23.2} / 51.6 \textcolor{blue}{18.5} / 62.2 \textcolor{blue}{19.8} / 72.9 \textcolor{blue}{17.6} & 60.2 \textcolor{blue}{8.7} / 45.5 \textcolor{blue}{8.1} / 52.9 \textcolor{blue}{9.2} / 68.3 \textcolor{blue}{7.3} & 51.8 \textcolor{blue}{11.0} / 41.5 \textcolor{blue}{7.7} / 48.6 \textcolor{blue}{7.5} / 62.9 \textcolor{blue}{8.5} \\ 
\midrule
& YCB-C+S+U  & ScanNet-C+S+U  & COCO-C+S+U   \\ \midrule
              & AP / PQ / Pre / Rec  & AP / PQ / Pre / Rec & AP / PQ / Pre / Rec  \\
AIR & 12.5 \textcolor{blue}{6.4} / 17.7 \textcolor{blue}{7.0} / 27.5 \textcolor{blue}{11.2} / 29.6 \textcolor{blue}{10.5} & 8.2 \textcolor{blue}{1.5} / 11.1 \textcolor{blue}{0.5} / 20.4 \textcolor{blue}{0.8} / 15.0 \textcolor{blue}{1.3} & 23.7 \textcolor{blue}{9.8} / 27.5 \textcolor{blue}{8.0} / 51.3 \textcolor{blue}{9.6} / 32.8 \textcolor{blue}{8.3} \\ 
MONet & 3.5 \textcolor{blue}{2.5} / 4.8 \textcolor{blue}{3.2} / 23.5 \textcolor{blue}{2.6} / 5.3 \textcolor{blue}{6.5} & 60.1 \textcolor{blue}{9,5} / 56.3 \textcolor{blue}{13.3} / 69.7 \textcolor{blue}{20.7} / 64.0 \textcolor{blue}{6.2} & 35.6 \textcolor{blue}{1.4} / 33.0 \textcolor{blue}{3.8} / 43.4 \textcolor{blue}{16.6} / 38.7 \textcolor{blue}{0.1} \\ 
IODINE & 65.3 \textcolor{blue}{9.2} / 55.9 \textcolor{blue}{10.4} / 73.9 \textcolor{blue}{4.6} / 67.7 \textcolor{blue}{8.4} & 64.6 \textcolor{blue}{3.2} / 52.1 \textcolor{blue}{5.1} / 61.3 \textcolor{blue}{7.6} / 68.2 \textcolor{blue}{2.2} & 64.8 \textcolor{blue}{7.3} / 53.4 \textcolor{blue}{6.5} / 65.8 \textcolor{blue}{8.5} / 68.2 \textcolor{blue}{2.5} \\ 
SlotAtt & 58.1 \textcolor{blue}{3.7} / 39.1 \textcolor{blue}{2.8} / 52.4 \textcolor{blue}{2.3} / 62.0 \textcolor{blue}{3.6} & 50.5 \textcolor{blue}{2.9} / 38.2 \textcolor{blue}{3.6} / 51.8 \textcolor{blue}{6.9} / 55.0 \textcolor{blue}{1.3} & 20.2 \textcolor{blue}{5.2} / 18.1 \textcolor{blue}{0.3} / 26.9 \textcolor{blue}{0.1} / 33.4 \textcolor{blue}{0.6}\\ 
\midrule
& YCB-C+S+T+U  & ScanNet-C+S+T+U  & COCO-C+S+T+U   \\ \midrule
              & AP / PQ / Pre / Rec  & AP / PQ / Pre / Rec & AP / PQ / Pre / Rec  \\ 
AIR & 24.8 \textcolor{blue}{20.1} / 23.2 \textcolor{blue}{11.9} / 33.7 \textcolor{blue}{16.8} / 39.1 \textcolor{blue}{20.1} & 10.0 \textcolor{blue}{1.8} / 13.5 \textcolor{blue}{1.8}/ 24.0 \textcolor{blue}{1.4} / 18.2 \textcolor{blue}{7.4} & 36.4 \textcolor{blue}{19.3} / 39.4 \textcolor{blue}{14.5} / 63.1 \textcolor{blue}{16.9} / 49.2 \textcolor{blue}{17.9} \\ 
MONet & 70.8 \textcolor{blue}{9.5} / 79.2 \textcolor{blue}{13.0} / 96.8 \textcolor{blue}{9.4} / 70.8 \textcolor{blue}{8.5} & 79.2 \textcolor{blue}{3.4} / 65.0 \textcolor{blue}{3.6} / 69.1 \textcolor{blue}{4.2} / 82.4 \textcolor{blue}{2.2} & 76.2 \textcolor{blue}{3.0} / 75.9 \textcolor{blue}{3.8} / 90.9 \textcolor{blue}{6.4} / 76.5 \textcolor{blue}{3.3} \\ 
IODINE & 72.4 \textcolor{blue}{4.8} / 54.2 \textcolor{blue}{6.7} / 65.1 \textcolor{blue}{5.8} / 74.3 \textcolor{blue}{5.8} & 76.1 \textcolor{blue}{8.7} / 56.5 \textcolor{blue}{9.2} / 66.9 \textcolor{blue}{7.0} / 80.3 \textcolor{blue}{6.2} & 81.4 \textcolor{blue}{3.4} / 59.3 \textcolor{blue}{5.1} / 70.5 \textcolor{blue}{7.7} / 84.3 \textcolor{blue}{6.7} \\ 
SlotAtt & 92.0 \textcolor{blue}{2.0} / 65.5 \textcolor{blue}{6.1} / 84.4 \textcolor{blue}{6.5} / 92.5 \textcolor{blue}{1.7} & 62.7 \textcolor{blue}{25.4} / 42.6 \textcolor{blue}{28.4} / 50.5 \textcolor{blue}{34.3} / 69.4 \textcolor{blue}{19.0} & 83.7 \textcolor{blue}{4.7} / 60.7 \textcolor{blue}{7.2} / 76.4 \textcolor{blue}{4.6} / 84.0 \textcolor{blue}{5.0}\\ \bottomrule
  \end{tabular}}
\end{table*}

\begin{table*}[t]
\centering
\caption{Quantitative results of IODINE and SlotAtt on datasets of Group 4 and their ablations. This table corresponds to Figure \ref{fig:bg_ablation} in Section \ref{sec:exp_background_factor}. Standard deviations are calculated over 3 runs (marked with \textcolor{blue}{{blue}}{).}}
\label{tab:app_bg_experiments}
\resizebox{\textwidth}{!}{
  \begin{tabular}{r c|c|c} \toprule
      & YCB$_{bg}$-C & ScanNet$_{bg}$-C & COCO$_{bg}$-C  \\ \midrule
              & AP / PQ / Pre / Rec / BG-Rec & AP / PQ / Pre / Rec / BG-Rec & AP / PQ / Pre / Rec / BG-Rec \\ 

IODINE & 3.4 \textcolor{blue}{3.3}/6.2 \textcolor{blue}{4.7}/9.8 \textcolor{blue}{3.7}/10.8 \textcolor{blue}{6.1}/99.9 \textcolor{blue}{4.8} & 13.6 \textcolor{blue}{9.3}/14.3 \textcolor{blue}{6.8}/21.0 \textcolor{blue}{6.5}/22.9 \textcolor{blue}{2.6}/98.2 \textcolor{blue}{4.2} & 1.7 \textcolor{blue}{1.6}/3.5 \textcolor{blue}{2.0}/9.2 \textcolor{blue}{2.1}/4.3 \textcolor{blue}{3.3}/98.3 \textcolor{blue}{2.9}\\
SlotAtt & 1.5 \textcolor{blue}{0.9}/4.5 \textcolor{blue}{1.3}/6.6 \textcolor{blue}{1.8}/9.5 \textcolor{blue}{3.0}/28.7 \textcolor{blue}{1.2} & 5.1 \textcolor{blue}{2.0}/9.7 \textcolor{blue}{1.2}/12.9 \textcolor{blue}{1.7}/18.8 \textcolor{blue}{1.9}/47.4 \textcolor{blue}{14.2} & 8.2 \textcolor{blue}{4.4}/12.3 \textcolor{blue}{4.2}/16.9 \textcolor{blue}{5.4}/23.2 \textcolor{blue}{7.2}/99.7 \textcolor{blue}{3.5} \\

\midrule
& YCB$_{bg}$-T & ScanNet$_{bg}$-T & COCO$_{bg}$-T  \\ \midrule
              & AP / PQ / Pre / Rec / BG-Rec & AP / PQ / Pre / Rec / BG-Rec & AP / PQ / Pre / Rec / BG-Rec \\ 
IODINE & 0.7 \textcolor{blue}{0.6}/3.4\textcolor{blue}{1.7}/5.3 \textcolor{blue}{2.5}/6.6 \textcolor{blue}{2.9}/91.6 \textcolor{blue}{0.1}& 10.6 \textcolor{blue}{0.0}/11.0 \textcolor{blue}{0.3}/15.1 \textcolor{blue}{0.6}/20.3 \textcolor{blue}{0.8}/84.5 \textcolor{blue}{0.6}& 1.2 \textcolor{blue}{0.1}/4.4 \textcolor{blue}{0.2}/6.5 \textcolor{blue}{0.1}/8.4 \textcolor{blue}{0.4}/92.7 \textcolor{blue}{0.2}\\
SlotAtt & 19.8 \textcolor{blue}{2.9}/21.2 \textcolor{blue}{1.5}/29.1 \textcolor{blue}{2.1}/40.1 \textcolor{blue}{2.5}/93.8 \textcolor{blue}{0.2}& 12.9 \textcolor{blue}{3.2}/16.7 \textcolor{blue}{2.5}/21.9 \textcolor{blue}{2.9}/30.1 \textcolor{blue}{3.1}/83.6 \textcolor{blue}{10.3}& 2.1 \textcolor{blue}{3.2}/4.5 \textcolor{blue}{4.7}/6.3 \textcolor{blue}{6.3}/9.5 \textcolor{blue}{7.5}/88.1 \textcolor{blue}{3.5} \\

\midrule
& YCB$_{bg}$-S & ScanNet$_{bg}$-S & COCO$_{bg}$-S  \\ \midrule
              & AP / PQ / Pre / Rec / BG-Rec & AP / PQ / Pre / Rec / BG-Rec & AP / PQ / Pre / Rec / BG-Rec \\ 
IODINE & 0.2\textcolor{blue}{0.3}/2.4 \textcolor{blue}{1.1}/3.4 \textcolor{blue}{1.5}/4.9 \textcolor{blue}{1.9}/12.5 \textcolor{blue}{5.6}& 2.5 \textcolor{blue}{2.3}/7.6 \textcolor{blue}{5.8}/10.2 \textcolor{blue}{6.8}/14.5 \textcolor{blue}{11.9}/25.9 \textcolor{blue}{10.3} & 0.2 \textcolor{blue}{0.2}/1.9 \textcolor{blue}{1.0}/2.5 \textcolor{blue}{1.2}/4.1 \textcolor{blue}{1.9}/10.4 \textcolor{blue}{0.8}\\
SlotAtt & 0.1 \textcolor{blue}{0.1}/1.6 \textcolor{blue}{0.6}/2.4 \textcolor{blue}{0.9}/3.3 \textcolor{blue}{1.5}/0.0 \textcolor{blue}{0.0} & 1.2 \textcolor{blue}{0.1}/4.6 \textcolor{blue}{0.9}/6.2 \textcolor{blue}{1.0}/9.5 \textcolor{blue}{2.1}/3.3 \textcolor{blue}{2.1}& 0.1 \textcolor{blue}{0.1}/1.2 \textcolor{blue}{0.0}/1.6 \textcolor{blue}{0.0}/2.8 \textcolor{blue}{0.0}/0.1 \textcolor{blue}{0.1}\\

\midrule
& YCB$_{bg}$-C+T & ScanNet$_{bg}$-C+T & COCO$_{bg}$-C+T  \\ \midrule
              & AP / PQ / Pre / Rec / BG-Rec & AP / PQ / Pre / Rec / BG-Rec & AP / PQ / Pre / Rec / BG-Rec \\ 
IODINE & 2.4 \textcolor{blue}{1.8}/5.2 \textcolor{blue}{0.6}/8.1 \textcolor{blue}{0.8}/9.9 \textcolor{blue}{0.7}/95.1 \textcolor{blue}{2.2}& 11.9 \textcolor{blue}{1.1}/17.1 \textcolor{blue}{1.3}/23.0 \textcolor{blue}{0.5}/29.3 \textcolor{blue}{3.1}/90.7 \textcolor{blue}{1.0}& 2.5 \textcolor{blue}{2.0}/6.3 \textcolor{blue}{0.5}/10.7 \textcolor{blue}{0.2}/9.9 \textcolor{blue}{1.6}/94.5 \textcolor{blue}{3.8}\\
SlotAtt & 14.1 \textcolor{blue}{4.1}/18.1 \textcolor{blue}{2.1}/25.0 \textcolor{blue}{3.1}/32.2\textcolor{blue}{1.8}/92.2 \textcolor{blue}{1.1}& 4.5 \textcolor{blue}{1.1}/10.1 \textcolor{blue}{2.0}/13.8 \textcolor{blue}{2.6}/19.1 \textcolor{blue}{3.3}/71.7 \textcolor{blue}{12.2}& 4.9 \textcolor{blue}{2.9}/7.8 \textcolor{blue}{1.6}/10.3 \textcolor{blue}{1.6}/15.1 \textcolor{blue}{3.0}/96.7 \textcolor{blue}{8.7}\\

\midrule
& YCB$_{bg}$-C+S & ScanNet$_{bg}$-C+S & COCO$_{bg}$-C+S  \\ \midrule
              & AP / PQ / Pre / Rec / BG-Rec & AP / PQ / Pre / Rec / BG-Rec & AP / PQ / Pre / Rec / BG-Rec \\ 
IODINE & 8.3 \textcolor{blue}{1.3}/10.0 \textcolor{blue}{0.6}/16.7 \textcolor{blue}{0.1}/15.6 \textcolor{blue}{2.1}/99.1 \textcolor{blue}{0.7}& 13.6 \textcolor{blue}{3.3}/17.7 \textcolor{blue}{2.0}/27.7 \textcolor{blue}{1.2}/26.1 \textcolor{blue}{4.3}/94.1 \textcolor{blue}{1.8}& 3.7 \textcolor{blue}{1.0}/8.4 \textcolor{blue}{0.2}/17.9 \textcolor{blue}{0.9}/10.7 \textcolor{blue}{0.7}/95.7 \textcolor{blue}{2.5}\\
SlotAtt & 0.2 \textcolor{blue}{0.5}/1.4 \textcolor{blue}{0.5}/2.2 \textcolor{blue}{1.5}/2.5 \textcolor{blue}{1.4}/67.4 \textcolor{blue}{1.2}& 0.4 \textcolor{blue}{1.1}/2.3 \textcolor{blue}{0.7}/4.5 \textcolor{blue}{2.1}/3.2 \textcolor{blue}{1.6}/ 66.5 \textcolor{blue}{2.4} & 3.5 \textcolor{blue}{1.0}/ 7.9 \textcolor{blue}{1.4}/ 11.2 \textcolor{blue}{1.8}/ 15.6 \textcolor{blue}{3.4}/ 93.1 \textcolor{blue}{3.1}\\

\midrule
& YCB$_{bg}$-T+S & ScanNet$_{bg}$-T+S & COCO$_{bg}$-T+S  \\ \midrule
              & AP / PQ / Pre / Rec / BG-Rec & AP / PQ / Pre / Rec / BG-Rec & AP / PQ / Pre / Rec / BG-Rec \\ 
IODINE & 2.0 \textcolor{blue}{1.4}/7.3 \textcolor{blue}{0.8}/10.6 \textcolor{blue}{1.8}/13.2 \textcolor{blue}{1.4}/92.8 \textcolor{blue}{5.3}& 12.2 \textcolor{blue}{1.0}/12.2 \textcolor{blue}{1.4}/16.4 \textcolor{blue}{2.4}/22.6 \textcolor{blue}{1.3}/86.0 \textcolor{blue}{1.3}& 2.3 \textcolor{blue}{0.4}/7.1 \textcolor{blue}{0.4}/10.4 \textcolor{blue}{0.5}/13.0 \textcolor{blue}{0.8}/88.8 \textcolor{blue}{0.0}\\
SlotAtt & 7.7 \textcolor{blue}{0.0}/13.5 \textcolor{blue}{0.0}/ 8.6 \textcolor{blue}{0.2}/26.6 \textcolor{blue}{0.2}/92.6 \textcolor{blue}{0.0} & 7.4 \textcolor{blue}{1.0}/11.0 \textcolor{blue}{0.3}/14.7 \textcolor{blue}{0.2}/19.7 \textcolor{blue}{0.2}/60.4 \textcolor{blue}{2.6}& 10.0 \textcolor{blue}{4.0}/13.9 \textcolor{blue}{7.7}/17.7 \textcolor{blue}{9.0}/26.2 \textcolor{blue}{7.0}/89.5 \textcolor{blue}{1.5}\\

\midrule
& YCB$_{bg}$-C+T+S & ScanNet$_{bg}$-C+T+S & COCO$_{bg}$-C+T+S  \\ \midrule
              & AP / PQ / Pre / Rec / BG-Rec & AP / PQ / Pre / Rec / BG-Rec & AP / PQ / Pre / Rec / BG-Rec \\ 
IODINE & 3.0 \textcolor{blue}{1.5}/5.9 \textcolor{blue}{2.5}/9.3 \textcolor{blue}{3.6}/11.3 \textcolor{blue}{3.5}/99.4 \textcolor{blue}{1.9}& 12.0 \textcolor{blue}{0.9}/15.8 \textcolor{blue}{3.0}/21.3 \textcolor{blue}{4.5}/28.0 \textcolor{blue}{3.5}/93.7 \textcolor{blue}{0.5}& 6.4 \textcolor{blue}{1.3}/10.9 \textcolor{blue}{0.3}/16.7 \textcolor{blue}{0.2}/17.7 \textcolor{blue}{0.4}/97.9 \textcolor{blue}{3.8}\\
SlotAtt & 14.8 \textcolor{blue}{4.2}/15.2 \textcolor{blue}{1.5}/20.4 \textcolor{blue}{3.1}/28.3 \textcolor{blue}{4.1}/90.0 \textcolor{blue}{4.2}&  2.1 \textcolor{blue}{0.3}/6.2 \textcolor{blue}{0.4}/8.5 \textcolor{blue}{0.4}/11.9 \textcolor{blue}{0.5}/36.3 \textcolor{blue}{9.5}&2.4 \textcolor{blue}{2.1}/6.0 \textcolor{blue}{2.8}/8.2 \textcolor{blue}{3.7}/12.0 \textcolor{blue}{4.8}/84.5 \textcolor{blue}{1.1}\\
\bottomrule
  \end{tabular}}
\end{table*}

\subsection{Additional Joint Ablations on Object- and Scene-Level Factors} \label{sec:joint_ablation_2}
\begin{figure*}[th]
    \setlength{\abovecaptionskip}{ 4 pt}
    \setlength{\belowcaptionskip}{ -6 pt}
    \centering
    \includegraphics[width=0.85\linewidth]{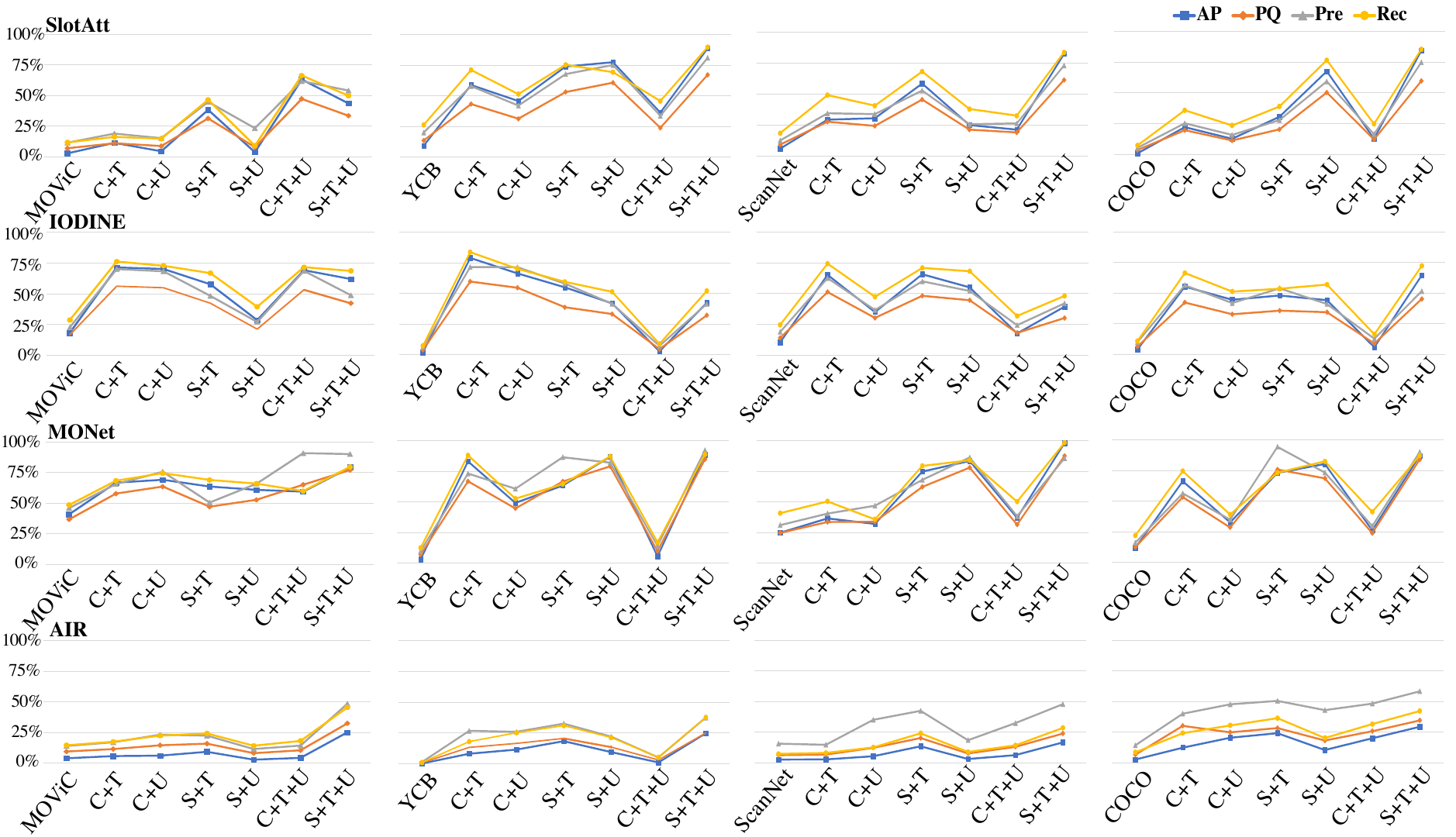} 
    \caption{\hl{Quantitative results of baselines on Group 2 datasets and their variants in Section} \ref{sec:joint_ablation_2}}
\label{fig:app_hybrid_ablation_stats}
\end{figure*}

\begin{figure*}[th]
    \setlength{\abovecaptionskip}{ 4 pt}
    \setlength{\belowcaptionskip}{ -6 pt}
    \centering
    \includegraphics[width=0.85\linewidth]{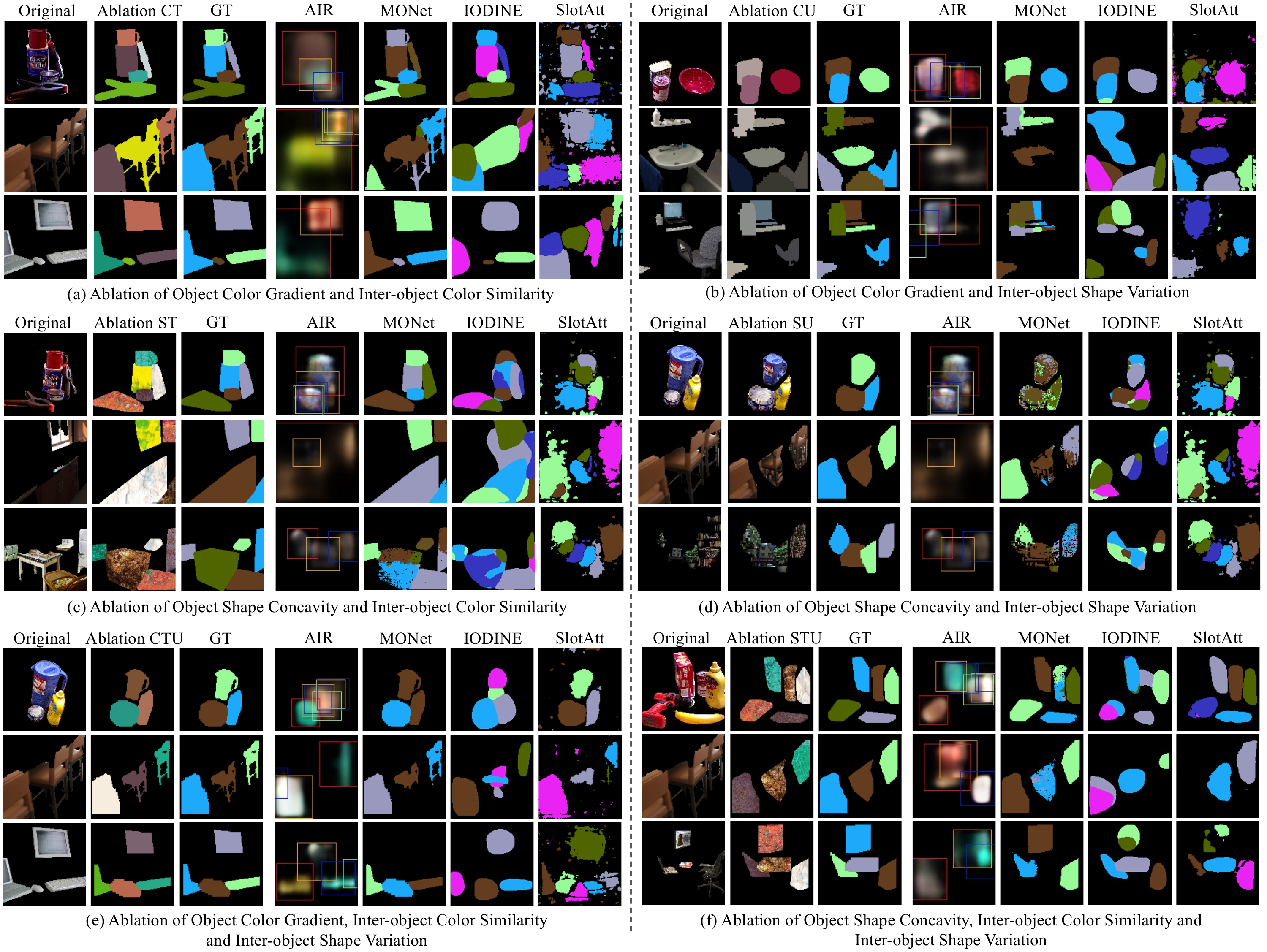} 
    \caption{Qualitative results of additional joint ablations on both object- and scene-level factors in Section \ref{sec:joint_ablation_2}.}
\label{fig:additional_joint_ablation_vis}
\end{figure*}

In addition to the experiments in Section \ref{sec:exp_joint_factors}, we generate additional 6 groups of datasets ablated with different combinations of object- and scene-level factors as follows and conduct corresponding experiments. Experiment results can be found in Figures \ref{fig:app_hybrid_ablation_stats}\&\ref{fig:additional_joint_ablation_vis}, and Table \ref{tab:app_hybrid_ablation_additional}.

\paragraph{Additional Ablated Datasets:} \label{sec:add_exp_joint_ablation}
\begin{itemize}[leftmargin=*]
\setlength{\itemsep}{1pt}
\setlength{\parsep}{1pt}
\setlength{\parskip}{1pt}
    \item \textit{Ablation of Object Color Gradient and Inter-object Color Similarity}: In each image of the three real-world datasets, we replace the object color by averaging all pixels of a distinctive texture, and keep the original shape unchanged, getting three ablated datasets: YCB-C+T/ScanNet-C+T/COCO-C+T.   
    \item \textit{Ablation of Object Color Gradient and Inter-object Shape Variation}: In each image of the three real-world datasets, we replace the object color by averaging its own texture, and then apply size normalization on the original shape of objects, getting three ablated datasets: YCB-C+U/ScanNet-C+U/COCO-C+U.  
    \item \textit{Ablation of Object Color Gradient, Inter-object Color Similarity and Inter-object Shape Variation}: In each image of the three real-world datasets, we replace the object color by averaging all pixels of a distinctive texture, and then apply size normalization on the original shape of objects. The ablated datasets are denoted as: YCB-C+T+U/ScanNet-C+T+U/COCO-C+T+U.   
    \item \textit{Ablation of Object Shape Concavity and Inter-object Color Similarity}: In each image of the three real-world datasets, we replace the object color by a distinctive texture, and modify the object shape as a convex hull, getting three ablated datasets: YCB-S+T/ScanNet-S+T/COCO-S+T.   
    \item \textit{Ablation of  Object Shape Concavity and Inter-object Shape Variation}: In each image of the three real-world datasets, we keep the texture of object unchanged, and modify the object shape as a convex hull followed by size normalization, getting three ablated datasets: YCB-S+U/ScanNet-S+U/COCO-S+U.
    \item \textit{Ablation of Object Shape Concavity, Inter-object Color Similarity and Inter-object Shape Variation}: In each image of the three real-world datasets, we replace the object color with a distinctive texture, and modify the object shape as a convex hull followed by size normalization. The ablated datasets are denoted as: YCB-S+T+U/ScanNet-S+T+U/COCO-S+T+U. 
\end{itemize}

\paragraph{Qualitative and Quantitative Results:} \label{sec:add_res_joint_ablation}
As shown in Table \ref{tab:app_hybrid_ablation_additional}, Figures \ref{fig:app_hybrid_ablation_stats}\&\ref{fig:additional_joint_ablation_vis}, all 6 additional combinations of object- and scene-level factors are explored, demonstrating consistent findings as our experiments in Section \ref{sec:exp_joint_factors}. Overall, all four methods show a high sensitivity to both object- and scene-level factors relating to appearance. This can be seen from the fact that: for datasets without ablations in appearance, \ie{}, the (S+U) ablated datasets, the object segmentation performance is inferior. By contrast, the object segmentation accuracy can be greatly improved on the datasets only with appearance factors ablated, \ie{}, the (C+T) datasets. Meanwhile, more regular shapes and uniform scales of objects still have a significant positive influence on the success of object segmentation especially when the appearance factors are combined in ablated datasets. 

To be specific, AIR \citep{Eslami2016} is quite sensitive to the scale of objects apart from object color gradient and inter-object color similarity. MONet \citep{Burgess2019} can obtain comparable performance to the simple synthetic datasets once object color gradient and inter-object color similarity are ablated. All four factors are closely relevant to the results of IODINE \citep{Greff2019} and SlotAtt \citep{Locatello2020}.

\begin{table*}[th]
\centering
\caption{Quantitative results on additional datasets ablated with both object- and scene-level factors. Standard deviations are calculated over 3 runs (marked with \textcolor{blue}{blue}).}
  \label{tab:app_hybrid_ablation_additional}
\resizebox{\textwidth}{!}{
  \begin{tabular}{r c|c|c} \toprule
& YCB-C+T & ScanNet-C+T & COCO-C+T  \\ \midrule
              & AP / PQ / Pre / Rec  & AP / PQ / Pre / Rec & AP / PQ / Pre / Rec \\ 
AIR & 8.0 \textcolor{blue}{6.1} / 13.2 \textcolor{blue}{7.3} / 26.6 \textcolor{blue}{17.6} / 17.9 \textcolor{blue}{7.3} & 2.8 \textcolor{blue}{4.7} / 6.7 \textcolor{blue}{6.5} / 14.8 \textcolor{blue}{4.0} / 8.1 \textcolor{blue}{14.8} & 12.4 \textcolor{blue}{9.4} / 30.2 \textcolor{blue}{23.0} / 40.1 \textcolor{blue}{25.6} / 24.2 \textcolor{blue}{13.8} \\ 
MONet & 82.9 \textcolor{blue}{9.6} / 66.7 \textcolor{blue}{0.5} / 73.0 \textcolor{blue}{1.2} / 87.8 \textcolor{blue}{12.2} & 36.5 \textcolor{blue}{28.4} / 33.5 \textcolor{blue}{21.8} / 40.4 \textcolor{blue}{24.1} / 50.3 \textcolor{blue}{19.7} & 66.1 \textcolor{blue}{5.6} / 53.2 \textcolor{blue}{3.7} / 56.2 \textcolor{blue}{2.3} / 74.4 \textcolor{blue}{6.4} \\ 
IODINE & 78.9 \textcolor{blue}{2.6} / 59.7 \textcolor{blue}{2.1} / 71.5 \textcolor{blue}{0.6} / 83.6 \textcolor{blue}{2.7} & 65.1 \textcolor{blue}{2.9} / 51.1 \textcolor{blue}{1.0} / 62.4 \textcolor{blue}{2.9} / 74.2 \textcolor{blue}{0.5}& 55.1 \textcolor{blue}{10.4} / 42.2 \textcolor{blue}{7.5} / 56.3 \textcolor{blue}{6.4} / 66.2 \textcolor{blue}{9.2} \\ 
SlotAtt & 58.7 \textcolor{blue}{12.1} / 43.2 \textcolor{blue}{5.0} / 57.8 \textcolor{blue}{9.4} / 71.0 \textcolor{blue}{9.4} & 29.2 \textcolor{blue}{2.9} / 27.6 \textcolor{blue}{1.3} / 34.4 \textcolor{blue}{1.4} / 49.2 \textcolor{blue}{1.2} & 22.1 \textcolor{blue}{8.6} / 19.6 \textcolor{blue}{4.1} / 25.2 \textcolor{blue}{4.1} / 35.8 \textcolor{blue}{8.1}\\
\midrule

& YCB-C+U  & ScanNet-C+U  & COCO-C+U   \\ \midrule
              & AP / PQ / Pre / Rec  & AP / PQ / Pre / Rec & AP / PQ / Pre / Rec  \\
AIR & 11.4 \textcolor{blue}{4.9} / 16.4 \textcolor{blue}{5.4} / 25.8 \textcolor{blue}{8.7} / 25.1 \textcolor{blue}{6.5} & 5.4 \textcolor{blue}{0.5} / 12.3 \textcolor{blue}{2.6} / 35.1 \textcolor{blue}{15.3} / 12.6 \textcolor{blue}{0.4} & 20.5 \textcolor{blue}{10.8} / 24.9 \textcolor{blue}{7.1} / 47.7 \textcolor{blue}{10.2} / 30.5 \textcolor{blue}{8.8}\\ 
MONet & 49.1 \textcolor{blue}{3.5} / 44.6 \textcolor{blue}{4.1} / 60.6 \textcolor{blue}{3.2} / 52.5 \textcolor{blue}{3.3} & 31.9 \textcolor{blue}{5.7} / 33.7 \textcolor{blue}{5.7} / 46.8 \textcolor{blue}{5.4} / 35.7 \textcolor{blue}{8.7} & 32.6 \textcolor{blue}{11.1} / 28.5 \textcolor{blue}{12.1} / 34.9 \textcolor{blue}{14.2} / 39.0 \textcolor{blue}{18.5}\\ 
IODINE & 66.3 \textcolor{blue}{0.1} / 54.6 \textcolor{blue}{0.0} / 71.3 \textcolor{blue}{4.1} / 70.0 \textcolor{blue}{0.2} & 34.8 \textcolor{blue}{20.9} / 30.0 \textcolor{blue}{14.7} / 36.4 \textcolor{blue}{20.9} / 47.2 \textcolor{blue}{14.0} & 44.4 \textcolor{blue}{8.9} / 32.6 \textcolor{blue}{7.9} / 41.7 \textcolor{blue}{12.3} / 51.1 \textcolor{blue}{7.6} \\ 
SlotAtt & 45.6 \textcolor{blue}{7.9} / 31.2 \textcolor{blue}{4.2} / 41.9 \textcolor{blue}{7.7} / 51.3 \textcolor{blue}{6.4} & 30.4 \textcolor{blue}{8.9} / 24.2 \textcolor{blue}{6.2} / 33.8 \textcolor{blue}{9.5} / 40.6 \textcolor{blue}{11.0} & 12.7 \textcolor{blue}{7.7} / 11.3 \textcolor{blue}{3.7} / 15.8 \textcolor{blue}{5.0} / 23.5 \textcolor{blue}{6.8} \\ 
\midrule

& YCB-S+T & ScanNet-S+T & COCO-S+T  \\ \midrule
              & AP / PQ / Pre / Rec  & AP / PQ / Pre / Rec & AP / PQ / Pre / Rec \\ 
AIR & 9.3 \textcolor{blue}{7.8} / 13.4 \textcolor{blue}{8.4} / 21.9 \textcolor{blue}{14.3} / 21.2 \textcolor{blue}{12.3} & 3.2 \textcolor{blue}{0.5} / 7.7 \textcolor{blue}{0.6} / 18.5 \textcolor{blue}{6.0} / 8.9 \textcolor{blue}{6.0} & 10.5 \textcolor{blue}{5.4} / 18.2 \textcolor{blue}{7.2} / 4.3 \textcolor{blue}{23.2} / 20.3 \textcolor{blue}{7.3} \\ 
MONet & 86.7 \textcolor{blue}{2.2} / 78.9 \textcolor{blue}{1.6} / 81.9 \textcolor{blue}{1.8} / 87.0 \textcolor{blue}{2.2} & 83.2 \textcolor{blue}{3.7} / 77.8 \textcolor{blue}{13.0} / 86.0 \textcolor{blue}{20.0} / 83.8 \textcolor{blue}{2.9} & 80.2 \textcolor{blue}{0.0} / 68.5 \textcolor{blue}{7.2} / 73.2 \textcolor{blue}{11.7} / 82.3 \textcolor{blue}{1.0} \\ 
IODINE & 41.9 \textcolor{blue}{17.9} / 33.2 \textcolor{blue}{11.4} / 41.5 \textcolor{blue}{12.0} / 51.3 \textcolor{blue}{14.7} & 54.9 \textcolor{blue}{23.1} / 44.4 \textcolor{blue}{16.2} / 52.0 \textcolor{blue}{16.7} / 68.0 \textcolor{blue}{21.6} & 44.1 \textcolor{blue}{1.1} / 34.3 \textcolor{blue}{0.7} / 41.0 \textcolor{blue}{0.3} / 56.8 \textcolor{blue}{0.6} \\ 
SlotAtt & 77.3 \textcolor{blue}{7.8} / 60.6 \textcolor{blue}{1.5} / 75.0 \textcolor{blue}{2.3} / 69.2 \textcolor{blue}{19.7} & 24.9 \textcolor{blue}{46.1} / 21.2 \textcolor{blue}{35.0} / 25.5 \textcolor{blue}{38.3} / 37.9 \textcolor{blue}{43.2} & 67.4 \textcolor{blue}{3.1} / 50.2 \textcolor{blue}{1.6} / 59.2 \textcolor{blue}{1.6} / 76.4 \textcolor{blue}{0.3} \\ 
\midrule

& YCB-S+U  & ScanNet-S+U  & COCO-S+U   \\ \midrule
              & AP / PQ / Pre / Rec  & AP / PQ / Pre / Rec & AP / PQ / Pre / Rec  \\
AIR & 0.8 \textcolor{blue}{5.4} / 2.9 \textcolor{blue}{7.1} / 5.0 \textcolor{blue}{10.0} / 5.2 \textcolor{blue}{12.7} & 6.3 \textcolor{blue}{0.2} / 13.1 \textcolor{blue}{3.3} / 32.4 \textcolor{blue}{16.3} / 14.3 \textcolor{blue}{1.2} & 20.0 \textcolor{blue}{9.5} / 25.8 \textcolor{blue}{10.0} / 48.3 \textcolor{blue}{15.6} / 31.6 \textcolor{blue}{10.9} \\ 
MONet & 5.5 \textcolor{blue}{1.2} / 9.8 \textcolor{blue}{0.4} / 14.1 \textcolor{blue}{0.8} / 17.0 \textcolor{blue}{0.2} & 36.9 \textcolor{blue}{2.9} / 31.4 \textcolor{blue}{2.6} / 38.2 \textcolor{blue}{3.6} / 50.1 \textcolor{blue}{0.1} & 26.1 \textcolor{blue}{3.5} / 23.8 \textcolor{blue}{1.8} / 29.6 \textcolor{blue}{1.6} / 41.2 \textcolor{blue}{10.0} \\ 
IODINE & 2.8 \textcolor{blue}{2.4} / 4.6 \textcolor{blue}{2.0} / 7.2 \textcolor{blue}{2.9} / 8.9 \textcolor{blue}{3.6} & 17.3 \textcolor{blue}{2.4} / 17.8 \textcolor{blue}{1.4} / 24.1 \textcolor{blue}{1.8} / 31.6 \textcolor{blue}{2.1} & 5.7 \textcolor{blue}{0.5} / 8.7 \textcolor{blue}{1.7} / 1.3 \textcolor{blue}{13.7} / 16.3 \textcolor{blue}{2.4} \\ 
SlotAtt & 36.2 \textcolor{blue}{3.3} / 23.8 \textcolor{blue}{4.1} / 33.6 \textcolor{blue}{5.5} / 45.6 \textcolor{blue}{1.7} & 21.1 \textcolor{blue}{0.9} / 18.9 \textcolor{blue}{0.2} / 26.2 \textcolor{blue}{0.1} / 32.5 \textcolor{blue}{2.0} & 12.7 \textcolor{blue}{7.1} / 12.1 \textcolor{blue}{2.3} / 16.4 \textcolor{blue}{2.1} / 24.7 \textcolor{blue}{5.2} \\ 
\midrule
      
& YCB-C+T+U & ScanNet-C+T+U & COCO-C+T+U  \\ \midrule
              & AP / PQ / Pre / Rec  & AP / PQ / Pre / Rec & AP / PQ / Pre / Rec \\ 
AIR & 18.2 \textcolor{blue}{12.0} / 20.5 \textcolor{blue}{7.9} / 32.5 \textcolor{blue}{13.5} / 30.9 \textcolor{blue}{9.0} & 13.5 \textcolor{blue}{6.6} / 20.2 \textcolor{blue}{9.1} / 42.4 \textcolor{blue}{25.5} / 24.1 \textcolor{blue}{5.6} & 24.2 \textcolor{blue}{11.1} / 28.3 \textcolor{blue}{8.5} / 50.5 \textcolor{blue}{13.1} / 36.4 \textcolor{blue}{9.7} \\ 
MONet & 63.2 \textcolor{blue}{6.9} / 66.5 \textcolor{blue}{11.8} / 86.3 \textcolor{blue}{13.0} / 64.4 \textcolor{blue}{5.7} & 74.6 \textcolor{blue}{5.6} / 62.1 \textcolor{blue}{6.6} / 67.8 \textcolor{blue}{7.2} / 79.2 \textcolor{blue}{3.2} & 73.0 \textcolor{blue}{3.3} / 75.7 \textcolor{blue}{10.3} / 94.2 \textcolor{blue}{23.1} / 73.1 \textcolor{blue}{7.5} \\ 
IODINE & 54.9 \textcolor{blue}{23.1} / 38.7 \textcolor{blue}{18.2} / 58.0 \textcolor{blue}{12.0} / 59.4 \textcolor{blue}{20.8} & 65.5 \textcolor{blue}{3.8} / 48.0 \textcolor{blue}{3.2} / 59.8 \textcolor{blue}{5.9} / 70.7 \textcolor{blue}{2.9} & 47.9 \textcolor{blue}{11.7} / 35.5 \textcolor{blue}{7.4} / 53.8 \textcolor{blue}{2.1} / 53.4 \textcolor{blue}{14.2} \\ 
SlotAtt & 73.8 \textcolor{blue}{4.6} / 53.0 \textcolor{blue}{2.9} / 67.6 \textcolor{blue}{7.8} / 75.2 \textcolor{blue}{4.5} & 58.5 \textcolor{blue}{8.1} / 45.5 \textcolor{blue}{10.2} / 52.9 \textcolor{blue}{8.9} / 68.3 \textcolor{blue}{11.9} & 30.5 \textcolor{blue}{27.6} / 20.3 \textcolor{blue}{18.0} / 27.8 \textcolor{blue}{23.0} / 38.9 \textcolor{blue}{23.3} \\ 
\midrule

& YCB-S+T+U  & ScanNet-S+T+U  & COCO-S+T+U   \\ \midrule
              & AP / PQ / Pre / Rec  & AP / PQ / Pre / Rec & AP / PQ / Pre / Rec  \\
AIR & 24.6 \textcolor{blue}{20.0} / 24.8 \textcolor{blue}{13.7} / 37.4 \textcolor{blue}{20.7} / 37.6 \textcolor{blue}{18.6} & 16.7 \textcolor{blue}{10.1} / 23.9 \textcolor{blue}{12.0} / 47.8 \textcolor{blue}{30.2} / 28.5 \textcolor{blue}{7.8} & 29.4 \textcolor{blue}{16.5} / 34.6 \textcolor{blue}{18.4} / 58.3 \textcolor{blue}{35.7} / 42.2 \textcolor{blue}{13.4} \\ 
MONet & 88.4 \textcolor{blue}{0.6} / 84.6 \textcolor{blue}{1.5} / 92.0 \textcolor{blue}{1.5} / 88.6 \textcolor{blue}{0.7} & 97.8 \textcolor{blue}{0.3} / 87.3 \textcolor{blue}{0.8} / 85.4 \textcolor{blue}{0.1} / 98.1 \textcolor{blue}{0.4} & 86.9 \textcolor{blue}{0.1} / 83.8 \textcolor{blue}{11.6} / 89.9 \textcolor{blue}{13.1} / 87.0 \textcolor{blue}{1.5} \\ 
IODINE & 42.3 \textcolor{blue}{15.0} / 32.1 \textcolor{blue}{9.7} / 41.0 \textcolor{blue}{11.5} / 52.0 \textcolor{blue}{15.7} & 39.1 \textcolor{blue}{4.7} / 29.9 \textcolor{blue}{1.8} / 41.8 \textcolor{blue}{7.2} / 47.9 \textcolor{blue}{2.4} & 64.3 \textcolor{blue}{9.7} / 45.1 \textcolor{blue}{6.9} / 51.4 \textcolor{blue}{5.6} / 72.0 \textcolor{blue}{11.7} \\ 
SlotAtt & 88.6 \textcolor{blue}{6.5} / 67.0 \textcolor{blue}{9.6} / 80.9 \textcolor{blue}{13.0} / 89.6 \textcolor{blue}{5.6} & 82.6 \textcolor{blue}{8.5} / 61.5 \textcolor{blue}{10.1} / 73.2 \textcolor{blue}{10.2} / 83.9 \textcolor{blue}{7.7} & 84.2 \textcolor{blue}{0.9} / 59.6 \textcolor{blue}{6.8} / 74.7 \textcolor{blue}{3.8} / 85.3 \textcolor{blue}{0.6}\\  \bottomrule
               
  \end{tabular}}
\end{table*}

\begin{figure}
    \setlength{\abovecaptionskip}{ 4 pt}
    \setlength{\belowcaptionskip}{ -8 pt}
\centering
\includegraphics[width=0.45\textwidth]{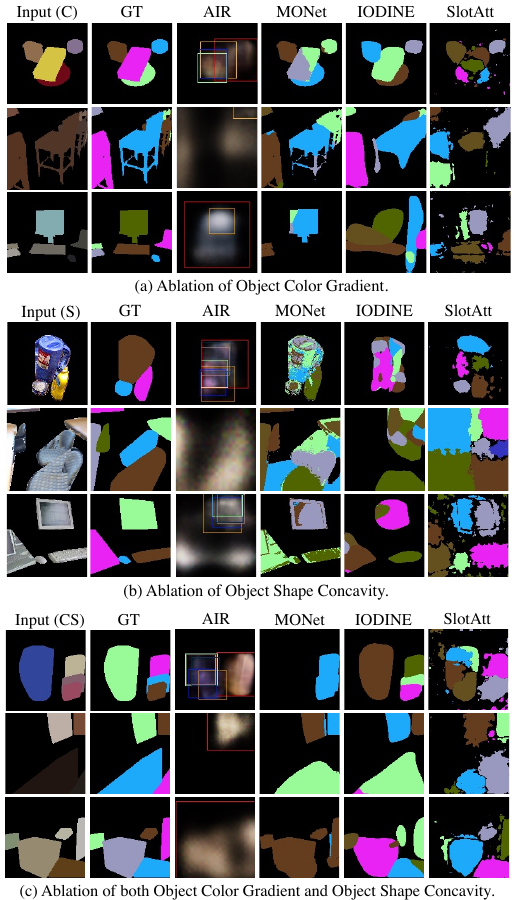}
\caption{Qualitative results of object-level factors ablations.}
\vspace{-0.3cm}
\label{fig:object_level_ablation_vis}
\end{figure}

\begin{figure}
     \centering
         \centering
         \includegraphics[width=0.45\textwidth]{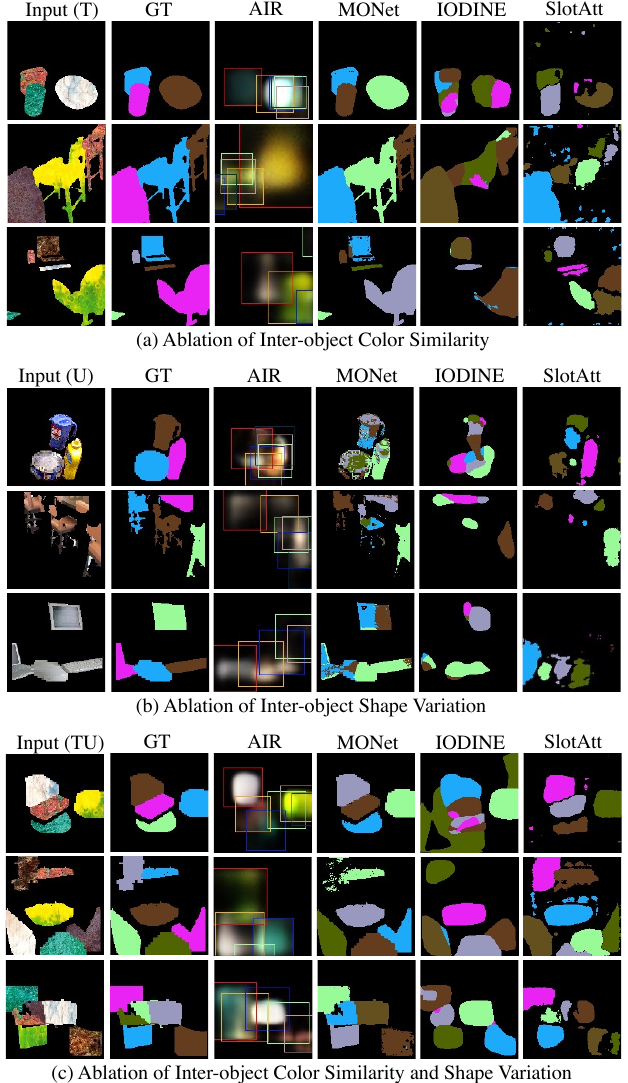}
\caption{Qualitative results of scene-level factors ablations.}
\label{fig:scene_level_ablation_vis}
\end{figure}

\begin{figure}
\centering
\includegraphics[width=0.45\textwidth]{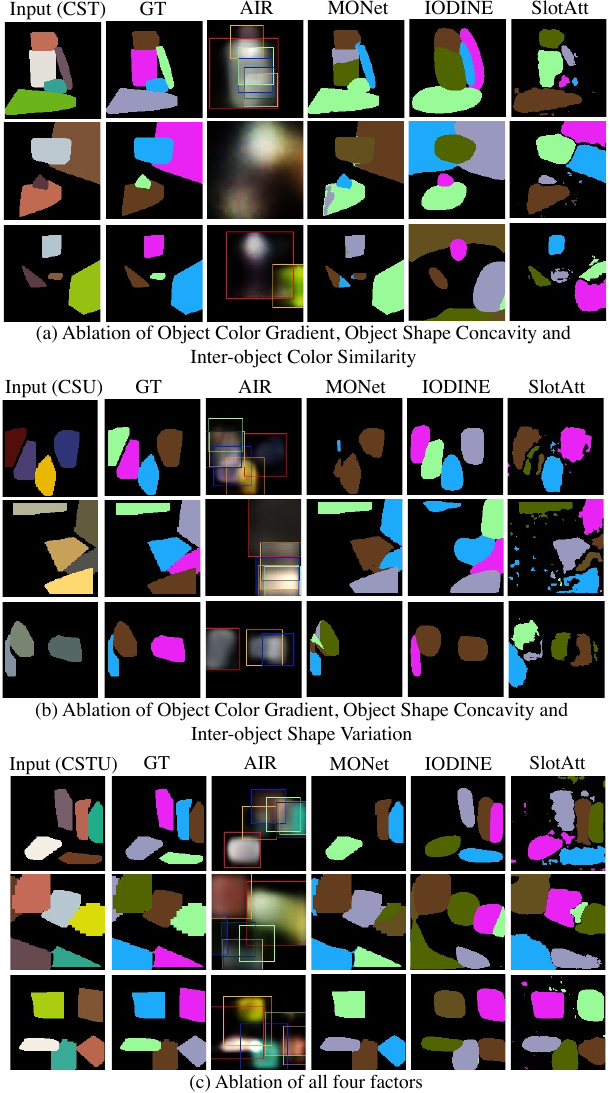}
\caption{Qualitative results of joint ablation on object- and scene-level factors.}
\label{fig:object_scene_level_ablation_vis}
\vspace{-0.2cm}
\end{figure}

\end{document}